\documentclass[12pt]{article}

\newtheorem{theorem}{Theorem}[section]
\newtheorem{definition}{Definition}[section]
\newtheorem{lemma}{Lemma}[section]

\newtheorem{remark}{Remark}[section]

\usepackage{graphicx}
\usepackage{amssymb,amsmath}
\usepackage[numbers]{natbib}
\oddsidemargin
-0.2in \topmargin -.60in
\begin{document}

\title{Generalization Bounds for Representative Domain Adaptation}

\small{\author{
Chao~Zhang$^{1}$,\quad Lei~Zhang$^{2}$,\quad Wei Fan$^{3}$, Jieping Ye$^{1,4}$\\
$^1$Center for Evolutionary
Medicine and Informatics, The Biodesign Institute, \\and $^4$Computer Science and Engineering,
Arizona State University,
Tempe, USA\\
\texttt{zhangchao1015@gmail.com; jieping.ye@asu.edu} \\
$^{2}$School of Computer Science and Technology,\\
Nanjing University of Science and Technology,
Nanjing, P.R. China\\
\texttt{zhanglei.njust@yahoo.com.cn} \\
$^{3}$Huawei Noah¡¯s Ark Lab,
Hong Kong, China\\
\texttt{david.fanwei@huawei.com} \\
}
}

%       \AND
% \name Lei Zhang \email zhanglei.njust@yahoo.com.cn \\
%\addr School of Computer Science and Technology \\
%Nanjing University of Science and Technology\\
%Nanjing, Jiangsu 210094, P.R. China
%       \AND
%       \name Jieping Ye \email jieping.ye@asu.edu \\
%      \addr  Center for Evolutionary Medicine and Informatics, The Biodesign Institute,\\
%        and Computer Science and Engineering, Arizona State University\\
%       Tempe, AZ 85287, U.S.A.}

\maketitle

\begin{abstract}%   <- trailing '%' for backward compatibility of .sty file
In this paper, we propose a novel framework to analyze the theoretical properties of the learning process for a representative type of domain adaptation, which combines data from multiple sources and one target (or briefly called representative domain adaptation). In particular, we use the integral probability metric to measure the difference between the distributions of two domains and meanwhile compare it with the $\mathcal{H}$-divergence and the discrepancy distance. We develop the Hoeffding-type, the Bennett-type and the McDiarmid-type deviation inequalities for multiple domains respectively, and then present the symmetrization inequality for representative domain adaptation. Next, we use the derived inequalities to obtain the Hoeffding-type and the Bennett-type generalization bounds respectively, both of which are based on the uniform entropy number. Moreover, we present the generalization bounds based on the Rademacher complexity. Finally, we analyze the asymptotic convergence and the rate of convergence of the learning process for representative domain adaptation. We discuss the factors that affect the asymptotic behavior of the learning process and the numerical experiments support our theoretical findings as well.
Meanwhile, we give a comparison with the existing results of domain adaptation and the classical results
under the same-distribution assumption.
\end{abstract}


%%%

\section{Introduction}

The generalization bound measures the probability that a function, chosen from a function class by an algorithm, has a sufficiently small error and it plays an important role in statistical learning theory \citep[see][]{Vapnik98,Bousquet04}. The generalization bounds have been widely used to study the consistency of the ERM-based learning process \citep{Vapnik98}, the asymptotic convergence of empirical process \citep{Vaart96} and the learnability of learning models \citep{Blumer89}. Generally, there are three essential aspects to obtain the generalization bounds of a specific learning process: complexity measures of function classes, deviation (or concentration) inequalities and symmetrization inequalities related to the learning process. For example, \citet{Vaart96} presented the generalization bounds based on the Rademacher complexity and the covering number, respectively. \citet{Vapnik98} gave the generalization bounds based on the Vapnik-Chervonenkis (VC) dimension. \citet{Bartlett05} proposed the local Rademacher complexity and obtained a sharp generalization bound for a particular function class $\{f\in\mathcal{F}:\mathrm{E}f^2<\beta\mathrm{E}f, \beta>0\}$. \citet{Hussain11} showed improved loss bounds for multiple kernel learning. \citet{zhang2013bennett}
analyzed the Bennett-type generalization bounds of the i.i.d. learning process.

It is noteworthy that the aforementioned results of statistical learning theory are all built under the assumption that training and test data are drawn from the same distribution (or briefly called the same-distribution assumption). This assumption may not be valid in the situation that training and test data have different distributions, which will arise in many practical applications including speech recognition \citep{Jiang07} and natural language processing \citep{Blitzer07a}. Domain adaptation has recently been proposed to handle this situation and it is aimed to apply a learning model, trained by using the samples drawn from a certain domain ({\it source domain}), to the samples drawn from another domain ({\it target domain}) with a different distribution \citep[see][]{Bickel07,Wu04,Blitzer06,Ben-David10,Bian12}. There have been some research works on the theoretical analysis of two types of domain adaptation. In the first type, the learner receives training data from several source domains, known as {\it domain adaptation with multiple sources} \citep[see][]{Ben-David10,Crammer07,Crammer08,Mansour08,Mansour09,zhang2012generalization}. In the second type, the learner minimizes a convex combination of the source and the target empirical risks, termed as {\it domain adaptation combining source and target data} \citep[see][]{Blitzer07,Ben-David10,zhang2012generalization}.

Without loss of generality, this paper is mainly concerned with a more representative (or general) type of domain adaptation, which combines data from multiple sources and one target (or briefly called representative domain adaptation). Evidently, it covers both of the aforementioned two types: domain adaptation with multiple sources and domain adaptation combining source and target. Thus, the results of this paper are more general than the previous works and some of existing results can be regarded as the special cases of this paper \citep[see][]{zhang2012generalization}. We brief the main contributions of this paper as follows.

\subsection{Overview of Main Results}

In this paper, we present a new framework to obtain the generalization bounds of the learning process for representative domain adaptation. Based on the resulting bounds, we then analyze the asymptotical properties of the learning process.
There are four major aspects in the framework: (i) the quantity measuring the difference between two domains; (ii) the complexity measure of function classes; (iii) the deviation inequalities for multiple domains; (iv) the symmetrization inequality for representative domain adaptation.

As shown in some previous works \citep{Mansour09,Mansour09b,Ben-David10}, one of the major challenges in the theoretical analysis of domain adaptation is to measure the difference between two domains. Different from the previous works, we use the integral probability metric to measure the difference between the distributions of two domains. Moreover, we also give a comparison with the quantities proposed in the previous works.

Generally, in order to obtain the generalization bounds of a learning process, one needs to develop the related deviation (or concentration) inequalities of the learning process. Here, we use a martingale method to develop the related Hoeffding-type, Bennett-type and McDiarmid-type deviation inequalities for multiple domains, respectively. Moreover, in the situation of domain adaptation, since the source domain differs from the target domain, the desired symmetrization inequality for domain adaptation should incorporate some quantity to reflect the difference. From this point of view, we then obtain the related symmetrization inequality incorporating the integral probability metric that measures the difference between the distributions of the source and the target domains.

By applying the derived inequalities, we obtain two types of generalization bounds of the learning process for representative domain adaptation: Hoeffding-type and Bennett-type, both of which are based on the uniform entropy number. Moreover, we use the McDiarmid-type deviation inequality to obtain the generalization bounds based on the Rademacher complexity.
It is noteworthy that, based on the relationship between the integral probability metric and the discrepancy distance (or $\mathcal{H}$-divergence), the proposed framework can also lead to the generalization bounds by incorporating the discrepancy distance  (or $\mathcal{H}$-divergence) [see Section \ref{sec:IPM} and Remark \ref{rem:result}].

Based on the resulting generalization bounds, we study the asymptotic convergence and the rate of convergence of the learning process for representative domain adaptation. In particular, we analyze the factors that affect the asymptotical behavior of the learning process and discuss the choices of parameters in the situation of representative domain adaptation.
The numerical experiments also support our theoretical findings.
Meanwhile, we compare our results with the existing results of domain adaptation and the related results under the same-distribution assumption. Note that the representative domain adaption refers to a more general situation that covers both of domain adaptation with multiple sources and domain adaptation combining source and target. Thus, our results include many existing works as special cases. Additionally, our analysis can be applied to analyze the key quantities studied in \citet{Mansour09b,Ben-David10} [see Section \ref{sec:IPM}].
%%%

\subsection{Organization of the Paper}

The rest of this paper is organized as follows. Section \ref{sec:notation} introduces the problem studied in this paper. Section \ref{sec:IPM} introduces the integral probability metric and then gives a comparison with other quantities.
In Section \ref{sec:UEN}, we introduce the uniform entropy number and the Rademacher complexity.
Section \ref{sec:bound1} provides the generalization bounds for representative domain adaptation. In Section \ref{sec:AG}, we analyze the asymptotic behavior of the learning process for representative domain adaptation. Section \ref{sec:experiment} shows the numerical experiments supporting our theoretical findings. We brief the related works in Section \ref{sec:compare} and the last section concludes the paper.
In Appendix \ref{app:devi&symm}, we present the deviation inequalities and the symmetrization inequality, and all proofs  are given in Appendix \ref{app:proof}.

%%%%

\section{Problem Setup}\label{sec:notation}

We denote $\mathcal{Z}^{(S_k)}:=\mathcal{X}^{(S_k)}\times\mathcal{Y}^{(S_k)}
\subset\mathbb{R}^I\times\mathbb{R}^J$ $(1\leq k\leq K)$ and $\mathcal{Z}^{(T)}:=\mathcal{X}^{(T)}\times\mathcal{Y}^{(T)}
\subset\mathbb{R}^I\times\mathbb{R}^J$ as the $k$-th source domain and the target domain, respectively. Set $L=I+J$.
Let $\mathcal{D}^{(S_k)}$ and $\mathcal{D}^{(T)}$ stand for the distributions of the input spaces $\mathcal{X}^{(S_k)}$ $(1\leq k\leq K)$ and $\mathcal{X}^{(T)}$, respectively. Denote $g^{(S_k)}_*:\mathcal{X}^{(S_k)}\rightarrow\mathcal{Y}^{(S_k)}$ and $g^{(T)}_*:\mathcal{X}^{(T)}\rightarrow\mathcal{Y}^{(T)}$ as the labeling functions of $\mathcal{Z}^{(S_k)}$ ($1\leq k\leq K$) and $\mathcal{Z}^{(T)}$, respectively.

In representative domain adaptation, the input-space distributions $\mathcal{D}^{(S_k)}$ $(1\leq k\leq K)$ and $\mathcal{D}^{(T)}$ differ from each other, or $g^{(S_k)}_*$ $(1\leq k\leq K)$ and $g^{(T)}_*$ differ from each other, or both cases occur.
There are some (but not enough) samples $\overline{{\bf Z}}_{1}^{N_{T}}:=\{{\bf z}_n^{(T)}\}_{n=1}^{N_T}=\big\{({\bf x}_n^{(T)},{\bf y}_n^{(T)})\big\}_{n=1}^{N_T}$ drawn from the target domain $\mathcal{Z}^{(T)}$ in addition to a large amount of i.i.d. samples ${\bf Z}_{1}^{N_k}=\{{\bf z}^{(k)}_n\}_{n=1}^{N_k}=\big\{({\bf x}^{(k)}_n,{\bf y}^{(k)}_n)\big\}_{n=1}^{N_k}$ drawn from each source domain $\mathcal{Z}^{(S_k)}$ $(1\leq k\leq K)$ with $N^{(T)}\ll N_k$ for any $1\leq k\leq K$.

Given two parameters $\tau\in[0,1)$ and ${\bf w}\in[0,1]^{K}$ with $\sum_{k=1}^Kw_k=1$, denote the convex combination of the weighted empirical risk of multiple-source data and the empirical risk of the target data as:
\begin{equation}\label{eq:emrisk.MC}
\mathrm{E}_{{\bf w}}^{\tau}(\ell\circ g):=\tau\mathrm{E}^{(T)}_{N_T}(\ell\circ g)+(1-\tau)\mathrm{E}^{(S)}_{{\bf w}}(\ell\circ g),
\end{equation}
where $\ell$ is the loss function,
\begin{equation}\label{eq:emrisk.t}
   \mathrm{E}^{(T)}_{N_T}(\ell\circ g):=\frac{1}{N_T}\sum_{n=1}^{N_T}\ell(g({\bf x}^{(T)}_n),{\bf
    y}^{(T)}_n),
\end{equation}
and
\begin{align}\label{eq:emrisk.M}
   \mathrm{E}^{(S)}_{{\bf w}}(\ell\circ g):=&\sum_{k=1}^Kw_k\mathrm{E}_{N_k}^{(S_k)}(\ell\circ g)
   =\sum_{k=1}^K\frac{w_k}{N_k}\sum_{n=1}^{N_k}\ell(g({\bf x}^{(k)}_n),{\bf
    y}^{(k)}_n).
\end{align}
Given a function class $\mathcal{G}$, we denote $g^{\tau}_{\bf w}\in\mathcal{G}$ as the function that minimizes the empirical quantity $\mathrm{E}^{\tau}_{{\bf w}}(\ell\circ g)$ over $\mathcal{G}$
and it is expected that $g^{\tau}_{\bf w}$ will perform well
on the
target expected risk:
\begin{equation}\label{eq:exrisk.t}
    \mathrm{E}^{(T)}(\ell\circ g)=\int\ell\big(g({\bf x}^{(T)}),{\bf y}^{(T)}\big)d\mathrm{P}({\bf z}^{(T)}),\;\;g\in\mathcal{G},
\end{equation}
that is, $g^{\tau}_{\bf w}$ approximates the labeling function $g^{(T)}_*$ as precisely as possible.

Note that when $\tau=0$, such a learning process provides the domain adaptation with multiple sources \citep[see][]{Crammer08,Mansour09,zhang2012generalization}; setting $K=1$ provides the domain adaptation combining source and target data \citep[see][]{Ben-David10,Blitzer07,zhang2012generalization}; setting $\tau=0$ and $K=1$
provides the basic domain adaptation with one single source \citep[see][]{Ben-David06}.

In this learning process, we are mainly interested in the following two types of quantities:
\begin{itemize}
\item $\mathrm{E}^{(T)}(\ell\circ g^{\tau}_{\bf w})-\mathrm{E}_{{\bf w}}^{\tau}(\ell\circ g^{\tau}_{\bf w})$, which corresponds to the estimation of the expected risk in the target domain $\mathcal{Z}^{(T)}$ from the empirical quantity $\mathrm{E}^{(S)}_{{\bf w}}(\ell\circ g)$;
\item $\mathrm{E}^{(T)}(\ell\circ g^{\tau}_{\bf w})-\mathrm{E}^{(T)}(\ell\circ \widetilde{g}^{(T)}_*)$, which corresponds to the performance of the algorithm for domain adaptation it uses,
\end{itemize}
where
$\widetilde{g}^{(T)}_*\in\mathcal{G}$ is the function that minimizes the expected risk $\mathrm{E}^{(T)}(\ell\circ g)$ over $\mathcal{G}$.

Recalling \eqref{eq:emrisk.MC} and \eqref{eq:exrisk.t}, since
\begin{equation*}
   \mathrm{E}_{{\bf w}}^{\tau}(\ell\circ \widetilde{g}^{(T)}_*)-\mathrm{E}_{{\bf w}}^{\tau}(\ell\circ g_{{\bf w}}^{\tau})\geq 0,
\end{equation*}
we have
\begin{align*}
 \mathrm{E}^{(T)}(\ell\circ g_{{\bf w}}^{\tau})=&\mathrm{E}^{(T)}(\ell\circ g_{{\bf w}}^{\tau})-\mathrm{E}^{(T)}(\ell\circ \widetilde{g}^{(T)}_*)+\mathrm{E}^{(T)}(\ell\circ \widetilde{g}^{(T)}_*)\\
%%%%%%%%%
\leq&\mathrm{E}_{{\bf w}}^{\tau}(\ell\circ \widetilde{g}^{(T)}_*)-\mathrm{E}_{{\bf w}}^{\tau}(\ell\circ g_{{\bf w}}^{\tau})+\mathrm{E}^{(T)}(\ell\circ g_{{\bf w}}^{\tau})-\mathrm{E}^{(T)}(\ell\circ \widetilde{g}^{(T)}_*)+\mathrm{E}^{(T)}(\ell\circ \widetilde{g}^{(T)}_*)\\
%%%%%%%%%%%%
\leq& 2\sup_{g\in\mathcal{G}}\big|\mathrm{E}^{(T)}(\ell\circ g)-\mathrm{E}_{{\bf w}}^{\tau}(\ell\circ g)\big|  +\mathrm{E}^{(T)}(\ell\circ \widetilde{g}^{(T)}_*),
\end{align*}
and thus
\begin{align*}
   0 \leq& \mathrm{E}^{(T)}(\ell\circ g_{{\bf w}}^{\tau})-\mathrm{E}^{(T)}(\ell\circ \widetilde{g}^{(T)}_*)
   \leq 2\sup_{g\in\mathcal{G}}\big|\mathrm{E}^{(T)}(\ell\circ g)-\mathrm{E}_{{\bf w}}^{\tau}(\ell\circ g)\big|.
\end{align*}
This shows that the asymptotic behaviors of the aforementioned two quantities, when the sample numbers $N_1,\cdots,N_K$ (or part of them) go to {\it infinity}, can both be described by the supremum:
\begin{equation}\label{eq:upbound1}
\sup_{g\in\mathcal{G}}\big|\mathrm{E}^{(T)}(\ell\circ g)-\mathrm{E}_{{\bf w}}^{\tau}(\ell\circ g)\big|,
\end{equation}
which is the so-called generalization bound of the learning process for representative domain adaptation.

For convenience, we define the loss function
class
\begin{equation}\label{eq:fuclass}
    \mathcal{F}:=\{{\bf z} \mapsto \ell(g({\bf x}),{\bf y}):g\in \mathcal{G}\},
\end{equation}
and call $\mathcal{F}$ as the function class in the rest of this paper.
By \eqref{eq:emrisk.MC}, \eqref{eq:emrisk.t}, \eqref{eq:emrisk.M} and \eqref{eq:exrisk.t}, we briefly denote for any $f\in\mathcal{F}$,
\begin{equation}\label{eq:short1.MC}
    \mathrm{E}^{(T)}f:=\int f({\bf z}^{(T)})d\mathrm{P}({\bf z}^{(T)})\;,
\end{equation}
and
\begin{align}\label{eq:short2.MC}
   \mathrm{E}^\tau_{\bf w}f:=&\tau\mathrm{E}^{(T)}_{N_T}f+(1-\tau)\mathrm{E}^{(S)}_{{\bf w}}f
   =\tau\mathrm{E}^{(T)}_{N_T}f+
   (1-\tau)\sum_{k=1}^Kw_k\mathrm{E}_{N_k}^{(S_k)}f\nonumber\\
   =&\frac{\tau}{N_T}\sum_{n=1}^{N_T}f({\bf z}^{(T)}_n)+\sum_{k=1}^K\frac{w_k(1-\tau)}{N_k}\sum_{n=1}^{N_k}f({\bf z}^{(k)}_n).
\end{align}
Thus, we equivalently rewrite the generalization bound \eqref{eq:upbound1} as
\begin{equation*}%\label{eq:upbound2}
\sup_{f\in\mathcal{F}}\big|\mathrm{E}^{(T)}f-\mathrm{E}_{\bf w}^\tau f\big|.
\end{equation*}

%%%%%%%%%%%%

\section{Integral Probability Metric}\label{sec:IPM}

In the theoretical analysis of domain adaptation, one of main challenges is to find a quantity to measure the difference between the source domain $\mathcal{Z}^{(S)}$ and the target domain $\mathcal{Z}^{(T)}$, and then one can use the quantity to derive generalization bounds for domain adaptation \citep[see][]{Mansour08,Mansour09,Ben-David10,Ben-David06}.
Different from the existing works \citep[e.g.][]{Mansour08,Mansour09,Ben-David10,Ben-David06},
we use the integral probability metric to measure the difference between $\mathcal{Z}^{(S)}$ and $\mathcal{Z}^{(T)}$. We also discuss the relationship between the integral probability metric and other quantities proposed in existing works: the {\it $\mathcal{H}$-divergence} and the {\it discrepancy distance} \citep[see][]{Ben-David10,Mansour09b}.

\subsection{Integral Probability Metric}

%As addressed in Section \ref{sec:notation}, if two domains $\mathcal{Z}^{(S)}$ and $\mathcal{Z}^{(T)}$ are different, there are three possibilities: the input-space distributions $\mathcal{D}^{(S)}$ and $\mathcal{D}^{(T)}$ differ from each other, or the labeling functions $g^{(S_k)}_*$ $(1\leq k\leq K)$ and $g^{(T)}_*$ differ from each other, or both of the two cases occur.

\citet{Ben-David10,Ben-David06} introduced the {\it $\mathcal{H}$-divergence} to derive the generalization bounds based on the VC dimension under the condition of ``$\lambda$-close". \citet{Mansour09b} obtained the generalization bounds based on the Rademacher complexity by using the {\it discrepancy distance}. Both quantities are
aimed to measure the difference between two input-space distributions $\mathcal{D}^{(S)}$ and $\mathcal{D}^{(T)}$. Moreover, \citet{Mansour09} used the R\'enyi divergence to measure the distance between two distributions. In this paper, we use the following quantity to measure the difference between the distributions of the
source and the target domains:
\begin{definition}\label{def:distance}
Given two domains $\mathcal{Z}^{(S)},\mathcal{Z}^{(T)}\subset\mathbb{R}^L$, let ${\bf z}^{(S)}$ and ${\bf z}^{(T)}$ be the random variables taking values from $\mathcal{Z}^{(S)}$ and $\mathcal{Z}^{(T)}$, respectively. Let $\mathcal{F}\subset\mathbb{R}^{\mathcal{Z}}$ be a function class.
We define
\begin{equation}\label{eq:distance}
D_{\mathcal{F}}(S,T):=\sup_{f\in\mathcal{F}}|\mathrm{E}^{(S)}f-\mathrm{E}^{(T)}f|,
\end{equation}
where the expectations $\mathrm{E}^{(S)}$ and $\mathrm{E}^{(T)}$ are taken with respect to the distributions of the domains $\mathcal{Z}^{(S)}$ and $\mathcal{Z}^{(T)}$, respectively.
\end{definition}

The quantity $D_{\mathcal{F}}(S,T)$ is termed as the integral probability metric that plays an important role in probability theory for measuring the difference between two probability distributions \citep[see][]{Zolotarev84,Rachev,Muller97,Reid11}. Recently, \citet{Sriperumbudur12} gave a further investigation and proposed an empirical method to compute the integral probability metric. As mentioned by \citet{Muller97} [see page 432],
the quantity $D_{\mathcal{F}}(S,T)$ is a semimetric and it is a metric if and only if the function class $\mathcal{F}$ separates the set of all signed measures with $\mu(\mathcal{Z}) = 0$. Namely, according to Definition \ref{def:distance}, given a non-trivial function class $\mathcal{F}$, the quantity $D_{\mathcal{F}}(S,T)$ is equal to {\it zero} if the domains $\mathcal{Z}^{(S)}$ and $\mathcal{Z}^{(T)}$ have the same distribution.

By \eqref{eq:fuclass}, the quantity $D_{\mathcal{F}}(S,T)$ can be equivalently rewritten as
\begin{align}\label{eq:distance2}
D_{\mathcal{F}}(S,T)=&\sup_{g\in\mathcal{G}}\Big|\mathrm{E}^{(S)}\ell(g({\bf x}^{(S)}),{\bf y}^{(S)})-\mathrm{E}^{(T)}\ell(g({\bf x}^{(T)}),{\bf y}^{(T)})\Big|\nonumber\\
=&\sup_{g\in\mathcal{G}}\Big|\mathrm{E}^{(S)}\ell\big(g({\bf x}^{(S)}),g_*^{(S)}({\bf x}^{(S)})\big)-\mathrm{E}^{(T)}\ell\big(g({\bf x}^{(T)}),g_*^{(T)}({\bf x}^{(T)})\big)\Big|.
\end{align}
Next, based on the equivalent form \eqref{eq:distance2}, we discuss the relationship between the quantity $D_{\mathcal{F}}(S,T)$ and other quantities including the {\it $\mathcal{H}$-divergence} and the {\it discrepancy distance}.

\subsection{$\mathcal{H}$-Divergence and Discrepancy Distance}\label{sup:mechanism}

Before the formal discussion, we briefly introduce the related quantities proposed in the previous works of \citet{Ben-David10,Mansour09b}.
\subsubsection{$\mathcal{H}$-Divergence}
In classification tasks, by setting $\ell$ as the absolute-value loss function ($\ell(x,y)=|x-y|$), \citet{Ben-David10} introduced a variant of the {\it $\mathcal{H}$-divergence}:
\begin{align*}%\label{eq:H-div}
   d_{\mathcal{H}\triangle\mathcal{H}}(\mathcal{D}^{(S)},\mathcal{D}^{(T)})
=\sup_{g_1,g_2\in\mathcal{H}}\Big|\mathrm{E}^{(S)}\ell\big(g_1({\bf x}^{(S)}),g_2({\bf x}^{(S)})\big)-\mathrm{E}^{(T)}\ell\big(g_1({\bf x}^{(T)}),g_2({\bf x}^{(T)})\big)\Big|
\end{align*}
with the condition of ``$\lambda$-close": there exists a $\lambda>0$ such that
\begin{align}\label{eq:lambda}
    \lambda\geq
\inf_{g\in\mathcal{G}}\left\{\int\ell\big(g({\bf x}^{(S)}),g_*^{(S)}({\bf x}^{(S)})\big)d\mathrm{P}({\bf z}^{(S)})+\int\ell\big(g({\bf x}^{(T)}),g_*^{(T)}({\bf x}^{(T)})\big)d\mathrm{P}({\bf z}^{(T)})\right\}.
\end{align}

 One of the main results in \citet{Ben-David10} can be summarized as follows: when ${\bf w}=(1,0,\cdots,0)$ or $\tau=0$, \citet{Ben-David10} derived the VC-dimension-based upper bounds of
\begin{equation}\label{eq:quantity.1}
   \mathrm{E}^{(T)}\ell\big(g_{\bf w}^\tau({\bf x}^{(T)}),g^{(T)}_*({\bf x}^{(T)})\big)-\mathrm{E}^{(T)}\ell\big(\widetilde{g}^{(T)}_*({\bf x}^{(T)}),g^{(T)}_*({\bf x}^{(T)})\big)
\end{equation}
by using the summation of $d_{\mathcal{H}\triangle\mathcal{H}}
(\mathcal{D}^{(S_k)},\mathcal{D}^{(T)})+2\lambda_k$, where $\widetilde{g}^{(T)}_*$ minimizes the expected risk $\mathrm{E}^{(T)}(\ell\circ g)$ over $g\in\mathcal{G}$ \citep[see][Theorems 3 $\&$ 4]{Ben-David10}.

There are two points that should be noted:
\begin{itemize}
  \item as addressed in Section \ref{sec:notation}, the quantity \eqref{eq:quantity.1} can be bounded by the generalization bound \eqref{eq:upbound1} and thus the analysis presented in this paper can be applied to study \eqref{eq:quantity.1};
  \item recalling \eqref{eq:lambda}, the condition of ``$\lambda$-close" actually places a restriction among the function class $\mathcal{G}$ and the labeling functions $g_*^{(S)},g_*^{(T)}$.
In the optimistic case, both of $g_*^{(S)}$ and $g_*^{(T)}$ are contained by the function class $\mathcal{G}$ and are the same, then $\lambda=0$.
\end{itemize}

\subsubsection{Discrepancy Distance}
In both classification and regression tasks, given a function class $\mathcal{G}$ and a loss function $\ell$, \citet{Mansour09b} defined the {\it discrepancy distance} as
\begin{align}\label{eq:d-distance}
\mathrm{disc}_{\ell}(\mathcal{D}^{(S)},\mathcal{D}^{(T)})=\sup_{g_1,g_2\in\mathcal{G}}\Big|\mathrm{E}^{(S)}\ell\big(g_1({\bf x}^{(S)}),g_2({\bf x}^{(S)})\big)-\mathrm{E}^{(T)}\ell\big(g_1({\bf x}^{(T)}),g_2({\bf x}^{(T)})\big)\Big|,
\end{align}
and then used this quantity to obtain the generalization bounds based on the Rademacher complexity. As mentioned by \citet{Mansour09b}, the quantities $d_{\mathcal{H}\triangle\mathcal{H}}(\mathcal{D}^{(S)},\mathcal{D}^{(T)})$
and $\mathrm{disc}_{\ell}(\mathcal{D}^{(S)},\mathcal{D}^{(T)})$ match in the setting of classification tasks with $\ell$ being the absolute-value loss function, while the usage of $\mathrm{disc}_{\ell}(\mathcal{D}^{(S)},\mathcal{D}^{(T)})$ does not require the ``$\lambda$-close" condition. Instead, the authors achieved the upper bound of
\begin{equation*}%\label{eq:sum.05}
 \mathrm{E}^{(T)}\ell\big(g({\bf x}^{(T)}),g^{(T)}_*({\bf x}^{(T)})\big)-\mathrm{E}^{(T)}\ell\big(\widetilde{g}^{(T)}_*({\bf x}^{(T)}),g^{(T)}_*({\bf x}^{(T)})\big),\;\;\forall g\in\mathcal{G}
\end{equation*}
by using the summation
\begin{equation*}%\label{eq:sum}
   \mathrm{disc}_{\ell}(\mathcal{D}^{(S)},\mathcal{D}^{(T)})+ \frac{1}{N}\sum_{n=1}^{N_S}\ell\big(g({\bf x}_n^{(S)}),\widetilde{g}^{(S)}_*({\bf x}^{(S)})\big)+\mathrm{E}^{(S)}\ell\big(\widetilde{g}^{(S)}_*({\bf x}^{(S)}),\widetilde{g}^{(T)}_*({\bf x}^{(S)})\big),
\end{equation*}
where $\widetilde{g}^{(S)}_*$ (resp. $\widetilde{g}^{(T)}_*$) minimizes the expected risk $\mathrm{E}^{(S)}(\ell\circ g)$ (resp. $\mathrm{E}^{(T)}(\ell\circ g)$) over $g\in\mathcal{G}$. It can be equivalently rewritten as follows \citep[see][Theorems 8 $\&$ 9]{Mansour09b}: the upper bound
\begin{equation}\label{eq:sum.06}
 \mathrm{E}^{(T)}\ell\big(g({\bf x}^{(T)}),g^{(T)}_*({\bf x}^{(T)})\big)-\frac{1}{N}\sum_{n=1}^{N_S}\ell\big(g({\bf x}_n^{(S)}),\widetilde{g}^{(S)}_*({\bf x}^{(S)})\big) ,\;\;\forall g\in\mathcal{G}
\end{equation}
can be bounded by using the summation
\begin{equation}\label{eq:sum.07}
  \mathrm{disc}_{\ell}(\mathcal{D}^{(S)},\mathcal{D}^{(T)})+ \mathrm{E}^{(T)}\ell\big(\widetilde{g}^{(T)}_*({\bf x}^{(T)}),g^{(T)}_*({\bf x}^{(T)})\big) +\mathrm{E}^{(S)}\ell\big(\widetilde{g}^{(S)}_*({\bf x}^{(S)}),\widetilde{g}^{(T)}_*({\bf x}^{(S)})\big).
\end{equation}

There are also two points that should be noted:
\begin{itemize}
  \item as addressed above, the quantity \eqref{eq:sum.06} can be bounded by the generalization bound \eqref{eq:upbound1} and thus the analysis presented in this paper can also be applied to study \eqref{eq:sum.06};
  \item similar to the condition of ``$\lambda$-close" [see \eqref{eq:lambda}], the summation \eqref{eq:sum.07}, in some sense, describes the behaviors of the labeling functions $g_*^{(S)}$ and $g_*^{(T)}$, because the functions $\widetilde{g}^{(S)}_*$ and $\widetilde{g}^{(T)}_*$ can be regarded as the approximations of $g_*^{(S)}$ and $g_*^{(T)}$ respectively.
\end{itemize}

Next, we discuss the relationship between $D_{\mathcal{F}}(S,T)$ and the aforementioned two quantities: the $\mathcal{H}$-divergence and the discrepancy distance. Recalling Definition \ref{def:distance}, since there is no limitation on the function class $\mathcal{F}$, the integral probability metric
$D_{\mathcal{F}}(S,T)$ can be used in both classification and regression tasks.
Therefore, we only consider the relationship between the integral probability metric $D_{\mathcal{F}}(S,T)$ and the {\it discrepancy distance} $\mathrm{disc}_{\ell}(\mathcal{D}^{(S)},\mathcal{D}^{(T)})$.

\subsection{Relationship between $D_{\mathcal{F}}(S,T)$ and $\mathrm{disc}_{\ell}(\mathcal{D}^{(S)},\mathcal{D}^{(T)})$}
\label{subsec:relation}

From Definition \ref{def:distance} and \eqref{eq:distance2}, the integral probability metric $D_{\mathcal{F}}(S,T)$ measures the difference between the distributions of the two domains $\mathcal{Z}^{(S)}$ and $\mathcal{Z}^{(T)}$. However, as addressed in Section \ref{sec:notation}, if a domain $\mathcal{Z}^{(S)}$ differs from another domain $\mathcal{Z}^{(T)}$, there are three possibilities: the input-space distribution $\mathcal{D}^{(S)}$ differs from $\mathcal{D}^{(T)}$, or $g_*^{(S)}$ differs from $g_*^{(T)}$, or both of them occur. Therefore, it is necessary to consider two kinds of differences: the difference between the input-space distributions $\mathcal{D}^{(S)}$ and $\mathcal{D}^{(T)}$ and the difference between the labeling functions $g_*^{(S)}$ and $g_*^{(T)}$. Next, we will show that the integral probability metric $D_{\mathcal{F}}(S,T)$ can be bounded by using two separate quantities that can measure the difference between $\mathcal{D}^{(S)}$ and $\mathcal{D}^{(T)}$ and the difference between $g_*^{(S)}$ and $g_*^{(T)}$, respectively.

As shown in \eqref{eq:d-distance}, the quantity $\mathrm{disc}_{\ell}(\mathcal{D}^{(S)},\mathcal{D}^{(T)})$ actually measures the difference between the input-space distributions $\mathcal{D}^{(S)}$ and $\mathcal{D}^{(T)}$.
Moreover, we introduce another quantity to measure the difference between the labeling functions $g_*^{(S)}$ and $g_*^{(T)}$:
\begin{definition}\label{def:distance2}
Given a loss function $\ell$ and a function class $\mathcal{G}$, we define
\begin{equation}\label{eq:Q}
Q^{(T)}_{\mathcal{G}}(g_*^{(S)},g_*^{(T)})
:=\sup_{g_1\in\mathcal{G}}\Big|\mathrm{E}^{(T)}\ell\big(g_1({\bf x}^{(T)}),g_*^{(T)}({\bf x}^{(T)})\big)-\mathrm{E}^{(T)}\ell\big(g_1({\bf x}^{(T)}),g_*^{(S)}({\bf x}^{(T)})\big)\Big|.
\end{equation}
\end{definition}
Note that if both of the loss function $\ell$ and the function class $\mathcal{G}$ are  non-trivial (or $\mathcal{F}$ is non-trivial), the quantity $Q^{(T)}_{\mathcal{G}}(g_*^{(S)},g_*^{(T)})$ is a (semi)metric between the labeling functions $g_*^{(S)}$ and $g_*^{(T)}$. In fact, it is not hard to verify that $Q^{(T)}_{\mathcal{G}}(g_*^{(S)},g_*^{(T)})$ satisfies the triangle inequality and is equal to {\it zero} if $g_*^{(S)}$ and $g_*^{(T)}$ match.

By combining \eqref{eq:distance2}, \eqref{eq:d-distance} and \eqref{eq:Q}, we have
\begin{align*}%\label{eq:d1}
\mathrm{disc}_{\ell}(\mathcal{D}^{(S)},\mathcal{D}^{(T)})=&\sup_{g_1,g_2\in\mathcal{G}}\Big|\mathrm{E}^{(S)}\ell\big(g_1({\bf x}^{(S)}),g_2({\bf x}^{(S)})\big)-\mathrm{E}^{(T)}\ell\big(g_1({\bf x}^{(T)}),g_2({\bf x}^{(T)})\big)\Big|\nonumber\\
\geq&\sup_{g_1\in\mathcal{G}}\Big|\mathrm{E}^{(S)}\ell\big(g_1({\bf x}^{(S)}),g_*^{(S)}({\bf x}^{(S)})\big)-\mathrm{E}^{(T)}\ell\big(g_1({\bf x}^{(T)}),g_*^{(S)}({\bf x}^{(T)})\big)\Big|\nonumber\\
=&\sup_{g_1\in\mathcal{G}}\Big|\mathrm{E}^{(S)}\ell\big(g_1({\bf x}^{(S)}),g_*^{(S)}({\bf x}^{(S)})\big)-\mathrm{E}^{(T)}\ell\big(g_1({\bf x}^{(T)}),g_*^{(T)}({\bf x}^{(T)})\big)\nonumber\\
&+\mathrm{E}^{(T)}\ell\big(g_1({\bf x}^{(T)}),g_*^{(T)}({\bf x}^{(T)})\big)-\mathrm{E}^{(T)}\ell\big(g_1({\bf x}^{(T)}),g_*^{(S)}({\bf x}^{(T)})\big)\Big|\nonumber\\
\geq&\sup_{g_1\in\mathcal{G}}\Big|\mathrm{E}^{(S)}\ell\big(g_1({\bf x}^{(S)}),g_*^{(S)}({\bf x}^{(S)})\big)-\mathrm{E}^{(T)}\ell\big(g_1({\bf x}^{(T)}),g_*^{(T)}({\bf x}^{(T)})\big)\Big|\nonumber\\
&-\sup_{g_1\in\mathcal{G}}\Big|\mathrm{E}^{(T)}\ell\big(g_1({\bf x}^{(T)}),g_*^{(T)}({\bf x}^{(T)})\big)-\mathrm{E}^{(T)}\ell\big(g_1({\bf x}^{(T)}),g_*^{(S)}({\bf x}^{(T)})\big)\Big|\nonumber\\
=&D_{\mathcal{F}}(S,T)-Q^{(T)}_{\mathcal{G}}(g_*^{(S)},g_*^{(T)}),
\end{align*}
and thus
\begin{equation}\label{eq:d2}
    D_{\mathcal{F}}(S,T)\leq
\mathrm{disc}_{\ell}(\mathcal{D}^{(S)},\mathcal{D}^{(T)})
+Q^{(T)}_{\mathcal{G}}(g_*^{(S)},g_*^{(T)}),
\end{equation}
which implies that the integral probability metric $D_{\mathcal{F}}(S,T)$ can be bounded by the summation of the {\it discrepancy distance} $\mathrm{disc}_{\ell}(\mathcal{D}^{(S)},\mathcal{D}^{(T)})$ and the quantity $Q^{(T)}_{\mathcal{G}}(g_*^{(S)},g_*^{(T)})$, which measure the difference between the input-space distributions $\mathcal{D}^{(S)}$ and $\mathcal{D}^{(T)}$ and the difference between the labeling functions $g_*^{(S)}$ and $g_*^{(T)}$, respectively.

Compared with \eqref{eq:lambda} and \eqref{eq:sum.07}, the integral probability metric $D_{\mathcal{F}}(S,T)$ provides a new mechanism to capture the difference between two domains, where the difference between labeling functions $g_*^{(S)}$ and $g_*^{(T)}$ is measured by a (semi)metric $Q^{(T)}_{\mathcal{G}}(g_*^{(S)},g_*^{(T)})$.

\begin{remark}\label{rem:IPM}
 As shown in \eqref{eq:distance2} and \eqref{eq:d-distance}, the integral probability metric $D_\mathcal{F}(S,T)$ takes the supremum of $g$ over $\mathcal{G}$, and the discrepancy distance $\mathrm{disc}_{\ell}(\mathcal{D}^{(S)},\mathcal{D}^{(T)})$ takes the supremum of $g_1$ and $g_2$ over $\mathcal{G}$ simultaneously. Consider a specific domain adaptation situation: the labeling function $g_*^{(S)}$ is close to $g_*^{(T)}$ and meanwhile both of them are contained in the function class $\mathcal{G}$. In this case, $D_\mathcal{F}(S,T)$ can be very small even though $\mathrm{disc}_{\ell}(\mathcal{D}^{(S)},\mathcal{D}^{(T)})$ is large. Thus, the integral probability metric
is more suitable for such domain adaptation setting than the discrepancy distance.

%There is a specific case in the situation of domain adaptation: $\mathcal{D}^{(S)}$ differs from $\mathcal{D}^{(T)}$ and meanwhile $g_*^{(S)}$ differs from $g_*^{(T)}$, while the distribution of the domain $\mathcal{Z}^{(S)}$ matches with that of the domain $\mathcal{Z}^{(T)}$. In this case, the integral probability metric $D_\mathcal{F}(S,T)$ equals to {\it zero}, but neither $\mathrm{disc}_{\ell}(\mathcal{D}^{(S)},\mathcal{D}^{(T)})$ nor $Q^{(T)}_{\mathcal{G}}(g_*^{(S)},g_*^{(T)})$ equals to {\it zero} for any non-trivial loss function $\ell$ and function class $\mathcal{G}$. Therefore, the integral probability metric $D_\mathcal{F}(S,T)$ is more suitable to this case than the $\mathcal{H}$-divergence $d_{\mathcal{H}\triangle\mathcal{H}}(\mathcal{D}^{(S)},\mathcal{D}^{(T)})$ and the discrepancy distance $\mathrm{disc}_{\ell}(\mathcal{D}^{(S)},\mathcal{D}^{(T)})$.
\end{remark}

%%%%

\section{Uniform Entropy Number and Rademacher Complexity}\label{sec:UEN}

In this section, we introduce the definitions of the uniform entropy number and the Rademacher complexity, respectively.

%%%

\subsection{Uniform Entropy Number}

Generally, the generalization bound of a certain learning process is achieved by incorporating the complexity measure of function classes, {\it e.g.}, the covering number, the VC dimension and the Rademacher complexity.
The results of this paper are based on the uniform entropy number that is derived from the concept of the covering number and we refer to \citet{Mendelson03} for more details about the uniform entropy number. The covering number of a function class $\mathcal{F}$ is defined as follows:
\begin{definition}\label{def:CovNum}
Let $\mathcal{F}$ be a function class and $d$ be a metric on $\mathcal{F}$. For any $\xi>0$, the covering number of $\mathcal{F}$ at radius $\xi$
with respect to the metric $d$, denoted by $\mathcal{N}(\mathcal{F},\xi,d)$ is the minimum size of a cover of radius $\xi$.
\end{definition}
In some classical results of statistical learning theory, the covering number is applied by letting $d$ be the distribution-dependent metric. For example, as shown in Theorem 2.3 of \citet{Mendelson03}, one can set $d$ as the norm $\ell_1({\bf Z}_1^N)$ and then derives the generalization bound of the i.i.d. learning process by incorporating the expectation of the covering number, that is, $\mathrm{E}\mathcal{N}(\mathcal{F},\xi,\ell_1({\bf Z}_1^N))$.  However, in the situation of domain adaptation, we only know the information of source domain, while the expectation $\mathrm{E}\mathcal{N}(\mathcal{F},\xi,\ell_1({\bf Z}_1^N))$ is dependent on distributions of both source and target domains because ${\bf z}=({\bf x},{\bf y})$. Therefore, the covering number is no longer applicable to our scheme for obtaining the generalization bounds for representative domain adaptation.
In contrast, the uniform entropy number is distribution-free and thus we choose it as the complexity measure of function classes to derive the generalization bounds.

For clarity of presentation, we give some useful notations for the following discussion.
For any $1\leq k\leq K$, given a sample set ${\bf Z}_{1}^{N_k}:=\{{\bf z}_{n}^{(k)}\}_{n=1}^{N_k}$ drawn from the source domain $\mathcal{Z}^{(S_k)}$, we denote ${\bf Z'}_{1}^{N_k}:=\{{\bf z'}_{n}^{(k)}\}_{n=1}^{N_k}$ as the sample set drawn from $\mathcal{Z}^{(S_k)}$ such that the ghost sample ${\bf z'}^{(k)}_{n}$ has the same distribution as that of ${\bf z}^{(k)}_{n}$ for any $1\leq n\leq N_k$ and any $1\leq k\leq K$. Again, given a sample set ${\bf \overline{Z}}_{1}^{N_T}=\{{\bf z}_n^{(T)}\}_{n=1}^{N_T}$ drawn from the target domain $\mathcal{Z}^{(T)}$, let ${\bf \overline{Z'}}_{1}^{N_T}=\{{\bf z'}_n^{(T)}\}_{n=1}^{N_T}$ be the ghost sample set of ${\bf \overline{Z}}_{1}^{N_T}$.
Denote ${\bf \overline{Z}}_{1}^{2N_T}:=\{{\bf \overline{Z}}_{1}^{N_T},{\bf \overline{Z}'}_{1}^{N_T}\}$ and ${\bf Z}_{1}^{2N_k}:=\{{\bf Z}_{1}^{N_k},{\bf Z'}_{1}^{N_k}\}$ for any $1\leq k\leq K$, respectively.
Given any $\tau\in[0,1)$ and any ${\bf w}=(w_1,\cdots,w_K)\in[0,1]^K$ with $\sum_{k=1}^Kw_k=1$, we introduce a variant of the $\ell_1$ norm: for any $f\in\mathcal{F}$,
\begin{align*}%\label{eq:L1.MC}
 & \|f\|_{\ell^{{\bf w},\tau}_1(\{{\bf Z}_{1}^{2N_k}\}_{k=1}^K,{\bf \overline{Z}}_{1}^{2N_T})}
 :=\frac{\tau}{2N_T}\sum_{n=1}^{N_T}\big(|f({\bf z}_n^{(T)})|+|f({\bf z'}_n^{(T)})|\big)
  +
  \sum_{k=1}^K\frac{(1-\tau)w_k}{2N_k}\sum_{n=1}^{N_k}\left(|f({\bf z}_n^{(k)})|+|f({\bf z'}_n^{(k)})|\right).
  \end{align*}
It is noteworthy that the variant $\ell^{{\bf w},\tau}_1$ of the $\ell_1$ norm is still a norm on the functional space, which can be easily verified by using the definition of norm, so we omit it here. In the situation of representative domain adaptation, by setting the metric $d$ as $\ell^{{\bf w},\tau}_1(\{{\bf Z}_{1}^{2N_k}\}_{k=1}^K,{\bf \overline{Z}}_{1}^{2N_T})$, we then define the uniform entropy number of $\mathcal{F}$ with respect to the metric $\ell^{{\bf w},\tau}_1(\{{\bf Z}_{1}^{2N_k}\}_{k=1}^K,{\bf \overline{Z}}_{1}^{2N_T})$ as
\begin{align}\label{eq:UEN1.MC}
\ln\mathcal{N}_1^{{\bf w},\tau}\big(\mathcal{F},\xi,2{\bf N}\big):=\sup_{\{{\bf Z}_{1}^{2N_k}\}_{k=1}^K,{\bf \overline{Z}}_{1}^{2N_T}}\ln\mathcal{N}\left(\mathcal{F},\xi,\ell^{{\bf w},\tau}_1(\{{\bf Z}_{1}^{2N_k}\}_{k=1}^K,{\bf \overline{Z}}_{1}^{2N_T})\right)
\end{align}
with ${\bf N}=N_T+\sum_{k=1}^KN_k$.

%%%

\subsection{Rademacher Complexity}

The Rademacher complexity is one of the most frequently used complexity measures of function classes and we refer to \citet{Vaart96,Mendelson03} for details.

\begin{definition}\label{def:Rade}
Let $\mathcal{F}$ be a function class and $\{{\bf z}_n\}_{n=1}^N$ be a sample set drawn from $\mathcal{Z}$. Denote $\{\sigma_n\}_{n=1}^N$ be a set of random variables independently taking either value from $\{-1,1\}$ with equal probability. The Rademacher complexity of $\mathcal{F}$ is defined as
\begin{equation}\label{eq:ExRade}
\mathcal{R}(\mathcal{F}):=\mathrm{E}\sup_{f\in\mathcal{F}}
\left\{\frac{1}{N}\sum_{n=1}^N\sigma_nf({\bf z}_n)\right\}
\end{equation}
with its empirical version given by
\begin{equation*}%\label{eq:EmRade}
\mathcal{R}_N(\mathcal{F}):=\mathrm{E}_{\sigma}\sup_{f\in\mathcal{F}}
\left\{\frac{1}{N}\sum_{n=1}^N\sigma_nf({\bf z}_n)\right\},
\end{equation*}
where $\mathrm{E}$ stands for the expectation taken with respect to all random variables $\{{\bf z}_n\}_{n=1}^N$ and $\{\sigma_n\}_{n=1}^N$, and $\mathrm{E}_{\sigma}$ stands for the expectation only taken with respect to the random variables $\{\sigma_n\}_{n=1}^N$.
\end{definition}

\section{Generalization Bounds for Representative Domain Adaptation}\label{sec:bound1}

Based on the uniform entropy number defined in \eqref{eq:UEN1.MC}, we first present two types of the
generalization bounds for representative domain adaptation: Hoeffding-type and Bennett-type, which are derived from the Hoeffding-type deviation inequality and the Bennett-type deviation inequality respectively. Moreover, we obtain
the bounds based on the Rademacher complexity via the McDiarmid-type deviation inequality.

\subsection{Hoeffding-type Generalization Bounds}

 The following theorem presents the Hoeffding-type generalization bound for representative domain adaptation:
\begin{theorem}\label{thm:Hcrate.MC}
Assume that $\mathcal{F}$ is a function class consisting of the bounded functions with the range $[a,b]$. Let $\tau\in[0,1)$ and ${\bf w}=(w_1,\cdots,w_K)\in[0,1]^K$ with $\sum_{k=1}^Kw_k=1$.
Then, given any $\xi>(1-\tau)D^{({\bf w})}_{\mathcal{F}}(S,T)$,
 we have for any $N_T,N_1,\cdots,N_K\in\mathbb{N}$ such that
 \begin{equation*}
   \frac{\tau^2(b-a)^2}{N_T(\xi')^2}
+\sum_{k=1}^K\frac{(1-\tau)^2w_k^2(b-a)^2}{N_k(\xi')^2}\leq \frac{1}{8},
 \end{equation*}
  with probability at least $1-\epsilon$,
\begin{align}\label{eq:crate1.MC}
&\sup_{f\in\mathcal{F}}
\big|\mathrm{E}^{\tau}_{\bf w}f-\mathrm{E}^{(T)}f\big|
\leq (1-\tau)D^{({\bf w})}_{\mathcal{F}}(S,T)+ \left(\frac{\ln\mathcal{N}^{{\bf w},\tau}_1(\mathcal{F},\xi'/8,2{\bf N})-\ln(\epsilon/8)}
{\frac{1}{32(b-a)^2\big(\frac{\tau^2}{N_T}+
\sum_{k=1}^{K}\frac{(1-\tau)^2w_k^2}{N_k}\big)}}\right)^{\frac{1}{2}},
%\nonumber
\end{align}
where $\xi':=\xi-(1-\tau)D^{({\bf w})}_{\mathcal{F}}(S,T)$, ${\bf N}=N_T+\sum_{k=1}^KN_k$,
 \begin{align*}%\label{eq:epsilon}
    \epsilon:=&8\mathcal{N}^{{\bf w},\tau}_1(\mathcal{F},\xi'/8,2{\bf N}) \exp\left\{-\frac{(\xi')^2}
{32(b-a)^2\big(\frac{\tau^2}{N_T}+
\sum_{k=1}^{K}\frac{(1-\tau)^2w_k^2}{N_k}\big)}\right\},
\end{align*}
and
\begin{equation}\label{eq:Dist.MC}
D^{({\bf w})}_{\mathcal{F}}(S,T):=\sum_{k=1}^Kw_kD_{\mathcal{F}}(S_k,T).
\end{equation}
%$D^{({\bf w})}_{\mathcal{F}}(S,T)$ is defined in \eqref{eq:Dist.MC}.
\end{theorem}

In the above theorem, we present the generalization bound derived from the Hoeffding-type deviation inequality. As shown in the theorem, the generalization bound $\sup_{f\in\mathcal{F}}|\mathrm{E}^{(T)}f-\mathrm{E}^{\tau}_{\bf w}f|$ can be bounded by the right-hand side of \eqref{eq:crate1.MC}.
Compared to the classical result under the same-distribution assumption \citep[see][Theorem 2.3 and Definition 2.5]{Mendelson03}: with probability at least $1-\epsilon$,
\begin{align}\label{eq:crate2.1}
    \sup_{f\in\mathcal{F}}
\big|\mathrm{E}_Nf-\mathrm{E}f\big|
\leq O\left(\sqrt{\frac{\ln\mathcal{N}_1\big(\mathcal{F},\xi,N\big)
-\ln(\epsilon/8)}{N}}\right)
\end{align}
with $\mathrm{E}_Nf$ being the empirical risk with respect to the  sample set ${\bf Z}_1^N$, there is a discrepancy quantity $(1-\tau)D^{({\bf w})}_{\mathcal{F}}(S,T)$ that is determined by three factors: the choice of ${\bf w}$, the choice of $\tau$ and the quantities $D_{\mathcal{F}}(S_k,T)$ ($1\leq k\leq K$). The two results will coincide if any source domain and the target domain match, that is, $D_{\mathcal{F}}(S_k,T)=0$ holds for any $1\leq k\leq K$.

\subsection{Bennett-type Generalization Bounds}

The above result is derived from the Hoeffding-type deviation inequality that only incorporates the information of the expectation. Recalling the classical Bennett's inequality \citep{Bennett62}, the Bennett-type inequalities are based on the information of the expectation and the variance (also see Appendix \ref{app:devi&symm}). Therefore, the Bennett-type results intuitively should provide a faster rate of convergence than that of the Hoeffding-type results. The following theorem presents the Bennett-type generalization bound for representative domain adaptation.

%We refer to \citet{zhang2013bennett} for the details of the Bennett-type generalization bounds of the i.i.d. learning process.

\begin{theorem}\label{thm:RB1.MC}
Under the notations of Theorem \ref{thm:Hcrate.MC}, set $w_k=N_k/\sum_{k=1}^KN_k$ ($1\leq k\leq K$) and $\tau=N_T/(N_T+\sum_{k=1}^KN_k)$.
Then, given any $\xi>(1-\tau)D^{({\bf w})}_{\mathcal{F}}(S,T)$,
 we have for any $N_T,N_1,\cdots,N_K\in\mathbb{N}$ such that
 \begin{equation*}
  N_T+\sum_{k=1}^K N_k\geq \frac{(b-a)^2}{8(\xi')^2}
 \end{equation*}
with $\xi'=\xi-(1-\tau)D^{({\bf w})}_{\mathcal{F}}(S,T)$,
 \begin{align}\label{eq:RB1.MC}
    \mathrm{Pr}\left\{\sup_{f\in\mathcal{F}}
\big|\mathrm{E}^{\tau}_{\bf w}f-\mathrm{E}^{(T)}f\big|>\xi\right\}
        \leq8\mathcal{N}^{{\bf w},\tau}_1(\mathcal{F},\xi'/8,2{\bf N})\exp\left\{\Big(N_T+\sum_{k=1}^KN_k\Big)\Gamma\left( \frac{\xi'}{(b-a)}\right)\right\},
\end{align}
where $\Gamma(x):=x-(x+1)\ln(x+1)$.
%\begin{align}\label{eq:Gamma}
%\Gamma(x):=x-(x+1)\ln(x+1).
%\end{align}
\end{theorem}

In the above theorem, we show that the probability that the generalization bound $\sup_{f\in\mathcal{F}}
\big|\mathrm{E}^{\tau}_{\bf w}f-\mathrm{E}^{(T)}f\big|$ is larger than a certain number $\xi>(1-\tau)D^{({\bf w})}_{\mathcal{F}}(S,T)$ can be bounded by the right-hand side of \eqref{eq:RB1.MC}. Compared with the Hoeffding-type result \eqref{eq:crate1.MC}, there are two limitations in this result:
\begin{itemize}
\item this generalization bound is actually the minimum value with respect to ${\bf w}$ and $\tau$, and does not reflect how the two parameters affect the bound. The result presented in the above theorem is not completely satisfactory  because it is hard to obtain the analytical expression of the inverse function of $A(\mathrm{e}^{ax}-1)+B(\mathrm{e}^{bx}-1)$ for any non-trivial $A,B,a,b>0$ (see Proof of Theorem \ref{thm:dineq.MC});
\item since it is also hard to obtain the analytical expression of the inverse function of $\Gamma(x)=x-(x+1)\ln(x+1)$, the result \eqref{eq:RB1.MC} cannot directly lead to the upper bound of $\sup_{f\in\mathcal{F}}
\big|\mathrm{E}^{\tau}_{\bf w}f-\mathrm{E}^{(T)}f\big|$, while the Hoeffding-type result \eqref{eq:crate1.MC} does. Instead, one generally uses $\frac{-x^2}{2+(2x/3)}$ to approximate the function $\Gamma(x)$, which leads to Bernstein-type alternative expression of the bound \eqref{eq:RB1.MC}:
%\footnote{http://ocw.mit.edu/courses/mathematics/18-465-topics-in-statistics-statistical-learning-theory-spring-2007/lecture-notes/l6.pdf}
\begin{align}\label{eq:Btype}
    \sup_{f\in\mathcal{F}}
\big|\mathrm{E}_Nf-\mathrm{E}f\big|\leq &(1-\tau)D^{({\bf w})}_{\mathcal{F}}(S,T) +\frac{
4(b-a)\big(\ln\mathcal{N}^{{\bf w},\tau}_1(\mathcal{F},\xi'/8,2{\bf N})-\ln(\epsilon/8)\big)}{3(N_T+\sum_{k=1}^KN_k)}
\nonumber\\
&+\frac{(b-a)\sqrt{2\big(\ln\mathcal{N}^{{\bf w},\tau}_1(\mathcal{F},\xi'/8,2{\bf N})-\ln(\epsilon/8)\big)}}{\sqrt{N_T+\sum_{k=1}^KN_k}}.
\end{align}
\end{itemize}

Compared to the Hoeffding-type result \eqref{eq:crate1.MC}, the alternative expression \eqref{eq:Btype} implies that the Bennett-type bound \eqref{eq:RB1.MC} does not provide stronger bounds for representative domain adaptation. First, the bound \eqref{eq:Btype} does not reflect how the parameters ${\bf w}$ and $\tau$ affect the performance of representative domain adaptation. Second, according to the Bernstein-type alternative expression \eqref{eq:Btype}, its rate of convergence is the same as that of the Hoeffding-type result \eqref{eq:crate1.MC}.

Next, we present a new alternative expression of \eqref{eq:RB1.MC}, which shows that the Bennett-type results can provide a faster rate of convergence than the Hoeffding-type bounds in addition to a more detailed description of the asymptotical behavior of the learning process.

\subsection{Alternative Expression of Bennett-type Generalization Bound}

Different from the Bernstein-type result \eqref{eq:Btype}, we introduce a new technique to deal with the term $\Gamma(x)$ and the details of the technique are referred to \citet{zhang2013bennett}.
Consider a function
\begin{equation}\label{eq:gamma}
\eta(c_1;x):=\frac{\ln\big(((x+1)
\ln(x+1)-x)/c_1\big)}{\ln(x)},
\end{equation}
and there holds that $\Gamma(x)=-c_1 x^{\eta(c_1,x)}\leq
-c_1 x^{\,\widetilde{\eta}}<-c_1 x^2$ for any
$0<\eta(c_1;x)\leq\widetilde{\eta}<2$ with $x\in(0,1/8]$ and $c_1\in(0.0075,0.4804)$. By replacing $\Gamma(x)$ with $-c_1 x^{\eta}$, we then obtain another alternative expression of the Bennett-type bound \eqref{eq:RB1.MC} as follows:

\begin{theorem}\label{thm:crate2.MC}
Under the notations of Theorem \ref{thm:Hcrate.MC}, set $w_k=N_k/\sum_{k=1}^KN_k$ ($1\leq k\leq K$) and $\tau=N_T/(N_T+\sum_{k=1}^KN_k)$.
Then, given any $\xi>(1-\tau)D_{\mathcal{F}}^{({\bf w})}(S,T)$,
 we have for any $N_T,N_1,\cdots,N_K\in\mathbb{N}$ such that
 \begin{equation*}
    N_T+\sum_{k=1}^K N_k\geq \frac{(b-a)^2}{8(\xi')^2}
 \end{equation*}
 with probability at least $1-\epsilon$,
\begin{align}\label{eq:crate2.MC}
&\sup_{f\in\mathcal{F}}
\big|\mathrm{E}^{\tau}_{\bf w}f-\mathrm{E}^{(T)}f\big|
\leq (1-\tau)D^{({\bf w})}_{\mathcal{F}}(S,T)+ (b-a)\left(\frac{\ln\mathcal{N}^{{\bf w},\tau}_1(\mathcal{F},\frac{\xi'}{8},2{\bf N})-\ln(\frac{\epsilon}{8})}
{N_T+\sum_{k=1}^KN_k}\right)^{\frac{1}{\eta}},
\end{align}
where $\xi':=\xi-(1-\tau)D^{({\bf w})}_{\mathcal{F}}(S,T)$,
$\epsilon:=8\mathcal{N}^{{\bf w},\tau}_1(\mathcal{F},\xi'/8,2{\bf N})\mathrm{e}^{(N_T+\sum_{k=1}^KN_k)\Gamma(x)}$ with $x\in(0,1/8]$ and
 $0<\eta(c_1;x)\leq\eta<2$.
\end{theorem}

This result shows that the Bennett-type bounds have a faster rate $o(N^{-\frac{1}{2}})$ of convergence than $O(N^{-\frac{1}{2}})$ of the Hoeffding-type results. Moreover, we can observe from the  numerical simulation
that the rate $O(N^{-\frac{1}{\eta}})$ varies w.r.t. $x$ for any $c_1\in(0.0075,0.4804)$ and especially, for any $c_1\in(0.0075,0.4434]$, the function $\eta(c_1;x)$ is monotonically decreasing in the interval $x\in(0,1/8]$, which implies that the rate will become faster as the discrepancy between the expected risk and the empirical quantity becomes bigger when $c_1\in(0.0075,0.4434]$. In contrast, the Hoeffding-type results have a consistent rate $O(N^{-\frac{1}{2}})$ regardless of the discrepancy. Therefore, although the Bennett-type bounds \eqref{eq:RB1.MC} and \eqref{eq:crate2.MC} do not reflect how the parameters ${\bf w}$ and $\tau$ affect the performance of the representative domain adaptation, they provide a more detailed
description of the asymptotical behavior of the learning process for representative domain adaptation.

\subsection{Generalization Bounds Based on Rademacher Complexity}

Based on the Rademacher complexity, we obtain the following generalization bounds for representative domain adaptation. Its proof is given in Appendix \ref{app:proof}.

\begin{theorem}\label{thm:RB.Rade.MC}
Assume that $\mathcal{F}$ is a function class consisting of bounded functions with the range $[a,b]$. Then, given any $\tau\in[0,1)$ and any ${\bf w}=(w_1,\cdots,w_K)\in[0,1]^K$ with $\sum_{k=1}^Kw_k=1$, we have with probability at least $1-\epsilon$,
\begin{align}\label{eq:RB.Rade.MC}
\sup_{f\in\mathcal{F}}
\big|\mathrm{E}^{\tau}_{\bf w}f-\mathrm{E}^{(T)}f\big|\leq& (1-\tau)D^{({\bf w})}_{\mathcal{F}}(S,T)+2(1-\tau)\sum_{k=1}^Kw_k\mathcal{R}^{(k)}(\mathcal{F})
+2\tau\mathcal{R}^{(T)}_{N_T}(\mathcal{F})\\
&+
2
\tau\sqrt{\frac{(b-a)^2\ln(4/\epsilon)}{2N_T}}+
\sqrt{\frac{(b-a)^2\ln(2/\epsilon)}{2}
\left(\frac{\tau^2}{N_T}+\sum_{k=1}^K\frac{(1-\tau)^2w_k^2}{N_k}\right)},\nonumber
%\nonumber\\
%\leq& (1-\tau)D^{({\bf w})}_{\mathcal{F}}(S,T)+2(1-\tau)\sum_{k=1}^Kw_k\mathcal{R}_{N_k}^{(k)}
%(\mathcal{F})
%+2\tau\mathcal{R}^{(T)}_{N_T}(\mathcal{F})\nonumber\\
%&+
%3
%\tau\sqrt{\frac{(b-a)^2\ln(4/\epsilon)}{2N_T}}+
%3\sqrt{\frac{(b-a)^2\ln(4/\epsilon)}{2}
%\left(\frac{\tau^2}{N_T}+\sum_{k=1}^K\frac{(1-\tau)^2w_k^2}{N_k}\right)},\nonumber
\end{align}
where $D^{({\bf w})}_{\mathcal{F}}(S,T)$ is defined in \eqref{eq:Dist.MC}, $\mathcal{R}_{N_T}^{(T)}$ is the empirical Rademacher complexity on the target domain $\mathcal{Z}^{(T)}$, and $\mathcal{R}^{(k)}(\mathcal{F})$ ($1\leq k\leq K$) are the Rademacher complexities on the source domains $\mathcal{Z}^{(S_k)}$.
%
%
% and $\mathcal{R}^{(k)}(\mathcal{F})$ (resp. $\mathcal{R}_{N_T}^{(T)}
%(\mathcal{F})$) ($1\leq k\leq K$) are the Rademacher complexities (resp. empirical Rademacher complexities) on the source domains $\mathcal{Z}^{(S_k)}$ (resp. target domain $\mathcal{Z}^{(T)}$), respectively.
\end{theorem}
Note that in the derived bound \eqref{eq:RB.Rade.MC}, we adopt an empirical Rademacher complexity $\mathcal{R}^{(T)}_{N_T}(\mathcal{F})$ that is based on the data drawn from the target domain $\mathcal{Z}^{(T)}$, because the distribution of $\mathcal{Z}^{(T)}$ is unknown in the situation of domain adaptation.
Similarly, the derived bound \eqref{eq:RB.Rade.MC} coincides with the related classical result under the assumption of same distribution \citep[see][Theorem 5]{Bousquet04}, when any source domain of $\{\mathcal{Z}^{(S_k)}\}_{k=1}^K$ and the target domain $\mathcal{Z}^{(T)}$ match, that is, $D^{({\bf w})}_{\mathcal{F}}(S,T)=D_{\mathcal{F}}(S_k,T)=0$ holds for any $1\leq k\leq K$.

Similar to the result \eqref{eq:crate2.MC}, we adopt the technique mentioned in \citet{zhang2013bennett} again and replace the term $\Gamma(x)$ with $-c x^{\eta}$ in the derived Bennett-type deviation inequality \eqref{eq:McDiarmid.Bennett} (see Appendix \ref{app:devi&symm}). Then, we obtain the Bennett-type generalization bounds based on the Rademacher complexity as follows:

\begin{theorem}\label{thm:RadeBound}
Under notations in Theorem \ref{thm:RB.Rade.MC}, we
have with probability at least $1-\epsilon$,
\begin{align}\label{eq:RadeBound}
\sup_{f\in\mathcal{F}}
\big|\mathrm{E}^{\tau}_{\bf w}f-\mathrm{E}^{(T)}f\big|\leq& (1-\tau)D^{({\bf w})}_{\mathcal{F}}(S,T)+2(1-\tau)\sum_{k=1}^Kw_k\mathcal{R}^{(k)}
(\mathcal{F})
+2\tau\mathcal{R}^{(T)}_{N_T}(\mathcal{F})\\
&+(b-a)\left(
2
\tau\sqrt[\eta]{\frac{\ln(4/\epsilon)}{c_2N_T}}+
\sqrt[\eta]{\frac{\ln(2/\epsilon)}{c_2}
\left(\frac{\tau^2}{N_T}
+\sum_{k=1}^K\frac{(1-\tau)^2w_k^2}{N_k}\right)}\right),\nonumber
%\nonumber\\
%\leq& (1-\tau)D^{({\bf w})}_{\mathcal{F}}(S,T)+2(1-\tau)\sum_{k=1}^Kw_k\mathcal{R}_{N_k}^{(k)}
%(\mathcal{F})
%+2\tau\mathcal{R}^{(T)}_{N_T}(\mathcal{F})\nonumber\\
%&+(b-a)\left(
%3
%\tau\sqrt[\eta]{\frac{\ln(4/\epsilon)}{c_2N_T}}+
%3\sqrt[\eta]{\frac{\ln(4/\epsilon)}{c_2}
%\left(\frac{\tau^2}{N_T}+\sum_{k=1}^K
%\frac{(1-\tau)^2w_k^2}{N_k}\right)}\right),\nonumber
\end{align}
where $c_2$ is taken from the interval $(0.0075,0.3863)$, $\epsilon:=\exp\big\{N\Gamma(x)\big\}$ and
 $0<\eta\big(c_2;x\big)\leq\eta<2$ $(x\in(0,1])$ with $\eta(c_2;x)$ defined in \eqref{eq:gamma}.
\end{theorem}
The results in the above theorem match with the Bennett-type bounds of the i.i.d. learning process shown in Theorem 4.3 of \citet{zhang2013bennett}, when any source domain of $\{\mathcal{Z}^{(S_k)}\}_{k=1}^K$ and the target domain $\mathcal{Z}^{(T)}$ match, that is, $D^{({\bf w})}_{\mathcal{F}}(S,T)=D_{\mathcal{F}}(S_k,T)=0$ holds for any $1\leq k\leq K$. The proof of this theorem is similar to that of Theorem \ref{thm:RB.Rade.MC}, so we omit it.

In addition, it is noteworthy that the Hoeffding-type results \eqref{eq:crate1.MC} and \eqref{eq:RB.Rade.MC} exhibit a tradeoff between the sample numbers $N_k$ ($1\leq k\leq K$) and $N_T$, which is associated with the choice of $\tau$. Although such a tradeoff has been discussed in some previous works \citep{Blitzer07,Ben-David10,zhang2012generalization}, the next section will show a rigorous theoretical analysis of the tradeoff in the situation of representative domain adaptation.

\begin{remark}\label{rem:result}
We have shown that $D_{\mathcal{F}}(S,T)$ can be bounded by the summation of the {\it discrepancy distance} $\mathrm{disc}_{\ell}(\mathcal{D}^{(S)},\mathcal{D}^{(T)})$ and the quantity $Q^{(T)}_{\mathcal{G}}(g_*^{(S)},g_*^{(T)})$, which measure the difference between distributions $\mathcal{D}^{(S)}$ and $\mathcal{D}^{(T)}$ and the difference between labeling functions $g_*^{(S)}$ and $g_*^{(T)}$, respectively. Thus, the presented generalization results \eqref{eq:crate1.MC}, \eqref{eq:RB1.MC}, \eqref{eq:Btype}, \eqref{eq:crate2.MC}, \eqref{eq:RB.Rade.MC} and \eqref{eq:RadeBound} can also be achieved by using the {\it discrepancy distance} (or $\mathcal{H}$-divergency) and the quantity $Q^{(T)}_{\mathcal{G}}(g_*^{(S)},g_*^{(T)})$ \citep[see][]{Ben-David10,Mansour09b}. In fact, one can directly replace $D_{\mathcal{F}}(S,T)$ with $\mathrm{disc}_{\ell}(\mathcal{D}^{(S)},\mathcal{D}^{(T)})
+Q^{(T)}_{\mathcal{G}}(g_*^{(S)},g_*^{(T)})$ and the derived results are similar to Theorem 9 of \citet{Mansour09b}. Alternatively, under the condition of ``$\lambda$-close" in classification setting, one can also replace $D_{\mathcal{F}}(S,T)$ with $ d_{\mathcal{H}\triangle\mathcal{H}}(\mathcal{D}^{(S)},\mathcal{D}^{(T)})
+c\lambda$ ($c>0$), and the derived bounds are similar to the results given by \citet{Ben-David10}. Thus, our results include previous works as special cases.
\end{remark}

%%%
\section{Asymptotic Behavior for Representative Domain Adaptation}\label{sec:AG}

In this section, we discuss the asymptotical convergence and the rate of convergence of the learning process for representative domain adaptation. We also give a comparison with the related results under the same-distribution assumption and the existing results for domain adaptation.

\subsection{Asymptotic Convergence}

From Theorem \ref{thm:Hcrate.MC}, the asymptotic convergence of the learning process for representative domain adaptation is affected by three factors: the uniform entropy number $\ln\mathcal{N}^{{\bf w},\tau}_1(\mathcal{F},\xi'/8,2{\bf N})$, the discrepancy term $D^{({\bf w})}_{\mathcal{F}}(S,T)$ and the choices of ${\bf w}, \tau$.

%%%
\begin{theorem}\label{thm:converge.MC}
Assume that $\mathcal{F}$ is a function class consisting of bounded functions with the range $[a,b]$. Given any $\tau\in[0,1)$ and any ${\bf w}=(w_1,\cdots,w_K)\in[0,1]^K$ with $\sum_{k=1}^Kw_k=1$, if the following condition holds: for any $1\leq k\leq K$ such that $w_k\not=0$,
\begin{equation}\label{eq:cond.MC}
  \lim_{N_k\rightarrow +\infty}\frac{\ln\mathcal{N}^{{\bf w},\tau}_1(\mathcal{F},\xi'/8,2{\bf N})}
{\frac{1}{\big(\frac{\tau^2}{N_T}+
\sum_{k=1}^{K}\frac{(1-\tau)^2w_k^2}{N_k}\big)}}<+\infty
\end{equation}
with ${\bf N}=N_T+\sum_{k=1}^KN_k$ and $\xi':=\xi-(1-\tau)D_{\mathcal{F}}(S,T)$,
then we have for any $\xi>(1-\tau)D^{({\bf w})}_{\mathcal{F}}(S,T)$,
\begin{equation}\label{eq:converge.MC}
    \lim_{N_k\rightarrow +\infty}\mathrm{Pr}
\left\{\sup_{f\in\mathcal{F}}
\big|\mathrm{E}^{\tau}_{\bf w}f-\mathrm{E}^{(T)}f\big|>\xi\right\}=0.
\end{equation}
\end{theorem}
As shown in Theorem \ref{thm:converge.MC}, if the choices of ${\bf w},\tau$ and the uniform entropy number $\ln\mathcal{N}^{{\bf w},\tau}_1(\mathcal{F},\xi'/8,2{\bf N})$ satisfy the condition \eqref{eq:cond.MC} with $\sum_{k=1}^Kw_k=1$,
the probability of the event $\sup_{f\in\mathcal{F}}
\big|\mathrm{E}^{\tau}_{\bf w}f-\mathrm{E}^{(T)}f\big|>\xi$ will converge to {\it zero} for any $\xi>(1-\tau)D^{({\bf w})}_{\mathcal{F}}(S,T)$, when the sample numbers $N_1,\cdots,N_K$ (or a part of them) go to {\it infinity}, respectively. This is partially in accordance with the classical result of the asymptotic convergence
of the learning process under the same-distribution assumption \citep[see][Theorem 2.3 and Definition 2.5]{Mendelson03}: the probability of the event that $\sup_{f\in\mathcal{F}}\big|\mathrm{E}f-\mathrm{E}_Nf\big|>\xi$ will converge to {\it zero} for any $\xi>0$, if the uniform entropy number $\ln\mathcal{N}_1\left(\mathcal{F},\xi,N\right)$ satisfies the following:
\begin{equation}\label{eq:cond2.MC}
  \lim_{N\rightarrow +\infty}\frac{\ln\mathcal{N}_1
\left(\mathcal{F},\xi,N\right)}{N}<+\infty.
\end{equation}

Note that in the learning process for representative domain adaptation, the uniform convergence of the empirical risk $\mathrm{E}^{\tau}_{\bf w}f$ to the expected risk $\mathrm{E}^{(T)}f$ may not hold, because the limit \eqref{eq:converge.MC} does not hold for any $\xi>0$ but for any $\xi>(1-\tau)D^{({\bf w})}_{\mathcal{F}}(S,T)$. By contrast, the limit \eqref{eq:converge.MC} holds for all $\xi>0$ in the learning process under the same-distribution assumption, if the condition \eqref{eq:cond2.MC} is satisfied. The two results coincide when any source domain $\mathcal{Z}^{(S_k)}$ ($1\leq k\leq K$) and the
target domain $\mathcal{Z}^{(T)}$ match, that is, $D^{({\bf w})}_{\mathcal{F}}(S,T)=D_{\mathcal{F}}(S_k,T)=0$ holds for any $1\leq k\leq K$.

Especially, if we set $w_k=N_k/\sum_{k=1}^KN_k$ ($1\leq k\leq K$) and $\tau=N_T/(N_T+\sum_{k=1}^KN_k)$, the result shown in Theorem \ref{thm:converge.MC} can also be derived from the generalization bound \eqref{eq:RB1.MC} that is of Bennett-type, because the function $\Gamma(x)$ is monotonically decreasing and smaller than ${\it zero}$ when $x>0$ (see Theorem \ref{thm:RB1.MC}).

\subsection{Rate of Convergence}
From \eqref{eq:crate1.MC}, the rate of convergence is affected by the choices of ${\bf w}$ and $\tau$.
According to the Cauchy-Schwarz inequality, setting $w_k=N_k/\sum_{k=1}^KN_k$ ($1\leq k\leq K$) and $\tau=N_T/(N_T+\sum_{k=1}^KN_k)$
%\begin{equation*}%\label{eq:rate.max}
 %\max\left\{\frac{1}{\big(\frac{\tau^2}{N_T}+
%\sum_{k=1}^{K}\frac{(1-\tau)^2w_k^2}{N_k}\big)}\right\}
%=N_T+\sum_{k=1}^KN_k,
%\end{equation*}
minimizes the second term of the right-hand side of \eqref{eq:crate1.MC} leading to a Hoeffding-type result:
\begin{align}\label{eq:Hcrate2.MC}
    &\sup_{f\in\mathcal{F}}
\big|\mathrm{E}_\tau f-\mathrm{E}^{(T)}f\big|
\leq\frac{\sum_{k=1}^KN_kD^{({\bf w})}_{\mathcal{F}}(S,T)}{N_T+\sum_{k=1}^KN_k}+ \left(\frac{\ln\mathcal{N}^{{\bf w},\tau}_1(\mathcal{F},\xi'/8,2{\bf N})-\ln(\epsilon/8)}
{\frac{N_T+\sum_{k=1}^KN_k}{32(b-a)^2}}
\right)^{\frac{1}{2}}.
\end{align}
This result implies that the fastest rate of convergence for the representative domain adaptation is up to $O(N^{-\frac{1}{2}})$ which is the same as the classical result \eqref{eq:crate2.1} of the learning process under the same-distribution assumption, if the discrepancy term $D^{({\bf w})}_{\mathcal{F}}(S,T)=0$.

On the other hand, the choice of $\tau$ is not only one of essential factors to the rate of convergence but also is associated with the tradeoff between the sample numbers $\{N_k\}_{k=1}^K$ and $N_T$.
As shown in \eqref{eq:Hcrate2.MC}, provided that the value of $\ln\mathcal{N}^{{\bf w},\tau}_1(\mathcal{F},\xi'/8,2{\bf N})$ is fixed, we can find that setting $\tau=\frac{N_T}{N_T+\sum_{k=1}^KN_k}$ can result in the fastest rate of convergence, while it can also cause the relatively larger discrepancy between the empirical risk $\mathrm{E}^{\tau}_{\bf w}f$ and the expected risk $\mathrm{E}^{(T)}f$, because the situation of representative domain adaptation is set up under the condition that $N_T\ll N_k$ for any $1\leq k\leq K$, which implies that $\frac{\sum_{k=1}^KN_k}{N_T+\sum_{k=1}^KN_k}\approx1$.

From Theorem \ref{thm:RB1.MC}, such a setting of ${\bf w}$ and $\tau$ leads to the Bennett-type result \eqref{eq:RB1.MC} as well.
% that can provide a faster rate $o\big(\frac{1}{N^{1/1.29}}\big)$ of convergence than the rate $O\big(\frac{1}{N^{1/2}}\big)$ of the Hoeffding-type result \eqref{eq:crate1.MC} in the large-deviation case.
It is noteworthy that the value $\tau=\frac{N_T}{N_T+N_S}$ has been mentioned in the section of ``Experimental Results" in \citet{Blitzer07}. Moreover, a similar trade-off strategy was also discussed in Section 5 of \citet{Lazaric11}. It is in accordance with our theoretical analysis of $\tau$ and the following numerical experiments support the theoretical findings as well.

%%%%
%%%%
\section{Numerical Experiments}\label{sec:experiment}

We have performed numerical experiments to verify the theoretical analysis of the asymptotic behavior of the learning process for representative domain adaptation. Without loss of generality, we only consider the case of $K=2$, {\it i.e.,} there are two source domains and one target domain. The experiment data are generated in the following way.

 For the target domain $\mathcal{Z}^{(T)}=\mathcal{X}^{(T)}\times\mathcal{Y}^{(T)}
 \subset\mathbb{R}^{100}\times\mathbb{R}$, we consider $\mathcal{X}^{(T)}$ as a Gaussian distribution $N(0,1)$ and draw $\{{\bf x}^{(T)}_n\}_{n=1}^{N_T}$ ($N_T=4000$) from $\mathcal{X}^{(T)}$ randomly and independently.
  Let $\beta\in\mathbb{R}^{100}$ be a random vector of a Gaussian distribution $N(1,5)$, and let the random vector $R\in\mathbb{R}^{100}$ be a noise term with $R\sim N(0,0.5)$.
 For any $1\leq n\leq N_T$, we randomly draw $\beta$ and $R$ from $N(1,5)$ and $N(0,0.01)$ respectively, and then generate $y^{(T)}_n\in\mathcal{Y}$ as follows:
\begin{equation*}%\label{eq:MS.T}
y^{(T)}_n=\langle{\bf x}^{(T)}_n, \beta\rangle+R.
\end{equation*}
The derived $\{({\bf x}_n^{(T)}, y_n^{(T)})\}_{n=1}^{N_T}$ ($N_T=4000$) are the samples of the target domain $\mathcal{Z}^{(T)}$ and will be used as the test data. We randomly pick $N'_T=100$ samples from them to form the objective function \eqref{eq:experiment1} and the rest $N''_T=3900$ are used for testing.

Similarly, we generate the sample set $\{({\bf x}_n^{(1)}, y_n^{(1)})\}_{n=1}^{N_1}$ ($N_1=2000$) of the source domain $\mathcal{Z}^{(S_1)}=\mathcal{X}^{(1)}\times\mathcal{Y}^{(1)}
\subset\mathbb{R}^{100}\times\mathbb{R}$: for any $1\leq n\leq N_1$,
\begin{equation*}%\label{eq:MS.1}
y^{(1)}_n=\langle{\bf x}^{(1)}_n, \beta\rangle+R,
\end{equation*}
where ${\bf x}^{(1)}_n\sim N(0.2,0.9)$, $\beta \sim N(1,5)$ and $R\sim N(0,0.5)$.

For the source domain $\mathcal{Z}^{(S_2)}=\mathcal{X}^{(2)}\times\mathcal{Y}^{(2)}
\subset\mathbb{R}^{100}\times\mathbb{R}$, the samples  $\{({\bf x}_n^{(2)}, y_n^{(2)})\}_{n=1}^{N_2}$ ($N_2=2000$) are generated in the following way: for any $1\leq n\leq N_2$,
\begin{equation*}%\label{eq:MS.2}
y^{(2)}_n=\langle{\bf x}^{(2)}_n, \beta\rangle+R,
\end{equation*}
where ${\bf x}^{(2)}_n\sim N(-0.2,1.2)$, $\beta \sim N(1,5)$ and $R\sim N(0,0.5)$.

\begin{figure}[ht]%%%%
\begin{center}
\includegraphics[height=6cm]{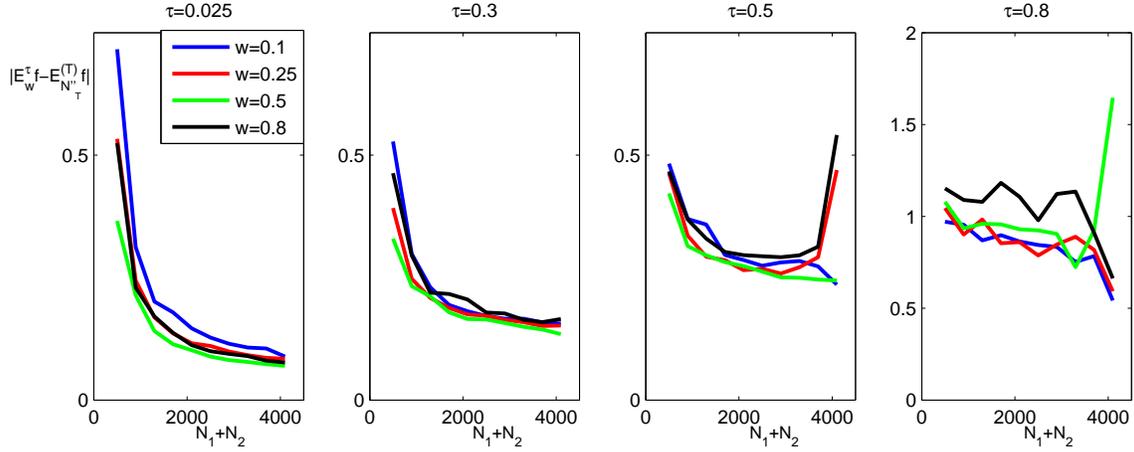}
%\put(-160,13){$N_1+N_2$}\put(-320,271){$|E^{(S)}_wf-E^{(T)}_{N_T}f|$}
\caption{Given $\tau\in\{0.025,0.3,0.5,0.8\}$, the curves for different $w\in\{0.1,0.25,0.5,0.8\}$.}
\label{fig:fig1}
\end{center}
\end{figure}

\begin{figure}[ht]%%%%
\begin{center}
\includegraphics[height=6cm]{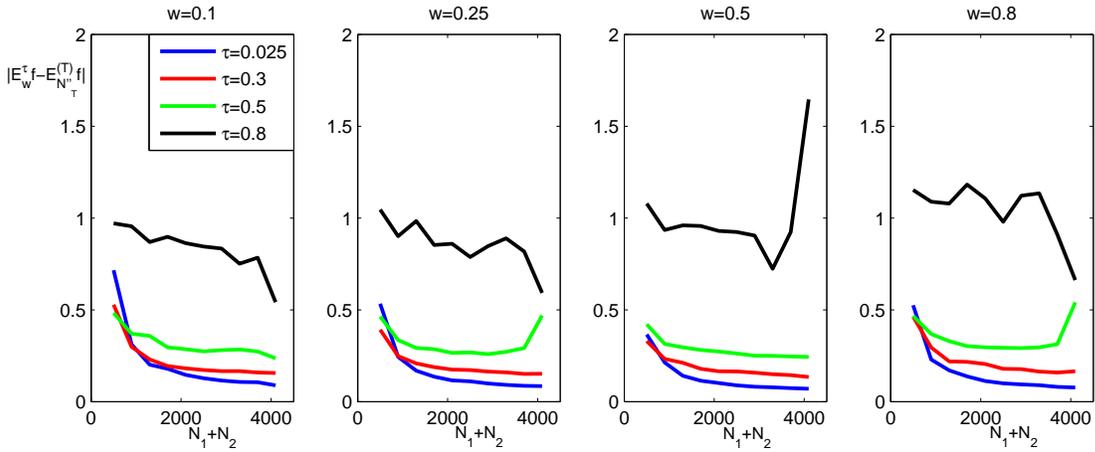}
%\put(-160,13){$N_1+N_2$}\put(-320,271){$|E^{(S)}_wf-E^{(T)}_{N_T}f|$}
\caption{Given $w\in\{0.1,0.25,0.5,0.8\}$, the curves for different $\tau\in\{0.025,0.3,0.5,0.8\}$.}
\label{fig:fig2}
\end{center}
\end{figure}

In this experiment, we use the method of Least Square Regression \citep{Liu:2009:SLEP:manual} to minimize the empirical risk
\begin{align}\label{eq:experiment1}
\mathrm{E}^{\tau}_w(\ell\circ g)=&\frac{\tau}{N'_T}\sum_{n=1}^{N'_T}\ell(g({\bf x}^{(T)}_n),y^{(T)}_n)+\frac{(1-\tau)w}{N_1}\sum_{n=1}^{N_1}\ell(g({\bf x}^{(1)}_n),y^{(1)}_n)\nonumber\\
&+\frac{(1-\tau)(1-w)}{N_2}\sum_{n=1}^{N_2}\ell(g({\bf x}^{(2)}_n),y^{(2)}_n)
 \end{align}
for different combination coefficients $w\in\{0.1,0.25,0.5,0.8\}$ and $\tau\in\{0.025,0.3,0.5,0.8\}$, respectively. Then, we
compute the discrepancy $|E^{\tau}_wf-E^{(T)}_{N''_T}f|$ for each $N_1+N_2$. Since $N_S\gg N'_T$, the initial $N_1$ and $N_2$ both equal to $200$. Each test is repeated $100$ times and the final result is the average of the $100$ results.
After each test, we increment both $N_1$ and $N_2$ by $200$ until $N_1=N_2=2000$. The experimental results are shown in Fig. \ref{fig:fig1} and Fig. \ref{fig:fig2}.

%%%%

%%%%

From Fig. \ref{fig:fig1} and Fig. \ref{fig:fig2}, we can observe that the choice of $\tau$ has a bigger impact on the performance of the learning process than the choice of $w$, and the learning fails when the value of $\tau$ becomes bigger than $0.5$.  This phenomenon can be explained as follows: recalling \eqref{eq:experiment1}, the bigger $\tau$ means that the learning process more relies on the data from the target, while the data from target are not sufficient in the situation of domain adaptation and thus the learning fails. However, for any $w\in\{0.1,0.25,0.5,0.8\}$, the curves of $|E^{\tau}_wf-E^{(T)}_{N''_T}f|$ ($\tau\in\{0.025,0.3\}$) are both decreasing when $N_1+N_2$ increases, which is in accordance with the theoretical results on the asymptotical convergence presented in Theorem \ref{thm:converge.MC}.

Moreover, we have theoretically analyzed how the choices of ${\bf w}$ and $\tau$ affect the rate of convergence of the learning process for representative domain adaptation. Our numerical experiments support the theoretical findings as well. In fact, in Fig. \ref{fig:fig1} and Fig. \ref{fig:fig2}, given any value of $w$, when $\tau \approx \frac{N'_T}{N_1+N_2+N'_T}$, the discrepancy $|E^{\tau}_w f-E^{(T)}_{N''_T}f|$ has the fastest rate of convergence, and
the rate becomes slower as $\tau$ is further away from $\frac{N'_T}{N_1+N_2+N'_T}$. On the other hand, given any value of $\tau\in\{0.025,0.3,0.5\}$, when $w=0.5$, the discrepancy $|E^{\tau}_wf-E^{(T)}_{N''_T}f|$ has the fastest rate of convergence, and the rate becomes slower as $w$ is further away from $0.5$. In this experiment, we set $N_1=N_2$ that implies that $N_2/(N_1+N_2)=0.5$. Thus, the experimental results are in accordance with the theoretical findings (see \eqref{eq:crate2.MC} and \eqref{eq:Hcrate2.MC}), {\it i.e.}, the setting $w=\frac{N_2}{N_1+N_2}$ and $\tau=\frac{N'_T}{N_1+N_2+N'_T}$ can provide the fastest rate of convergence of the learning process for representative domain adaptation.
%%%

\section{Prior Works}\label{sec:compare}

There have been some previous works on the theoretical analysis of domain adaptation with multiple sources \citep[see][]{Ben-David10,Crammer07,Crammer08,Mansour08,Mansour09} and domain adaptation combining source and target data \citep[see][]{Blitzer07,Ben-David10}.

In \citet{Crammer07,Crammer08}, the function class and the loss function are assumed to satisfy the conditions of ``$\alpha$-triangle inequality" and ``uniform convergence bound". Moreover, one has to get some prior information about the disparity between any source domain and the target domain. Under these conditions, some generalization bounds were obtained by using the classical techniques developed under the same-distribution assumption.

\citet{Mansour08} proposed another framework to study the problem of domain adaptation with multiple sources. In this framework, one needs to know some prior knowledge including the exact distributions of the source domains and the hypothesis function with a small loss on each source domain. Furthermore, the target domain and the hypothesis function on the target domain were deemed as the mixture of the source domains and the mixture of the
hypothesis functions on the source domains, respectively. Then, by introducing the R\'enyi divergence, \citet{Mansour09} extended their previous work \citep{Mansour08} to a more general setting, where the distribution of the target domain can be arbitrary and one only needs to know an approximation of the exact distribution of each source domain.
\citet{Ben-David10} also discussed the situation of domain adaptation with the mixture of source domains.

In \citet{Ben-David10,Blitzer07}, domain adaptation combining source and target data was originally proposed and meanwhile a theoretical framework was presented to analyze its properties for the classification tasks by introducing the $\mathcal{H}$-divergence. Under the condition of ``$\lambda$-close", the authors achieved the generalization bounds based on the VC dimension.

\citet{Mansour09b} introduced the {\it discrepancy distance} $\mathrm{disc}_{\ell}(\mathcal{D}^{(S)},\mathcal{D}^{(T)})$ to capture the difference between domains and this quantity can be used in both classification and regression tasks. By extending the classical results of statistical learning theory, the authors obtained the generalization bounds based on the Rademacher complexity for domain adaptation.

% and this proposition is supported by the experiment results shown in Fig. \ref{fig:experiment}.

\section{Conclusion}

In this paper, we study the theoretical properties of the learning process for the so-called representative domain adaptation, which combines data from multiple sources and one target. In particular, we first use the integral probability metric $D_{\mathcal{F}}(S,T)$ to measure the difference between the distributions of two domains. Different from the $\mathcal{H}$-divergence and the discrepancy distance, the integral probability metric can provide a new mechanism to measure the difference between two domains. Additionally, we show that the theoretical analysis in this paper can also be applied to study domain adaptation settings in previous works (see Section \ref{sec:IPM}).

%We present the generalization bounds based on the uniform entropy number and the bounds based on the Rademacher complexity, respectively. By using the derived bounds, we then analyze the asymptotical convergence and the rate of convergence of the learning process.

Then, we develop the Hoeffding-type, the Bennett-type and the McDiarmid-type deviation inequalities for different domains, respectively. We also obtain the symmetrization inequality for representative domain adaptation, which incorporates the discrepancy term $(1-\tau)D^{({\bf w})}_{\mathcal{F}}(S,T)$ which reflects the ``knowledge-transferring" from the source to the target. By applying these inequalities, we achieve two types of generalization bounds for representative domain adaptation: Hoeffding-type and the Bennett-type. They are based on the uniform entropy number and the Rademacher complexity, respectively.

By using the derived bounds, we point out that the asymptotic convergence of the learning process is determined by the complexity of the function class $\mathcal{F}$ measured by the uniform entropy number.
This is partially in accordance with the classical result under the same-distribution assumption \citep[see][Theorem 2.3 and Definition 2.5]{Mendelson03}. We also show that the rate of convergence is affected by the choices of parameters ${\bf w}$ and $\tau$. The setting of $w_k=\frac{N_k}{\sum_{k=1}^KN_k}$ ($1\leq k\leq K$) and $\tau=\frac{N_T}{N_T+\sum_{k=1}^KN_k}$ can lead to the fastest rate of the bounds and the numerical experiments support our theoretical findings as well.

Moreover, we discuss the difference between the Hoeffding-type and the Bennett-type results. The Hoeffding-type results \eqref{eq:crate1.MC} and \eqref{eq:RB.Rade.MC} have well-defined expressions that can explicitly reflect how the parameters ${\bf w}$ and $\tau$ affect the performance of the representative domain adaptation, and its rate of convergence is up to $O(N^{-\frac{1}{2}})$ consistently. In contrast, although the Bennett-type bounds \eqref{eq:RB1.MC} and \eqref{eq:RadeBound} do not reflect the effect of the parameters ${\bf w}$ and $\tau$, they have a faster rate $o(N^{-\frac{1}{2}})$ than the Hoeffding-type results, and meanwhile, provide a more detailed description of the asymptotical behavior of the learning process for representative domain adaptation. The two types complement with each other.

%(with its Bernstain-type alternative expression \eqref{eq:Btype})

Since representative domain adaptation covers domain adaptation with multiple sources and domain adaptation combining source and target, the results of this paper are more general and some of existing results are included as special cases \citep[e.g.][]{zhang2012generalization}. Moreover, it is noteworthy that the generalization bounds \eqref{eq:crate1.MC}, \eqref{eq:RB1.MC}, \eqref{eq:Btype} and \eqref{eq:crate2.MC} can lead to the results based on the fat-shattering dimension, respectively \citep[see][Theorem 2.18]{Mendelson03}. According to Theorem 2.6.4 of \citet{Vaart96}, the bounds based on the VC dimension can also be obtained from the results \eqref{eq:crate1.MC}, \eqref{eq:RB1.MC}, \eqref{eq:Btype} and \eqref{eq:crate2.MC}, respectively.

%%%%

\bibliography{Ref-Domain}

\begin{thebibliography}{33}
\expandafter\ifx\csname natexlab\endcsname\relax\def\natexlab#1{#1}\fi
\expandafter\ifx\csname url\endcsname\relax
  \def\url#1{{\tt #1}}\fi

\bibitem[Bartlett et~al.(2005)Bartlett, Bousquet, and Mendelson]{Bartlett05}
P.L. Bartlett, O.~Bousquet, and S.~Mendelson.
\newblock Local rademacher complexities.
\newblock {\em Annals of Statistics}, 33\penalty0 (4):\penalty0 1497--1537,
  2005.

\bibitem[Ben-David et~al.(2010)Ben-David, Blitzer, Crammer, Kulesza, Pereira,
  and Vaughan]{Ben-David10}
S.~Ben-David, J.~Blitzer, K.~Crammer, A.~Kulesza, F.~Pereira, and J.W. Vaughan.
\newblock A theory of learning from different domains.
\newblock {\em Machine Learning}, 79\penalty0 (1-2):\penalty0 151--175, 2010.

\bibitem[Ben-David et~al.(2007)Ben-David, Blitzer, Crammer, and
  Pereira]{Ben-David06}
S.~Ben-David, J.~Blitzer, K.~Crammer, and F.~Pereira.
\newblock Analysis of representations for domain adaptation.
\newblock {\em Advances in neural information processing systems}, 19:\penalty0
  137, 2007.

\bibitem[Bennett(1962)]{Bennett62}
G.~Bennett.
\newblock Probability inequalities for the sum of independent random variables.
\newblock {\em Journal of the American Statistical Association}, 57\penalty0
  (297):\penalty0 33--45, 1962.

\bibitem[Bian et~al.(2012)Bian, Tao, and Rui]{Bian12}
W.~Bian, D.~Tao, and Y.~Rui.
\newblock Cross-domain human action recognition.
\newblock {\em IEEE Transactions on Systems, Man, and Cybernetics, Part B:
  Cybernetics}, 42\penalty0 (2):\penalty0 298--307, 2012.

\bibitem[Bickel et~al.(2007)Bickel, Br{\"u}ckner, and Scheffer]{Bickel07}
S.~Bickel, M.~Br{\"u}ckner, and T.~Scheffer.
\newblock Discriminative learning for differing training and test
  distributions.
\newblock In {\em Proceedings of the 24th international conference on Machine
  learning}, pages 81--88. ACM, 2007.

\bibitem[Blitzer et~al.(2007{\natexlab{a}})Blitzer, Crammer, Kulesza, Pereira,
  and Wortman]{Blitzer07}
J.~Blitzer, K.~Crammer, A.~Kulesza, F.~Pereira, and J.~Wortman.
\newblock Learning bounds for domain adaptation.
\newblock {\em Advances in neural information processing systems}, 20:\penalty0
  129--136, 2007{\natexlab{a}}.

\bibitem[Blitzer et~al.(2007{\natexlab{b}})Blitzer, Dredze, and
  Pereira]{Blitzer07a}
J.~Blitzer, M.~Dredze, and F.~Pereira.
\newblock Biographies, bollywood, boom-boxes and blenders: Domain adaptation
  for sentiment classification.
\newblock {\em Annual Meeting-Association For Computational Linguistics},
  45\penalty0 (1):\penalty0 440, 2007{\natexlab{b}}.

\bibitem[Blitzer et~al.(2006)Blitzer, McDonald, and Pereira]{Blitzer06}
J.~Blitzer, R.~McDonald, and F.~Pereira.
\newblock Domain adaptation with structural correspondence learning.
\newblock In {\em Proceedings of the 2006 Conference on Empirical Methods in
  Natural Language Processing}, pages 120--128. Association for Computational
  Linguistics, 2006.

\bibitem[Blumer et~al.(1989)Blumer, Ehrenfeucht, Haussler, and
  Warmuth]{Blumer89}
A.~Blumer, A.~Ehrenfeucht, D.~Haussler, and M.K. Warmuth.
\newblock Learnability and the vapnik-chervonenkis dimension.
\newblock {\em Journal of the ACM (JACM)}, 36\penalty0 (4):\penalty0 929--965,
  1989.

\bibitem[Bousquet(2002)]{Bousquet02}
O.~Bousquet.
\newblock A bennett concentration inequality and its application to suprema of
  empirical processes.
\newblock {\em Comptes Rendus Mathematique}, 334\penalty0 (6):\penalty0
  495--500, 2002.

\bibitem[Bousquet et~al.(2004)Bousquet, Boucheron, and Lugosi]{Bousquet04}
O.~Bousquet, S.~Boucheron, and G.~Lugosi.
\newblock Introduction to statistical learning theory.
\newblock {\em Advanced Lectures on Machine Learning}, pages 169--207, 2004.

\bibitem[Crammer et~al.(2007)Crammer, Kearns, and Wortman]{Crammer07}
K.~Crammer, M.~Kearns, and J.~Wortman.
\newblock Learning from multiple sources.
\newblock {\em Advances in Neural Information Processing Systems}, 19:\penalty0
  321, 2007.

\bibitem[Crammer et~al.(2008)Crammer, Kearns, and Wortman]{Crammer08}
K.~Crammer, M.~Kearns, and J.~Wortman.
\newblock Learning from multiple sources.
\newblock {\em The Journal of Machine Learning Research}, 9:\penalty0
  1757--1774, 2008.

\bibitem[Hoeffding(1963)]{Hoeffding63}
W.~Hoeffding.
\newblock Probability inequalities for sums of bounded random variables.
\newblock {\em Journal of the American statistical association}, 58\penalty0
  (301):\penalty0 13--30, 1963.

\bibitem[Hussain and Shawe-Taylor(2011)]{Hussain11}
Z.~Hussain and J.~Shawe-Taylor.
\newblock Improved loss bounds for multiple kernel learning.
\newblock {\em J. Mach. Learn. Res.-Proc. Track}, 15:\penalty0 370--377, 2011.

\bibitem[Jiang and Zhai(2007)]{Jiang07}
J.~Jiang and C.~Zhai.
\newblock Instance weighting for domain adaptation in nlp.
\newblock {\em Annual Meeting-Association For Computational Linguistics},
  45\penalty0 (1):\penalty0 264, 2007.

\bibitem[Lazaric and Restelli(2011)]{Lazaric11}
A.~Lazaric and M.~Restelli.
\newblock Transfer from multiple mdps.
\newblock {\em Advances in neural information processing systems}, 2011.

\bibitem[Liu et~al.(2009)Liu, Ji, and Ye]{Liu:2009:SLEP:manual}
J.~Liu, S.~Ji, and J.~Ye.
\newblock {\em SLEP: Sparse Learning with Efficient Projections}.
\newblock Arizona State University, 2009.
\newblock URL \url{http://www.public.asu.edu/$\sim$jye02/Software/SLEP}.

\bibitem[Mansour et~al.(2009{\natexlab{a}})Mansour, Mohri, and
  Rostamizadeh]{Mansour09b}
Y.~Mansour, M.~Mohri, and A.~Rostamizadeh.
\newblock Domain adaptation: Learning bounds and algorithms.
\newblock In {\em The 22nd Annual Conference on Learning Theory (COLT 2009)},
  2009{\natexlab{a}}.

\bibitem[Mansour et~al.(2009{\natexlab{b}})Mansour, Mohri, and
  Rostamizadeh]{Mansour08}
Y.~Mansour, M.~Mohri, and A.~Rostamizadeh.
\newblock Domain adaptation with multiple sources.
\newblock {\em Advances in neural information processing systems}, 21:\penalty0
  1041--1048, 2009{\natexlab{b}}.

\bibitem[Mansour et~al.(2009{\natexlab{c}})Mansour, Mohri, and
  Rostamizadeh]{Mansour09}
Y.~Mansour, M.~Mohri, and A.~Rostamizadeh.
\newblock Multiple source adaptation and the r{\'e}nyi divergence.
\newblock In {\em Proceedings of the Twenty-Fifth Conference on Uncertainty in
  Artificial Intelligence}, pages 367--374, 2009{\natexlab{c}}.

\bibitem[Mendelson(2003)]{Mendelson03}
S.~Mendelson.
\newblock A few notes on statistical learning theory.
\newblock {\em Advanced Lectures on Machine Learning}, pages 1--40, 2003.

\bibitem[M{\"u}ller(1997)]{Muller97}
A.~M{\"u}ller.
\newblock Integral probability metrics and their generating classes of
  functions.
\newblock {\em Advances in Applied Probability}, 29\penalty0 (2):\penalty0
  429--443, 1997.

\bibitem[Rachev(1991)]{Rachev}
S.T. Rachev.
\newblock {\em Probability metrics and the stability of stochastic models}.
\newblock New York: Wiley, 1991.

\bibitem[Reid and Williamson(2011)]{Reid11}
M.~Reid and B.~Williamson.
\newblock Information, divergence and risk for binary experiments.
\newblock {\em Journal of Machine Learning Research}, 12:\penalty0 731--817,
  2011.

\bibitem[Sriperumbudur et~al.(2012)Sriperumbudur, Fukumizu, Gretton,
  Sch{\"o}lkopf, and Lanckriet]{Sriperumbudur12}
B.K. Sriperumbudur, K.~Fukumizu, A.~Gretton, B.~Sch{\"o}lkopf, and G.R.G.
  Lanckriet.
\newblock On the empirical estimation of integral probability metrics.
\newblock {\em Electronic Journal of Statistics}, 6:\penalty0 1550--1599, 2012.

\bibitem[Van~der Vaart and Wellner(1996)]{Vaart96}
A.~Van~der Vaart and J.~Wellner.
\newblock {\em Weak Convergence and Empirical Processes: with Aapplications to
  Statistics}.
\newblock Springer, 1996.

\bibitem[Vapnik(1998)]{Vapnik98}
V.N. Vapnik.
\newblock {\em Statistical Learning Theory}.
\newblock Wiley, 1998.

\bibitem[Wu and Dietterich(2004)]{Wu04}
P.~Wu and T.G. Dietterich.
\newblock Improving svm accuracy by training on auxiliary data sources.
\newblock In {\em Proceedings of the twenty-first international conference on
  Machine learning}, page 110. ACM, 2004.

\bibitem[Zhang(2013)]{zhang2013bennett}
C.~Zhang.
\newblock Bennett-type generalization bounds: Large-deviation case and faster
  rate of convergence.
\newblock {\em The Conference on Uncertainty in Artificial Intelligence (UAI)},
  2013.

\bibitem[Zhang et~al.(2012)Zhang, Zhang, and Ye]{zhang2012generalization}
C.~Zhang, L.~Zhang, and J.~Ye.
\newblock Generalization bounds for domain adaptation.
\newblock {\em Advances in neural information processing systems (NIPS)}, 2012.

\bibitem[Zolotarev(1984)]{Zolotarev84}
V.M. Zolotarev.
\newblock Probability metrics.
\newblock {\em Theory of Probability and its Application}, 28\penalty0
  (1):\penalty0 278--302, 1984.

\end{thebibliography}
\bibliographystyle{plainnat}

\newpage

\appendix

\section{Deviation Inequalities and Symmetrization Inequalities}\label{app:devi&symm}

By adopting a martingale method, we develop the Hoeffding-type, the Bennett-type and the McDiarmid-type deviation inequalities for multiple domains, respectively. Moreover, we present a symmetrization inequality for representative domain adaptation.

\subsection{Deviation Inequalities for Multiple Domains}

Deviation (or concentration) inequalities play an essential role in obtaining the generalization bounds for a certain learning process. Generally, specific deviation inequalities need to be developed for different learning processes. There are many popular deviation and concentration inequalities, for example, Hoeffding's inequality \citep{Hoeffding63}, McDiarmid's inequality \citep[see][]{Bousquet04}, Bennett's inequality \citep{Bennett62}, Bernstein's inequality and Talagrand's inequality. We refer to \citet{Bousquet04,Bousquet02} for their application to the learning process (or empirical process). Note that these results are all built under the same-distribution assumption, and thus they are
not applicable (or at least cannot be directly applied) to the learning process of the representative domain adaptation considered in this paper, where the samples are drawn from multiple domains.
Next, we extend the classical Hoeffding's inequality, Bennett's inequality and McDiarmid's inequality to the scenario of multiple domains, respectively.

\subsubsection{Hoeffding-type Deviation Inequality}

We first present the Hoeffding-type deviation inequality for multiple domains, where the random variables can take values from different domains.

\begin{theorem}\label{thm:Hdineq.MC}
Assume that $f$ is a bounded function with the range $[a,b]$. Let ${\bf Z}_{1}^{N_k}=\{{\bf z}^{(k)}_n\}_{n=1}^{N_k}$
and $\overline{{\bf Z}}_{1}^{N_{T}}:=\{{\bf z}_n^{(T)}\}_{n=1}^{N_T}$ be the sets of i.i.d. samples drawn from the source domain $\mathcal{Z}^{(S_k)}\subset\mathbb{R}^L$ $(1\leq k\leq K)$ and the target domain $\mathcal{Z}^{(T)}\subset\mathbb{R}^L$, respectively. Given $\tau\in[0,1)$ and  ${\bf w}\in[0,1]^{K}$ with $\sum_{k=1}^Kw_k=1$, we define a function $F^\tau_{\bf w}:\mathbb{R}^{L(N_T+\sum_{k=1}^KN_k)}\rightarrow\mathbb{R}$ as
\begin{align}\label{eq:F.MC}
    F^\tau_{\bf w}\left(\{{\bf Z}_{1}^{N_k}\}_{k=1}^K,\overline{{\bf Z}}_{1}^{N_{T}}\right)
:=\tau \Big(\prod_{k=1}^KN_k\Big)\sum_{n=1}^{N_T}f({\bf z}^{(T)}_n)+(1-\tau)N_T\sum_{k=1}^Kw_k
    \Big(\prod_{i\not=k}N_i\Big)\sum_{n=1}^{N_k}f({\bf z}^{(k)}_n).
\end{align}
Then, we have for any $\xi>0$,
\begin{align}\label{eq:Hdineq.MC1}
&\mathrm{Pr}\left\{\big|\mathrm{E}^{(*)}F^{\tau}_{\bf w}-F^{\tau}_{\bf w}\big(\{{\bf Z}_{1}^{N_k}\}_{k=1}^K,\overline{{\bf Z}}_{1}^{N_{T}}\big)\big|>\xi\right\}\\
    \leq&2\exp\left\{-\frac{2\xi^2}
{N_T(b-a)^2\big(\prod_{k=1}^KN_k\big)
\Big(\tau^2\big(\prod_{k=1}^KN_k\big)+
\sum_{k=1}^{K}(1-\tau)^2N_Tw_k^2\big(\prod_{i\not=k}N_i\big)\Big)}
\right\},\nonumber
\end{align}
where the expectation $\mathrm{E}^{(*)}$ is taken on all source domains $\{\mathcal{Z}^{(S_k)}\}_{k=1}^K$ and the target domain $\mathcal{Z}^{(T)}$.
\end{theorem}
This result is an extension of the classical Hoeffding's inequality under the same-distribution assumption \citep{Bousquet04}. Compared to the classical result, the resulted deviation inequality \eqref{eq:Hdineq.MC1} is suitable to the scenario of multiple domains. These two inequalities coincide when there is only one domain or all domains match.

\subsubsection{Bennett-type Deviation Inequality}

It is noteworthy that Hoeffding's inequality is obtained by only using the information of the expectation of the random variable \citep{Hoeffding63}. If the information of the variance is also taken into consideration, one can further obtain Bennett's inequality \citep{Bennett62}. Similar to the above, we generalize the classical Bennett's inequality to a more general setting, where the random variables can take values from different domains.

\begin{theorem}\label{thm:dineq.MC}
Under the notations of Theorem \ref{thm:Hdineq.MC},
then we have for any $\alpha>0$ and $\xi>0$,
\begin{align*}%\label{eq:dineq.MC1}
&\mathrm{Pr}\left\{\big|\mathrm{E}^{(*)}F^{\tau}_{\bf w}-F^{\tau}_{\bf w}\big(\{{\bf Z}_{1}^{N_k}\}_{k=1}^K,\overline{{\bf Z}}_{1}^{N_{T}}\big)\big|>\xi\right\}
    \leq2\mathrm{e}^{\Phi(\alpha)-\alpha\xi},
\end{align*}
where the expectation $\mathrm{E}^{(*)}$ is taken on all source domains $\{\mathcal{Z}^{(S_k)}\}_{k=1}^K$ and the target domain $\mathcal{Z}^{(T)}$, and
\begin{align*}%\label{eq:dineq.MC2}
   \Phi(\alpha):=&N_T\left(\mathrm{e}^{\alpha\tau(\prod_{k=1}^KN_k)(b-a)}-1
-\alpha\tau\Big(\prod_{k=1}^KN_k\Big)(b-a)\right)\nonumber\\
&+\sum_{k=1}^KN_k\left(\mathrm{e}^{\alpha(1-\tau)N_Tw_k(\prod_{i\not=k}N_i)(b-a)}
-1-\alpha(1-\tau)N_Tw_k\Big(\prod_{i\not=k}N_i\Big)(b-a)\right).
\end{align*}
Furthermore, by setting $w_k=N_k/\sum_{k=1}^KN_k$ ($1\leq k\leq K$) and $\tau=N_T/(N_T+\sum_{k=1}^KN_k)$, then we have
\begin{align}\label{eq:dineq.MC}
&\mathrm{Pr}\left\{\big|\mathrm{E}^{(*)}F^{\tau}_{\bf w}-F^{\tau}_{\bf w}\big(\{{\bf Z}_{1}^{N_k}\}_{k=1}^K,\overline{{\bf Z}}_{1}^{N_{T}}\big)\big|>\xi\right\}\nonumber\\
    \leq&2\exp\left\{\Big(N_T+\sum_{k=1}^KN_k\Big)\Gamma\left( \frac{\xi}{ N_T\big(\prod_{k=1}^KN_k\big)(b-a)}\right)\right\},
\end{align}
where
\begin{align}\label{eq:Gamma}
\Gamma(x):=x-(x+1)\ln(x+1).
\end{align}
\end{theorem}

Compared to the classical Bennett's inequality \citep{Bousquet02,Bennett62}, the derived inequality \eqref{eq:dineq.MC} is suitable to the scenario of multiple domains and these two inequalities coincide when there is only one domain or all the domains match.

Differing from the Hoeffding-type inequality \eqref{eq:Hdineq.MC1}, the derived inequality \eqref{eq:dineq.MC} does not explicitly reflect how the choices of ${\bf w}$ and $\tau$ affect the right-hand side of the inequality. The presented result is not completely satisfying, because it is hard to obtain the analytical expression of the inverse function of $\Phi'(\alpha)$ and then we cannot achieve the analytical result that incorporates the parameters ${\bf w}$ and $\tau$ (see the proofs of Theorems \ref{thm:Hdineq.MC} \& \ref{thm:dineq.MC}). Instead, by the method of Lagrange multiplier (see Lemma \ref{lem:sec.lem}), we have shown that setting $w_k=N_k/\sum_{k=1}^KN_k$ ($1\leq k\leq K$) and $\tau=N_T/(N_T+\sum_{k=1}^KN_k)$ can result in the minimum of the term $\Phi(\alpha)$ with respect to ${\bf w}$ and $\tau$, and then get the Bennett-type deviation inequality \eqref{eq:dineq.MC}.%\footnote{In the process of obtaining the Hoeffding-type deviation inequality, the corresponding function is a quadratic function, so we can obtain an explicit result with respect to the parameters ${\bf w}$ and $\tau$ (see the proof of Theorem \ref{thm:Hdineq.MC}).}

By Cauchy-Schwarz inequality, such a setting of ${\bf w}$ and $\tau$ can also lead to the minimum of the Hoeffding-type result \eqref{eq:dineq.MC}. Because of its well-defined expression, we can use the Hoeffding-type result to analyze how the parameters ${\bf w}$ and $\tau$ affect the generalization bounds.
However, the Bennett-type results can provide a faster rate $o(N^{-\frac{1}{2}})$ of convergence and give a more detailed description to the asymptotical behavior of the learning process than the Hoeffding-type results, which consistently provide the rate $O(N^{-\frac{1}{2}})$ regardless of the discrepancy between the expected and the empirical risks.

%However, since the Bennett-type result is obtained by using the information of both expectation and variance, we will also show that it can provide a faster rate of convergence than the Hoeffding-type result in the large-deviation case (see Remark \ref{rem:large} and Theorem \ref{thm:crate2.MC}).

\subsubsection{McDiarmid-type Deviation Inequality}

The following is the classical McDiarmid's inequality that is one of the most frequently used deviation inequalities in statistical learning theory and has been widely used to obtain generalization bounds based on the Rademacher complexity under the assumption of same distribution \citep[see][Theorem 6]{Bousquet04}.

\begin{theorem}[McDiamid's Inequality]\label{thm:Mcdiarmid0}
Let ${\bf z}_{1},\cdots,{\bf z}_{N}$ be $N$ independent random variables taking values from the domain $\mathcal{Z}$. Assume that the function $H:\mathcal{Z}^N\rightarrow\mathbb{R}$  satisfies the condition of bounded difference:
for all $1\leq n\leq N$,
\begin{align}\label{eq:dif.cond}
&\sup_{{\bf z}_{1},\cdots,{\bf z}_{N},{\bf z'}_n}\Big|H\big({\bf z}_{1},\cdots,{\bf z}_n,\cdots,{\bf z}_{N}\big)
-H\big({\bf z}_{1},\cdots,{\bf z}'_n,\cdots,{\bf z}_{N}\big)\Big|\leq c_n.
\end{align}
Then, for any $\xi>0$
\begin{equation*}
\mathrm{Pr}\left\{H\big({\bf z}_{1},\cdots,{\bf z}_n,\cdots,{\bf z}_{N}\big)-\mathrm{E}\Big\{H\big({\bf z}_{1},\cdots,{\bf z}_n,\cdots,{\bf z}_{N}\big)\Big\}\geq\xi\right\}
\leq\exp\left\{-2\xi^2/\sum_{n=1}^{N}c_n^2\right\}.
\end{equation*}
\end{theorem}
As shown in Theorem \ref{thm:Mcdiarmid0}, the classical McDiarmid's inequality is valid under the condition that random variables ${\bf z}_{1},\cdots,{\bf z}_{N}$ are independent and drawn from the same domain. Next, we generalize this inequality to a more general setting, where the independent random variables can take values from different domains.

\begin{theorem}\label{thm:Mcdiarmid}
Given independent domains $\mathcal{Z}^{(S_k)}$ ($1\leq k\leq K$), let ${\bf Z}_{1}^{N_k}:=\{{\bf z}_n^{(k)}\}_{n=1}^{N_k}$ be $N_k$ independent random variables taking values from the domain $\mathcal{Z}^{(S_k)}$ for any $1\leq k\leq K$. Assume that the function $H:\big(\mathcal{Z}^{(S_1)}\big)^{N_1}\times\cdots\times\big(\mathcal{Z}^{(S_K)}\big)^{N_K}\rightarrow\mathbb{R}$  satisfies the condition of bounded difference:
for all $1\leq k\leq K$ and $1\leq n\leq N_k$,
\begin{align}\label{eq:dif.cond.MC}
&\sup_{{\bf Z}_{1}^{N_1},\cdots,{\bf Z}_{1}^{N_K},{\bf z'}^{(k)}_n}\Big|H\big({\bf Z}_{1}^{N_1},\cdots,{\bf Z}_{1}^{N_{k-1}},{\bf z}_1^{(k)},\cdots,{\bf z}_n^{(k)},\cdots,{\bf z}_{N_k}^{(k)},{\bf Z}_{1}^{N_{k+1}},\cdots,{\bf Z}_{1}^{N_K}\big)\nonumber\\
&\qquad\qquad-H\big({\bf Z}_{1}^{N_1},\cdots,{\bf Z}_{1}^{N_{k-1}},{\bf z}_1^{(k)},\cdots,{\bf z'}_n^{(k)},\cdots,{\bf z}_{N_k}^{(k)},{\bf Z}_{1}^{N_{k+1}},\cdots,{\bf Z}_{1}^{N_K}\big)\Big|\leq c^{(k)}_n.
\end{align}
Then, for any $\xi>0$
\begin{equation}\label{eq:McDiarmid.MC}
\mathrm{Pr}\left\{H\big({\bf Z}_{1}^{N_1},\cdots,{\bf Z}_{1}^{N_K}\big)-\mathrm{E}\Big\{H\big({\bf Z}_{1}^{N_1},\cdots,{\bf Z}_{1}^{N_K}\big)\Big\}\geq\xi\right\}
\leq\exp\left\{-2\xi^2/\sum_{k=1}^{K}\sum_{n=1}^{N_k}(c^{(k)}_n)^2\right\}.
\end{equation}
Furthermore, if all $c^{(k)}_n$ ($1\leq k\leq K;1\leq n\leq N_k$) are equal to $c$, then there holds that for any $\xi>0$
\begin{align}\label{eq:McDiarmid.Bennett}
\mathrm{Pr}\left\{H\big({\bf Z}_{1}^{N_1},\cdots,{\bf Z}_{1}^{N_K}\big)-\mathrm{E}\Big\{H\big({\bf Z}_{1}^{N_1},\cdots,{\bf Z}_{1}^{N_K}\big)\Big\}\geq\xi\right\}
\leq\exp\left\{\left(\sum_{k=1}^KN_k\right)\Gamma\left(\frac{\xi}
{c\sum_{k=1}^KN_k}\right)\right\}.
\end{align}
\end{theorem}
Similarly, the derived inequality \eqref{eq:McDiarmid.MC} coincides with the classical one (see Theorem \ref{thm:Mcdiarmid0}) when there is only one domain or all domains match. The inequality \eqref{eq:McDiarmid.Bennett} is also a generalized version of the classical Bennett's inequality.

%%%%%%%%%%%

\subsection{Symmetrization Inequalities}

Symmetrization inequalities are mainly used to replace the expected risk by an empirical risk computed on another sample set that is independent of the given sample set but has the same distribution. In this manner, the generalization bounds can be achieved based on a certain complexity measure, for example, the covering number and the VC dimension. However, the classical symmetrization result is built under the same-distribution assumption \citep[see][]{Bousquet04}. Here, we propose a symmetrization inequality for representative domain adaptation.

%The symmetrization inequality for domain adaptation is presented in the following theorem:

\begin{theorem}\label{thm:sym.MC}
Assume that $\mathcal{F}$ is a function class with the range $[a,b]$. Let the sample sets
$\{{\bf Z}_1^{N_k}\}_{k=1}^K$ and $\{{\bf Z'}_1^{N_k}\}_{k=1}^K$
be drawn from the multiple sources $\{\mathcal{Z}^{(S_k)}\}_{k=1}^K$ respectively,
and ${\bf \overline{Z}}_{1}^{N_T}$ and ${\bf \overline{Z'}}_{1}^{N_T}$
be drawn from the target domain $\mathcal{Z}^{(T)}$.
Then, for any $\tau\in[0,1)$ and ${\bf w}\in[0,1]^{K}$ with $\sum_{k=1}^Kw_k=1$, given any $\xi>(1-\tau)D_{\mathcal{F}}^{({\bf w})}(S,T)$, we have for any $N_T,N_1,\cdots,N_K\in\mathbb{N}$
such that
\begin{equation}\label{eq:sym.condition}
  \frac{\tau^2(b-a)^2}{N_T(\xi')^2}
+\sum_{k=1}^K\frac{(1-\tau)^2w_k^2(b-a)^2}{N_k(\xi')^2}\leq \frac{1}{8}
\end{equation}
with $\xi':=\xi-(1-\tau)D_{\mathcal{F}}^{({\bf w})}(S,T)$,
\begin{align}\label{eq:sym1.MC}
    \mathrm{Pr}\left\{\sup_{f\in\mathcal{F}}\big|\mathrm{E}^{(T)}f
-\mathrm{E}^{\tau}_{\bf w}f\big|>\xi\right\}
    \leq2\mathrm{Pr}\left\{\sup_{f\in\mathcal{F}}\big|
\mathrm{E'}^{\tau}_{\bf w}f-\mathrm{E}^{\tau}_{\bf w}f\big|>\frac{\xi'}{2}\right\},
    \end{align}
where
\begin{equation*}%\label{eq:Dist.MC}
D^{({\bf w})}_{\mathcal{F}}(S,T):=\sum_{k=1}^Kw_kD_{\mathcal{F}}(S_k,T).
\end{equation*}
\end{theorem}

%%%
This theorem shows that given $\xi>(1-\tau)D_{\mathcal{F}}^{({\bf w})}(S,T)$, the probability of the event:
\begin{equation*}
\sup_{f\in\mathcal{F}}\big|\mathrm{E}^{(T)}f-\mathrm{E}^{\tau}_{\bf w}f\big|>\xi
\end{equation*}
can be bounded by using the probability of the event:
\begin{equation}
\sup_{f\in\mathcal{F}}\big|\mathrm{E'}^{\tau}_{\bf w}f-\mathrm{E}^{\tau}_{\bf w}f\big|>\frac{\xi-(1-\tau)D_{\mathcal{F}}^{({\bf w})}(S,T)}{2}
\end{equation}
that is only determined by the characteristics of the sample sets $\{{\bf Z}_1^{N_k}\}_{k=1}^K$, $\{{\bf Z'}_1^{N_k}\}_{k=1}^K$, ${\bf \overline{Z}}_{1}^{N_T}$ and ${\bf \overline{Z'}}_{1}^{N_T}$, when the condition \eqref{eq:sym.condition} is satisfied. Compared to the classical symmetrization result under the same-distribution assumption \citep[see][]{Bousquet04}, there is a discrepancy term $(1-\tau)D_{\mathcal{F}}^{({\bf w})}(S,T)$ in the derived inequality, which embodies the ``knowledge-transferring" in the learning process for representative domain adaptation. Especially, the two results will coincide when any source domain and the target domain match, that is, $D_{\mathcal{F}}^{({\bf w})}(S,T)=D_{\mathcal{F}}(S_k,T)=0$ holds for any $1\leq k\leq K$.

\section{Proofs of Main Results}\label{app:proof}

Here, we prove the main results of this paper including Theorem \ref{thm:Hdineq.MC}, Theorem \ref{thm:dineq.MC}, Theorem \ref{thm:Mcdiarmid}, Theorem \ref{thm:sym.MC}, Theorem \ref{thm:Hcrate.MC}, Theorem \ref{thm:RB1.MC} and Theorem \ref{thm:RB.Rade.MC}.

\subsection{Proof of Theorem \ref{thm:Hdineq.MC} }

The proof of Theorem \ref{thm:Hdineq.MC} is processed by a martingale method. Before the formal proofs, we need to introduce some essential notations.

Let $\overline{{\bf Z}}_{1}^{N_{T}}:=\{{\bf z}_n^{(T)}\}_{n=1}^{N_T}$ be the sample set drawn from the target domain $\mathcal{Z}^{(T)}$ and $\{{\bf Z}_{1}^{N_k}\}_{k=1}^K$ be the sample sets drawn from multiple sources $\{\mathcal{Z}^{(S_k)}\}_{k=1}^K$, respectively.
Given $\tau\in[0,1)$ and  ${\bf w}\in[0,1]^{K}$ with $\sum_{k=1}^Kw_k=1$, we denote
\begin{align}\label{eq:F1.MC}
    F_{\bf w}\big(\{{\bf Z}_{1}^{N_k}\}_{k=1}^K\big):=&(1-\tau)N_T\sum_{k=1}^Kw_k
    \Big(\prod_{i\not=k}N_i\Big)\sum_{n=1}^{N_k}f({\bf z}^{(k)}_n);\nonumber\\
    F_T({\bf \overline{Z}}_{1}^{N_T}):=&\tau \big(\prod_{k=1}^KN_k\big)\sum_{n=1}^{N_T}f({\bf z}^{(T)}_n).
\end{align}
Recalling \eqref{eq:F.MC}, it is evident that $$F^\tau_{\bf w}\big(\{{\bf Z}_{1}^{N_k}\}_{k=1}^K,\overline{{\bf Z}}_{1}^{N_{T}}\big)=F_{\bf w}\big(\{{\bf Z}_{1}^{N_k}\}_{k=1}^K\big)+F_T(\overline{{\bf  Z}}_{1}^{N_T}).$$

Define a random variable
\begin{equation}\label{eq:Sn1.M}
    S^{(k)}_{n}:=\mathrm{E}^{(S)}\left\{F_{\bf w}\big(\{{\bf Z}_{1}^{N_k}\}_{k=1}^K\big)|{\bf Z}_{1}^{N_1},{\bf Z}_{1}^{N_2},\cdots,{\bf Z}_{1}^{N_{k-1}},{\bf Z}_1^{n}\right\},\;\; 1\leq k\leq K,\;0\leq n\leq N_k,
\end{equation}
where
$${\bf Z}_{1}^{n}=\{{\bf z}^{(k)}_1,{\bf z}^{(k)}_2,\cdots,{\bf z}^{(k)}_{n}\}\subseteq{\bf Z}_{1}^{N_k},  \mbox{  and   }   {\bf Z}_{1}^0=\varnothing.$$
It is clear that
$$S_0^{(1)}=\mathrm{E}^{(S)}F_{\bf w} \mbox{  and }  S^{(K)}_{N_K}=F_{\bf w}(\{{\bf Z}_{1}^{N_k}\}_{k=1}^K),$$
where $\mathrm{E}^{(S)}$ stands for the expectation taken on all source domains $\{\mathcal{Z}^{(S_k)}\}_{k=1}^K$.

Then, according to \eqref{eq:F1.MC} and \eqref{eq:Sn1.M},
we have for any $1\leq k\leq K$ and $1\leq n\leq N_k$:
\begin{align}%\label{eq:Sn2.M}
S^{(k)}_n-S_{n-1}^{(k)}
=&\mathrm{E}^{(S)}\left\{F_{\bf w}(\{{\bf Z}_{1}^{N_k}\}_{k=1}^K)\big|{\bf Z}_{1}^{N_1},{\bf Z}_{1}^{N_2},\cdots,{\bf Z}_{1}^{N_{k-1}},{\bf Z}_1^{n}\right\}\nonumber\\
&-\mathrm{E}^{(S)}\left\{F_{\bf w}(\{{\bf Z}_{1}^{N_k}\}_{k=1}^K)\big|{\bf Z}_{1}^{N_1},{\bf Z}_{1}^{N_2},\cdots,{\bf Z}_{1}^{N_{k-1}},{\bf Z}_1^{n-1}\right\}\nonumber\\
    =&(1-\tau)N_T\mathrm{E}^{(S)}\left\{\sum_{k=1}^Kw_k
    \Big(\prod_{i\not=k}N_i\Big)\sum_{n=1}^{N_k}f({\bf z}^{(k)}_n)\big|{\bf Z}_{1}^{N_1},{\bf Z}_{1}^{N_2},\cdots,{\bf Z}_{1}^{N_{k-1}},{\bf Z}_1^{n}\right\}\nonumber\\
    & -(1-\tau)N_T\mathrm{E}^{(S)}\left\{\sum_{k=1}^Kw_k\Big(\prod_{i\not=k}N_i\Big)
    \sum_{n=1}^{N_k}f({\bf z}^{(k)}_n)\big|{\bf Z}_{1}^{N_1},{\bf Z}_{1}^{N_2},\cdots,{\bf Z}_{1}^{N_{k-1}},{\bf Z}_1^{n-1}\right\}\nonumber\\
     =&(1-\tau)N_T\sum_{l=1}^{k-1}w_l\Big(\prod_{i\not=l}N_i\Big)\sum_{j=1}^{N_l}f({\bf z}^{(l)}_j)+(1-\tau)N_Tw_k\big(\prod_{i\not=k}N_i\big)\sum_{j=1}^nf({\bf z}_j^{(k)})\nonumber\\
     &+(1-\tau)N_T\mathrm{E}^{(S)}\left\{\sum_{l=k+1}^Kw_l
     \Big(\prod_{i\not=l}N_i\Big)\sum_{j=1}^{N_l}f({\bf z}^{(l)}_j)+w_k\Big(\prod_{i\not=k}N_i\Big)\sum_{j=n+1}^{N_k}f({\bf z}^{(k)}_j)\right\}\nonumber\\
     &-(1-\tau)N_T
    \sum_{l=1}^{k-1}w_l\Big(\prod_{i\not=l}N_i\Big)\sum_{j=1}^{N_l}f({\bf z}^{(l)}_j)-(1-\tau)N_Tw_k\big(\prod_{i\not=k}N_i\big)\sum_{j=1}^{n-1}f({\bf z}_j^{(k)})\nonumber
     \end{align}
     \begin{align}\label{eq:Sn2.M}
     &-(1-\tau)N_T\mathrm{E}^{(S)}\left\{\sum_{l=k+1}^Kw_l
     \Big(\prod_{i\not=l}N_i\Big)\sum_{j=1}^{N_l}f({\bf z}^{(l)}_j)+w_k\Big(\prod_{i\not=k}N_i\Big)\sum_{j=n}^{N_k}f({\bf z}^{(k)}_j)\right\}\nonumber\\
    =&(1-\tau)N_Tw_k\Big(\prod_{i\not=k}N_i\Big)\left(f({\bf z}_n^{(k)})-\mathrm{E}^{(S_k)}f\right).
\end{align}

Moreover, we define another random variable:
\begin{align*}%\label{eq:Sn1.C}
    T_n:=&\mathrm{E}^{(T)}\left\{F_T({\bf \overline{Z}}_{1}^{N_T})|{\bf \overline{Z}}_1^n\right\},\; \mbox{$0\leq n\leq N_T$,}
\end{align*}
where
\begin{align*}
{\bf\overline{Z}}_{1}^n=\{{\bf z}^{(T)}_1,\cdots,{\bf z}^{(T)}_n\}\subseteq{\bf \overline{Z}}_{1}^{N_{T}} \; \mbox{  with   } \;  {\bf \overline{Z}}_{1}^0:=\varnothing.
\end{align*}
It is clear that $T_0=\mathrm{E}^{(T)}F_T$ and   $T_{N_T}=F_T({\bf\overline{Z}}_{1}^{N_T})$.
Similarly, we also have for any $1\leq n\leq N_T$,
\begin{align}\label{eq:Sn2.C}
T_n-T_{n-1}
=\tau \Big(\prod_{k=1}^KN_k\Big)\left(f({\bf z}_n^{(T)})-\mathrm{E}^{(T)}f\right).
\end{align}

\subsubsection{Proof of Theorem \ref{thm:Hdineq.MC}}

In order to prove Theorem \ref{thm:Hdineq.MC}, we need the following inequality resulted from Hoeffding's lemma.

\begin{lemma}\label{lem:Hoeffding}
Let $f$ be a function with the range $[a,b]$. Then, the following holds for any $\alpha>0$:
\begin{align*}%\label{eq:Hoeffding}
&\mathrm{E}
\left\{\mathrm{e}^{\alpha(f({\bf z}^{(S)})-\mathrm{E}^{(S)}f)}\right\}
\leq \mathrm{e}^{\frac{\alpha^2(b-a)^2}{8}}.
\end{align*}
\end{lemma}

{\it Proof.} We consider
$$(f({\bf z}^{(S)})-\mathrm{E}^{(S)}f)$$
as a random variable. Then, it is clear that
$$\mathrm{E}\{f({\bf z}^{(S)})-\mathrm{E}^{(S)}f\}=0.$$
Since the value of $\mathrm{E}^{(S)}f$ is a constant denoted as $e$, we have
$$a-e\leq f({\bf z}^{(S)})-\mathrm{E}^{(S)}f\leq b-e.$$
According to Hoeffding's lemma, we then have
\begin{align*}%\label{eq:Hoeffding1}
&\mathrm{E}
\left\{\mathrm{e}^{\alpha(f({\bf z}^{(S)})-\mathrm{E}^{(S)}f)}\right\}
\leq \mathrm{e}^{\frac{\alpha^2(b-a)^2}{8}}.
\end{align*}
 This completes the proof. \hfill $\blacksquare$

We are now ready to prove Theorem \ref{thm:Hdineq.MC}.

{\it Proof of Theorem \ref{thm:Hdineq.MC}.}
According to \eqref{eq:F.MC} and \eqref{eq:F1.MC}, we have
\begin{align*}%\label{eq:main0.C}
F^{\tau}_{\bf w}\big(\{{\bf Z}_{1}^{N_k}\}_{k=1}^K,\overline{{\bf Z}}_{1}^{N_{T}}\big)-\mathrm{E}^{(*)}F^{\tau}_{\bf w}
    =&F_{\bf w}\big(\{{\bf Z}_{1}^{N_k}\}_{k=1}^K\big)+F_T({\bf \overline{Z}}_{1}^{N_T})-
    \mathrm{E}^{(*)}\{F_{\bf w}+F_T\}\nonumber\\
    =& F_{\bf w}\big(\{{\bf Z}_{1}^{N_k}\}_{k=1}^K\big)-\mathrm{E}^{(S)}F_{\bf w}+F_T({\bf \overline{Z}}_{1}^{N_T})-\mathrm{E}^{(T)}F_T,
    \end{align*}
where the expectation $\mathrm{E}^{(S)}$ is taken on all sources $\{\mathcal{Z}^{(S_k)}\}_{k=1}^K$ and $\mathrm{E}^{(T)}$ is taken on the target domain $\mathcal{Z}^{(T)}$.

According to \eqref{eq:F.MC}, \eqref{eq:Sn2.M}, \eqref{eq:Sn2.C}, Lemma \ref{lem:Hoeffding}, Markov's inequality and the law of iterated expectation, we have for any $\alpha>0$,
\begin{align}\label{eq:Hmain1.MC}
   &\mathrm{Pr}\left\{F^{\tau}_{\bf w}\big(\{{\bf Z}_{1}^{N_k}\}_{k=1}^K,\overline{{\bf Z}}_{1}^{N_{T}}\big)-\mathrm{E}^{(*)}F^{\tau}_{\bf w}>\xi\right\}\nonumber\\
   =&\mathrm{Pr}\left\{ F_{\bf w}\big(\{{\bf Z}_{1}^{N_k}\}_{k=1}^K\big)-\mathrm{E}^{(S)}F_{\bf w}+F_T({\bf \overline{Z}}_{1}^{N_T})-\mathrm{E}^{(T)}F_T>\xi\right\}\nonumber\\
   \leq& \mathrm{e}^{-\alpha\xi}
\mathrm{E}\left\{\mathrm{e}^{\alpha\left(F_{\bf w}\big(\{{\bf Z}_{1}^{N_k}\}_{k=1}^K\big)-\mathrm{E}^{(S)}F_{\bf w}+F_T({\bf \overline{Z}}_{1}^{N_T})-\mathrm{E}^{(T)}F_T\right)}\right\}\nonumber\\
=&\mathrm{e}^{-\alpha\xi}
\mathrm{E}\left\{\mathrm{e}^{\alpha\big(\sum_{k=1}^K\sum_{n=1}^{N_k}
(S^{(k)}_{n}-S^{(k)}_{n-1})
+\sum_{n=1}^{N_T}(T_n-T_{n-1})\big)}\right\}\nonumber\\
=&\mathrm{e}^{-\alpha\xi}\mathrm{E}\left\{\mathrm{E}
\left\{\mathrm{e}^{\alpha\big(\sum_{k=1}^K\sum_{n=1}^{N_k}
(S^{(k)}_{n}-S^{(k)}_{n-1})
+\sum_{n=1}^{N_T}(T_n-T_{n-1})\big)}\big|{\bf Z}_{1}^{N_1},\cdots,{\bf Z}_{1}^{N_{K-1}},{\bf Z}_{1}^{N_{K}-1}\right\}\right\}\nonumber\\
=&\mathrm{e}^{-\alpha\xi}\mathrm{E}\Big\{\mathrm{e}^
{\alpha\big(\sum_{k=1}^{K}\sum_{n=1}^{N_k}
(S^{(k)}_{n}-S^{(k)}_{n-1})-
(S^{(K)}_{N_K}-S^{(K)}_{N_K-1})+\sum_{n=1}^{N_T}(T_n-T_{n-1})\big)}\nonumber\\
&\times\mathrm{E}
\Big\{\mathrm{e}^{\alpha\big(S^{(K)}_{N_K}-S^{(K)}_{N_K-1}\big)}
\big|{\bf Z}_{1}^{N_1},\cdots,{\bf Z}_{1}^{N_{K-1}},{\bf Z}_{1}^{N_{K}-1}\Big\}\Big\}\nonumber\\
=&\mathrm{e}^{-\alpha\xi}\mathrm{E}\left\{\mathrm{e}^
{\alpha\big(\sum_{k=1}^{K}\sum_{n=1}^{N_k}
(S^{(k)}_{n}-S^{(k)}_{n-1})-
(S^{(K)}_{N_K}-S^{(K)}_{N_K-1})
+\sum_{n=1}^{N_T}(T_n-T_{n-1})\big)}\mathrm{E}
\Big\{\mathrm{e}^{\alpha w_KN_T(\prod_{i\not=K}N_i)(f({\bf z}_N^{(K)})-\mathrm{E}^{(S_K)}f)}\Big\}\right\}\nonumber\\
\leq&\mathrm{e}^{-\alpha\xi}\mathrm{E}\left\{\mathrm{e}^
{\alpha\big(\sum_{k=1}^{K}\sum_{n=1}^{N_k}
(S^{(k)}_{n}-S^{(k)}_{n-1})-
(S^{(K)}_{N_K}-S^{(K)}_{N_K-1})+\sum_{n=1}^{N_T}(T_n-T_{n-1})\big)}\right\}\nonumber\\
&\times\exp\left\{\frac{\alpha^2(1-\tau)^2N_T^2w_K^2(\prod_{i\not=K}N_i)^2
(b-a)^2}{8}\right\}\nonumber\\
=&\mathrm{e}^{-\alpha\xi}\mathrm{E}\left\{\mathrm{e}^
{\alpha\big(\sum_{k=1}^{K}\sum_{n=1}^{N_k}
(S^{(k)}_{n}-S^{(k)}_{n-1})-
(S^{(K)}_{N_K}-S^{(K)}_{N_K-1})+\sum_{n=1}^{N_T-1}(T_n-T_{n-1})\big)}\mathrm{E}
\Big\{\mathrm{e}^{\alpha\left(T_{N_T}-T_{N_T-1}\right)}\Big|\overline{{\bf Z}}_1^{N_T-1}\Big\}\right\}\nonumber\\
&\times\exp\left\{\frac{\alpha^2(1-\tau)^2N_T^2w_K^2(\prod_{i\not=K}N_i)^2
(b-a)^2}{8}\right\}\nonumber\\
\leq&\mathrm{e}^{-\alpha\xi}\mathrm{E}\Big\{\mathrm{e}^
{\alpha\big(\big(\sum_{k=1}^{K}\sum_{n=1}^{N_k}
(S^{(k)}_{n}-S^{(k)}_{n-1})-
(S^{(K)}_{N_K}-S^{(K)}_{N_K-1})\big)
+\sum_{n=1}^{N_T-1}(T_n-T_{n-1})\big)}\Big\}\nonumber\\
&\times\exp\left\{\frac{\alpha^2\tau^2(\prod_{k=1}^KN_k)^2(b-a)^2}{8}\right\}
\exp\left\{\frac{\alpha^2(1-\tau)^2N_T^2w_K^2(\prod_{i\not=K}N_i)^2
(b-a)^2}{8}\right\},
\end{align}
where ${\bf Z}_1^{N_K-1}:=\{{\bf z}_1^{(K)},\cdots,{\bf z}_{N_K-1}^{(K)}\}\subset{\bf Z}_1^{N_K}$ and $\overline{{\bf Z}}_1^{N_T-1}:=\{{\bf z}_1^{(T)},\cdots,{\bf z}_{N_T-1}^{(T)}\}\subset\overline{{\bf Z}}_1^{N_T}$.

%%%%%%%%%%%%%%%%%%%%%%%%

Following \eqref{eq:Hmain1.MC}, we have
\begin{equation}\label{eq:Hmain2.MC}
   \mathrm{Pr}\left\{F^{\tau}_{\bf w}\big(\{{\bf Z}_{1}^{N_k}\}_{k=1}^K,\overline{{\bf Z}}_{1}^{N_{T}}\big)-\mathrm{E}^{(*)}F^{\tau}_{\bf w}>\xi\right\}
   \leq\mathrm{e}^{\Phi(\alpha)-\alpha\xi},
\end{equation}
where
\begin{equation}\label{eq:Hmain3.MC}
   \Phi(\alpha)=N_T\frac{\alpha^2\tau^2
(\prod_{k=1}^KN_k)^2(b-a)^2}{8}+\sum_{k=1}^KN_k
\frac{\alpha^2(1-\tau)^2N_T^2w_k^2(\prod_{i\not=k}N_i)^2(b-a)^2}{8}.
\end{equation}
Similarly, we can obtain
\begin{align}\label{eq:Hmain4.MC}
    \mathrm{Pr}\left\{\mathrm{E}^{(*)}F^{\tau}_{\bf w}-F^{\tau}_{\bf w}\big(\{{\bf Z}_{1}^{N_k}\}_{k=1}^K,\overline{{\bf Z}}_{1}^{N_{T}}\big)>\xi\right\} \leq\mathrm{e}^{\Phi(\alpha)-\alpha\xi}.
\end{align}
Note that $\Phi(\alpha)-\alpha\xi$ is a quadratic function with respect to $\alpha>0$ and thus the minimum value $\min_{\alpha>0}\left\{\Phi(\alpha)-\alpha\xi\right\}$ is achieved
when
\begin{equation}\label{eq:Hmain5.MC}
   \alpha=\frac{4\xi}{N_T(b-a)^2\big(\prod_{k=1}^KN_k\big)
\Big(\tau^2\big(\prod_{k=1}^KN_k\big)+
\sum_{k=1}^{K}(1-\tau)^2N_Tw_k^2\big(\prod_{i\not=k}N_i\big)\Big)}.
\end{equation}
By combining \eqref{eq:Hmain2.MC}, \eqref{eq:Hmain3.MC}, \eqref{eq:Hmain4.MC} and \eqref{eq:Hmain5.MC}, we arrive at
\begin{align*}%\label{eq:deviation3}
    &\mathrm{Pr}\left\{\big|\mathrm{E}^{(*)}F^{\tau}_{\bf w}-F^{\tau}_{\bf w}\big(\{{\bf Z}_{1}^{N_k}\}_{k=1}^K,\overline{{\bf Z}}_{1}^{N_{T}}\big)\big|>\xi\right\}\nonumber\\
    \leq&2\exp\left\{-\frac{2\xi^2}
{N_T(b-a)^2\big(\prod_{k=1}^KN_k\big)
\Big(\tau^2\big(\prod_{k=1}^KN_k\big)+
\sum_{k=1}^{K}(1-\tau)^2N_Tw_k^2\big(\prod_{i\not=k}N_i\big)\Big)}
\right\}.
\end{align*}
This completes the proof.  \hfill $\blacksquare$

%%%

\subsection{Proof of Theorem \ref{thm:dineq.MC}}

To prove Theorem \ref{thm:dineq.MC}, we also need the following two inequalities. The first one has been mentioned in the proof of the classical Bennett's inequality \citep{Bennett62}.

\begin{lemma}\label{lem:Bennett}
Let $f$ be a function with the range $[a,b]$. Then, the following holds for any $\alpha>0$:
\begin{align*}%\label{eq:Bennett}
&\mathrm{E}
\left\{\mathrm{e}^{\alpha(f({\bf z})-\mathrm{E}f)}\right\}
\leq \exp\left\{\mathrm{e}^{\alpha(b-a)}-1-\alpha(b-a)\right\}.
\end{align*}
\end{lemma}

{\it Proof.} We consider
$$Z:=f({\bf z})-\mathrm{E}f$$
as a random variable. Then, it is clear that
$\mathrm{E}Z=0$, $|Z|\leq b-a$ and $\mathrm{E}Z^2\leq (b-a)^2$.

For any $\alpha>0$, we expand
\begin{align*}
    \mathrm{E}\mathrm{e}^{\alpha Z}=&\mathrm{E}\left\{\sum_{s=0}^{\infty}\frac{(\alpha Z)^s}{s!}\right\}
    =\sum_{s=0}^{\infty}\alpha^s\frac{\mathrm{E}Z^s}{s!}\nonumber\\
    =&1+\sum_{s=2}^{\infty}\frac{\alpha^s}{s!}\mathrm{E}Z^2Z^{s-2}
    \leq 1+\sum_{s=2}^{\infty}\frac{\alpha^s}{s!}\mathrm{E}(b-a)^2(b-a)^{s-2}
    \nonumber\\
    =&1+\frac{(b-a)^2}{(b-a)^2}\sum_{s=2}^{\infty}\frac{\alpha^s(b-a)^s}{s!}
    =1+\left(\mathrm{e}^{\alpha(b-a)}-1-\alpha(b-a)\right)\nonumber\\
    \leq&\exp\left\{\mathrm{e}^{\alpha(b-a)}-1-\alpha(b-a)\right\}.
\end{align*}
This completes the proof.\hfill $\blacksquare$

The second lemma is given as follows:

\begin{lemma}\label{lem:sec.lem}
Let $h(x)=\mathrm{e}^x-1-x$ ($x\geq0$) and ${\bf w}=(w_1,\cdots,w_K)\in[0,1]^K$. Given any $\{N_k\}_{k=1}^K\in\mathbb{N}^K$, the solution to the following optimization problem:
\begin{equation}\label{eq:sec.lem1}
 \min_{{\bf w}\in[0,1]^K}
 \left\{\sum_{k=1}^KN_kh\Big(w_k\big(\prod_{i\not=k}N_i\big)\Big)\right\}
 \;\;s.t.\;\sum_{k=1}^Kw_k=1
\end{equation}
is given by: for any $1\leq k\leq K$,
\begin{equation*}%\label{eq:sec.lem2}
 w_k=\frac{N_k}{N_1+N_2+\cdots+N_K}.
\end{equation*}
\end{lemma}

{\it Proof.} The method of Lagrange multipliers is applied to solve this optimization problem.
In fact, we introduce a new variable $\lambda$ to form a
Lagrange function:
\begin{equation*}%\label{eq:form1}
 F({\bf w},\lambda)=\left\{\sum_{k=1}^KN_kh\Big(w_k\big
 (\prod_{i\not=k}N_i\big)\Big)\right\}+\lambda(\sum_{k=1}^Kw_k-1),
\end{equation*}
and then solve the equation
\begin{equation}\label{eq:form2}
 \nabla_{{\bf w},\lambda} F({\bf w},\lambda)=0,
\end{equation}
whose solution is also the solution to the optimization problem \eqref{eq:sec.lem1}.

From \eqref{eq:form2}, we have
\begin{equation*}%\label{eq:form3}
    \left\{
      \begin{array}{ll}
        \frac{\partial F}{\partial w_1}= & \big(\prod_{k=1}^KN_k\big)
    h'\Big(w_1\big(\prod_{i\not=1}N_i\big)\Big)+\lambda=0 \\
        \frac{\partial F}{\partial w_2}= & \big(\prod_{k=1}^KN_k\big)
    h'\Big(w_2\big(\prod_{i\not=2}N_i\big)\Big)+\lambda=0 \\
        \quad\vdots & \\
        \frac{\partial F}{\partial w_K}= & \big(\prod_{k=1}^KN_k\big)
    h'\Big(w_K\big(\prod_{i\not=K}N_i\big)\Big)+\lambda=0 \\
        \frac{\partial F}{\partial \lambda}= &\sum_{k=1}^Kw_k-1=0,
      \end{array}
    \right.
\end{equation*}
and thus
\begin{equation*}%\label{eq:form4}
h'\Big(w_1\big(\prod_{i\not=1}N_i\big)\Big)
=h'\Big(w_2\big(\prod_{i\not=2}N_i\big)\Big)
=\cdots
=h'\Big(w_K\big(\prod_{i\not=K}N_i\big)\Big)
\end{equation*}
with $\sum_{k=1}^Kw_k-1=0$. Since $h'(x)=\mathrm{e}^x-1$ is a strictly monotonic increasing function with $h'(x)>0$ for any $x>0$, we further have
\begin{equation}\label{eq:form5}
w_1\big(\prod_{i\not=1}N_i\big)
=w_2\big(\prod_{i\not=2}N_i\big)
=\cdots
=w_K\big(\prod_{i\not=K}N_i\big)
\end{equation}
with $\sum_{k=1}^Kw_k-1=0$. According to \eqref{eq:form5}, we obtain the solution to the optimization problem \eqref{eq:sec.lem1}: for any $1\leq k\leq K$,
\begin{align*}%\label{eq:sec.lem9}
    w_k=\frac{N_k}{N_1+N_2+\cdots+N_K}.
\end{align*}
This completes the proof.
\hfill $\blacksquare$

We are now ready to prove Theorem \ref{thm:dineq.MC}.

{\it Proof of Theorem \ref{thm:dineq.MC}.}
Similar to the proof of Theorem \ref{thm:Hdineq.MC}, according to \eqref{eq:F.MC}, \eqref{eq:Sn2.M}, \eqref{eq:Sn2.C}, Lemma \ref{lem:Bennett}, Markov's inequality and the law of iterated expectation, we have for any $\alpha>0$,
\begin{align}\label{eq:main1.MC}
   &\mathrm{Pr}\left\{F^{\tau}_{\bf w}\big(\{{\bf Z}_{1}^{N_k}\}_{k=1}^K,\overline{{\bf Z}}_{1}^{N_{T}}\big)-\mathrm{E}^{(*)}F^{\tau}_{\bf w}>\xi\right\}\nonumber\\
\leq&\mathrm{e}^{-\alpha\xi}\mathrm{E}\Big\{\mathrm{e}^
{\alpha\big(\big(\sum_{k=1}^{K}\sum_{n=1}^{N_k}
(S^{(k)}_{n}-S^{(k)}_{n-1})-
(S^{(K)}_{N_K}-S^{(K)}_{N_K-1})\big)+\sum_{n=1}^{N_T-1}(T_n-T_{n-1})\big)}\Big\}\nonumber\\
&\times\exp\left\{\mathrm{e}^{\alpha\tau(\prod_{k=1}^KN_k)(b-a)}-1
-\alpha\tau\Big(\prod_{k=1}^KN_k\Big)(b-a)\right\}\nonumber\\
&\times\exp\left\{\mathrm{e}^{\alpha(1-\tau)N_Tw_K(\prod_{i\not=K}N_i)(b-a)}
-1-\alpha(1-\tau)N_Tw_K\Big(\prod_{i\not=K}N_i\Big)(b-a)\right\}.
\end{align}

Following \eqref{eq:main1.MC}, we arrive at
\begin{equation}\label{eq:main2.MC}
   \mathrm{Pr}\left\{F^{\tau}_{\bf w}\big(\{{\bf Z}_{1}^{N_k}\}_{k=1}^K,\overline{{\bf Z}}_{1}^{N_{T}}\big)-\mathrm{E}^{(*)}F^{\tau}_{\bf w}>\xi\right\}
   \leq\mathrm{e}^{\Phi(\alpha)-\alpha\xi},
\end{equation}
where
\begin{equation}\label{eq:main3.MC}
   \Phi(\alpha)=\Psi(\alpha)+\sum_{k=1}^K\Upsilon_k(\alpha)
\end{equation}
with
\begin{align*}%\label{eq:Psi.MC}
    \Psi(\alpha)
=N_T\left(\mathrm{e}^{\alpha\tau(\prod_{k=1}^KN_k)(b-a)}-1
-\alpha\tau\Big(\prod_{k=1}^KN_k\Big)(b-a)\right),
\end{align*}
and for any $1\leq k\leq K$,
\begin{align}\label{eq:Upsilon.MC}
    \Upsilon_k(\alpha)=
N_k\left(\mathrm{e}^{\alpha(1-\tau)N_Tw_k(\prod_{i\not=k}N_i)(b-a)}
-1-\alpha(1-\tau)N_Tw_k\Big(\prod_{i\not=k}N_i\Big)(b-a)\right).
\end{align}
Note that the value of $\Phi(\alpha)$ is determined by $\alpha$ and the choices of ${\bf w}$ and $\tau$. We first minimize $\Phi(\alpha)$ with respect to ${\bf w}$ and $\tau$.
According to Lemma \ref{lem:sec.lem} and \eqref{eq:Upsilon.MC}, under the condition that $\sum_{k=1}^Kw_k=1$, we have
\begin{align}\label{eq:Upsilon.MC1}
   \widetilde{\Upsilon}(\alpha):=&\min_{{\bf w}\in[0,1]^K}\left\{\sum_{k=1}^K\Upsilon_k(\alpha)\right\}\nonumber\\
=&\sum_{k=1}^K
N_k\left(\mathrm{e}^{\alpha(1-\tau)\frac{N_T\prod_{k=1}^KN_k}
{\sum_{k=1}^KN_K}(b-a)}
-1-\alpha(1-\tau)\frac{N_T\prod_{k=1}^KN_k}
{\sum_{k=1}^KN_K}(b-a)\right),
\end{align}
which is achieved when $w_k=N_k/\sum_{k=1}^KN_k$ ($1\leq k\leq K$).

Again, by Lemma \ref{lem:sec.lem} and \eqref{eq:Upsilon.MC1},
setting
 \begin{equation*}
    \tau=\frac{N_T}{N_T+\sum_{k=1}^KN_k}
 \end{equation*}
leads to
\begin{align*}%\label{eq:Psi.MC1}
    \widetilde{\Phi}(\alpha):=\mathop{\min_{{\bf w}\in[0,1]^K}}_{\tau\in(0,1]}\left\{\Phi(\alpha)\right\}
=
\Big(N_T+\sum_{k=1}^KN_k\Big)\left(\mathrm{e}^{\alpha(b-a)
\frac{N_T\prod_{k=1}^KN_k}
{N_T+\sum_{k=1}^KN_K}}
-1-\alpha(b-a)\frac{N_T\prod_{k=1}^KN_k}
{N_T+\sum_{k=1}^KN_k}\right).
\end{align*}
We are now ready to minimize $\widetilde{\Phi}(\alpha)-\alpha\xi$ with respect to $\alpha$.
Note that $\widetilde{\Phi}(\alpha)$ is infinitely differentiable for $\alpha>0$ with
\begin{align}\label{eq:Hn1.MC}
\widetilde{\Phi}'(\alpha)
=(b-a)N_T\Big(\prod_{k=1}^KN_k\Big)
\left(\mathrm{e}^{\alpha(b-a)\frac{N_T\prod_{k=1}^KN_k}
{N_T+\sum_{k=1}^KN_k}}-1\right)>0,
\end{align}
and
\begin{align}\label{eq:Hn2.MC}
    \widetilde{\Phi}''(\alpha)=\frac{(b-a)^2N_T^2\Big(\prod_{k=1}^KN_k\Big)^2}
{N_T+\sum_{k=1}^KN_k}
\mathrm{e}^{\alpha(b-a)\frac{N_T\prod_{k=1}^KN_k}
{N_T+\sum_{k=1}^KN_k}}>0.
\end{align}

Denote $\varphi(\alpha):=\widetilde{\Phi}'(\alpha)$. According to \eqref{eq:Hn1.MC} and \eqref{eq:Hn2.MC}, for any $\xi>0$, the minimum
$\min_{\alpha>0}\left\{\widetilde{\Phi}(\alpha)-\alpha\xi\right\}$ is achieved
when $\varphi(\alpha)-\xi=0$.
By \eqref{eq:Hn1.MC}, we have $\varphi(0)=0$ and $\varphi^{-1}(0)=0$.
Since $\widetilde{\Phi}(0)=0$, we arrive at
\begin{align*}%\label{eq:Hn3}
\widetilde{\Phi}\left(\varphi^{-1}(\xi)\right)
    =&\int_{0}^{\varphi^{-1}(\xi)}\varphi(s)ds\nonumber\\
    =&\int_{0}^{\xi}sd\varphi^{-1}(s)\nonumber\\
=&\xi\varphi^{-1}(\xi)-0\varphi^{-1}(0)
-\int_{0}^{\xi}\varphi^{-1}(s)ds\nonumber\\
=&\xi\varphi^{-1}(\xi)-\int_{0}^{\xi}\varphi^{-1}(s)ds.
\end{align*}
Thus, we have for any $\xi>0$,
\begin{align}\label{eq:main6.MC}
    \min_{\alpha>0}\left\{\widetilde{\Phi}(\alpha)-\alpha\xi\right\}
    =&-\int_{0}^{\xi}\varphi^{-1}(s)ds\nonumber\\
    =&-\int_{0}^{\xi}
\frac{N_T+\sum_{k=1}^KN_k}{N_T\big(\prod_{k=1}^KN_k\big)(b-a)}
\ln\left(1+\frac{s}{ N_T\big(\prod_{k=1}^KN_k\big)(b-a)}
\right)ds\nonumber\\
=&\Big(N_T+\sum_{k=1}^KN_k\Big)\Gamma\left( \frac{\xi}{ N_T\big(\prod_{k=1}^KN_k\big)(b-a)}\right).
\end{align}
By combining \eqref{eq:main1.MC}, \eqref{eq:main2.MC}, \eqref{eq:main3.MC} and \eqref{eq:main6.MC}, if $w_k=\frac{N_k}{\sum_{k=1}^KN_k}$ ($1\leq k\leq K$) and $\tau=\frac{N_T}{N_T+\sum_{k=1}^KN_k}$,
 we have
\begin{align*}%\label{eq:cor.dev2}
    &\mathrm{Pr}\left\{F^{\tau}_{\bf w}\big(\{{\bf Z}_{1}^{N_k}\}_{k=1}^K,\overline{{\bf Z}}_{1}^{N_{T}}\big)-\mathrm{E}^{(*)}F^{\tau}_{\bf w}>\xi\right\}\nonumber\\
\leq&\exp\left\{\Big(N_T+\sum_{k=1}^KN_k\Big)\Gamma\left( \frac{\xi}{ N_T\big(\prod_{k=1}^KN_k\big)(b-a)}\right)\right\},
\end{align*}
where $\Gamma(x)$ is defined in \eqref{eq:Gamma}.
Similarly, under the same conditions, we also have
\begin{align}\label{eq:cor.dev3}
    &\mathrm{Pr}\left\{\mathrm{E}^{(*)}F^{\tau}_{\bf w}-F^{\tau}_{\bf w}\big(\{{\bf Z}_{1}^{N_k}\}_{k=1}^K,\overline{{\bf Z}}_{1}^{N_{T}}\big)>\xi\right\}\nonumber\\
\leq&\exp\left\{\Big(N_T+\sum_{k=1}^KN_k\Big)\Gamma\left( \frac{\xi}{ N_T\big(\prod_{k=1}^KN_k\big)(b-a)}\right)\right\}.
\end{align}
This completes the proof. \hfill $\blacksquare$

%%%%%%%%%

\subsection{Proof of Theorem \ref{thm:Mcdiarmid}}

{\it Proof of Theorem \ref{thm:Mcdiarmid}.} Define a random variable
\begin{equation}\label{eq:Tn1}
    T^{(k)}_{n}:=\mathrm{E}\left\{H(\{{\bf Z}_{1}^{N_k}\}_{k=1}^K)|{\bf Z}_{1}^{N_1},{\bf Z}_{1}^{N_2},\cdots,{\bf Z}_{1}^{N_{k-1}},{\bf Z}_1^{n}\right\},\;\; 1\leq k\leq K,\;0\leq n\leq N_k,
\end{equation}
where
$${\bf Z}_{1}^{n}=\{{\bf z}^{(k)}_1,{\bf z}^{(k)}_2,\cdots,{\bf z}^{(k)}_{n}\}\subseteq{\bf Z}_{1}^{N_k},  \mbox{  and   }   {\bf Z}_{1}^0=\varnothing.$$
It is clear that
$$T_0^{(1)}=\mathrm{E}\big\{H(\{{\bf Z}_{1}^{N_k}\}_{k=1}^K)\big\} \mbox{  and }  T^{(K)}_{N_K}=H(\{{\bf Z}_{1}^{N_k}\}_{k=1}^K),$$
and thus
\begin{equation}\label{eq:Tn2}
 H(\{{\bf Z}_{1}^{N_k}\}_{k=1}^K)-\mathrm{E}\big\{H(\{{\bf Z}_{1}^{N_k}\}_{k=1}^K)\big\}
 =T^{(K)}_{N_K}-T_0^{(1)}
 =\sum_{k=1}^K\sum_{n=1}^{N_k}(T^{(k)}_{n}-T^{(k)}_{n-1}).
\end{equation}

Denote for any $1\leq k\leq K$ and $1\leq n\leq N_k$,
\begin{align}\label{eq:UL1}
 U_n^{(k)}=&\sup_{\mu}\left\{T^{(k)}_{n}\big|_{{\bf z}^{(k)}_n=\mu}-T^{(k)}_{n-1}\right\};\nonumber\\
 L_n^{(k)}=&\inf_{\nu}\left\{T^{(k)}_{n}\big|_{{\bf z}^{(k)}_n=\nu}-T^{(k)}_{n-1}\right\}.\nonumber
\end{align}
It follows from the definition of \eqref{eq:Tn1} that $L_n^{(k)}\leq(T^{(k)}_{n}-T^{(k)}_{n-1})\leq U_n^{(k)}$ and thus results in
\begin{equation}\label{eq:UL2}
T^{(k)}_{n}-T^{(k)}_{n-1}\leq U_n^{(k)}-L_n^{(k)}=\sup_{\mu,\nu}\left\{T^{(k)}_{n}\big|_{{\bf z}^{(k)}_n=\mu}-T^{(k)}_{n}\big|_{{\bf z}^{(k)}_n=\nu}\right\}\leq c^{(k)}_n.
\end{equation}
Moreover, by the law of iterated expectation, we also have for any $1\leq k\leq K$ and $1\leq n\leq N_k$
\begin{equation}\label{eq:UL3}
\mathrm{E}\left\{T^{(k)}_{n}-T^{(k)}_{n-1}|{\bf Z}_{1}^{N_1},{\bf Z}_{1}^{N_2},\cdots,{\bf Z}_{1}^{N_{k-1}},{\bf Z}_1^{n-1}\right\}=0.
\end{equation}
According to Hoeffding inequality \citep[see][]{Hoeffding63}, given an $\alpha>0$, the condition \eqref{eq:dif.cond.MC} leads to for any $1\leq k\leq K$ and $1\leq n\leq N_k$,
\begin{equation}\label{eq:Tn3}
    \mathrm{E}\left\{\mathrm{e}^{\alpha(T^{(k)}_{n}-T^{(k)}_{n-1})}|{\bf Z}_{1}^{N_1},{\bf Z}_{1}^{N_2},\cdots,{\bf Z}_{1}^{N_{k-1}},{\bf Z}_1^{n-1}\right\}\leq\mathrm{e}^{\alpha^2(c^{(k)}_n)^2/8}.
\end{equation}

Subsequently, according to Markov's inequality, \eqref{eq:Tn2}, \eqref{eq:UL2}, \eqref{eq:UL3} and \eqref{eq:Tn3},  we have for any $\alpha>0$,
\begin{align*}%\label{eq:Main.McD}
&\mathrm{Pr}\left\{H\big({\bf Z}_{1}^{N_1},\cdots,{\bf Z}_{1}^{N_K}\big)-\mathrm{E}\Big\{H\big({\bf Z}_{1}^{N_1},\cdots,{\bf Z}_{1}^{N_K}\big)\Big\}\geq\xi\right\}\nonumber\\
\leq&\mathrm{e}^{-\alpha\xi}\mathrm{E}\Big\{\mathrm{e}^{\alpha\big(H\big({\bf Z}_{1}^{N_1},\cdots,{\bf Z}_{1}^{N_K}\big)-\mathrm{E}\big\{H\big({\bf Z}_{1}^{N_1},\cdots,{\bf Z}_{1}^{N_K}\big)\big\}\big)}\Big\}\nonumber\\
=&\mathrm{e}^{-\alpha\xi}\mathrm{E}
\Big\{\mathrm{e}^{\alpha\big(\sum_{k=1}^K\sum_{n=1}^{N_k}(T^{(k)}_{n}
-T^{(k)}_{n-1})\big)}\Big\}\nonumber\\
=&\mathrm{e}^{-\alpha\xi}\mathrm{E}\left\{\mathrm{E}
\left\{\mathrm{e}^{\alpha
\big(\sum_{k=1}^K\sum_{n=1}^{N_k}(T^{(k)}_{n}
-T^{(k)}_{n-1})\big)}|{\bf Z}_{1}^{N_1},\cdots,{\bf Z}_{1}^{N_{k-1}},{\bf Z}_1^{N_K-1}\right\} \right\}\nonumber\\
=&\mathrm{e}^{-\alpha\xi}\mathrm{E}\Big\{\mathrm{e}^{\alpha\big(\sum_{k=1}^{K-1}
\sum_{n=1}^{N_k}(T^{(k)}_{n}
-T^{(k)}_{n-1})+\sum_{n=1}^{N_K-1}(T^{(K)}_{n}
-T^{(K)}_{n-1})\big)}\Big\{\mathrm{e}^{\alpha\big(T^{(K)}_{N_K}
-T^{(K)}_{N_K-1}\big)}|{\bf Z}_{1}^{N_1},\cdots,{\bf Z}_{1}^{N_{K-1}},{\bf Z}_1^{N_K-1}\Big\} \Big\}\nonumber\\
\leq&\mathrm{e}^{-\alpha\xi}\mathrm{E}\Big\{\mathrm{e}^{\alpha\big(\sum_{k=1}^{K-1}
\sum_{n=1}^{N_k}(T^{(k)}_{n}
-T^{(k)}_{n-1})+\sum_{n=1}^{N_K-1}(T^{(K)}_{n}
-T^{(K)}_{n-1})\big)}\Big\}\mathrm{e}^{\alpha^2(c^{(K)}_{N_K})^2/8} \nonumber\\
\leq & \mathrm{e}^{-\alpha\xi}\prod_{k=1}^K\prod_{n=1}^{N_k}
\exp\left\{\frac{\alpha^2(c^{(k)}_n)^2}{8}\right\}
=\exp\left\{-\alpha\xi+\alpha^2\sum_{k=1}^K\sum_{n=1}^{N_k}
\frac{(c^{(k)}_n)^2}{8}\right\}.
\end{align*}
The above bound is minimized by setting $$\alpha^*=\frac{4\xi}{\sum_{k=1}^K\sum_{n=1}^{N_K}(c^{(k)}_n)^2},$$
and its minimum value is
$$\exp\left\{-2\xi^2/\sum_{k=1}^{K}\sum_{n=1}^{N_k}(c^{(k)}_n)^2\right\}.$$

In the similar way, the proof of the inequality \eqref{eq:McDiarmid.Bennett} follows the way of proving Theorem \ref{thm:dineq.MC}, so we omit it.
This completes the proof. \hfill$\blacksquare$

%%%%%%
\subsection{Proof of Theorem \ref{thm:sym.MC}}\label{app:pf.sym}
%%%
{\it Proof of Theorem \ref{thm:sym.MC}.}
Let $\widehat{f}$ be the function achieving the supremum: $$\sup_{f\in\mathcal{F}}\big|\mathrm{E}^{(T)}f-\mathrm{E}^{\tau}_{\bf w}f\big|$$ with respect to the sample sets $\{{\bf Z}_1^{N_k}\}_{k=1}^K$ and ${\bf \overline{Z}}_{1}^{N_T}$. According to \eqref{eq:short1.MC}, \eqref{eq:short2.MC}, \eqref{eq:distance} and \eqref{eq:Dist.MC}, we arrive at
\begin{align*}%\label{eq:sym00.MC}
 &|\mathrm{E}^{(T)}\widehat{f}-\mathrm{E}^{\tau}_{\bf w}\widehat{f}|\nonumber\\
=&|\tau\mathrm{E}^{(T)}\widehat{f}+(1-\tau)\mathrm{E}^{(T)}\widehat{f}
-\tau\mathrm{E}^{(T)}_{N_T}\widehat{f}-(1-\tau)\mathrm{E}^{(S)}_{{\bf w}}\widehat{f}|\nonumber\\
=&|\tau\mathrm{E}^{(T)}\widehat{f}+(1-\tau)\mathrm{E}^{(T)}\widehat{f}
-\tau\mathrm{E}^{(T)}_{N_T}\widehat{f}
-(1-\tau)\mathrm{E}^{(S)}_{{\bf w}}\widehat{f}+(1-\tau)\overline{\mathrm{E}}^{(S)}\widehat{f}
-(1-\tau)\overline{\mathrm{E}}^{(S)}\widehat{f}|\nonumber\\
=&|\tau(\mathrm{E}^{(T)}\widehat{f}
-\mathrm{E}^{(T)}_{N_T}\widehat{f})
+(1-\tau)(\mathrm{E}^{(T)}\widehat{f}
-\overline{\mathrm{E}}^{(S)}\widehat{f})
+(1-\tau)(\overline{\mathrm{E}}^{(S)}\widehat{f}-\mathrm{E}^{(S)}_{{\bf w}}\widehat{f})|\nonumber\\
 \leq& (1-\tau)D_\mathcal{F}^{({\bf w})}(S,T)
+|\tau(\mathrm{E}^{(T)}\widehat{f}
-\mathrm{E}^{(T)}_{N_T}\widehat{f})
+(1-\tau)(\overline{\mathrm{E}}^{(S)}
\widehat{f}-\mathrm{E}^{(S)}_{{\bf w}}\widehat{f})|,
\end{align*}
and thus,
\begin{align}\label{eq:sym01.MC}
 &\mathrm{Pr}\left\{|\mathrm{E}^{(T)}\widehat{f}-\mathrm{E}^{\tau}_{\bf w}\widehat{f}|>\xi\right\}\nonumber\\
 \leq& \mathrm{Pr}\left\{(1-\tau)D_\mathcal{F}^{({\bf w})}(S,T)
+|\tau(\mathrm{E}^{(T)}\widehat{f}
-\mathrm{E}^{(T)}_{N_T}\widehat{f})
+(1-\tau)(\overline{\mathrm{E}}^{(S)}
\widehat{f}-\mathrm{E}^{(S)}_{{\bf w}}\widehat{f})|>\xi\right\},
\end{align}
where the expectation $\overline{\mathrm{E}}^{(S)}\widehat{f}$ is defined as
\begin{equation*}\label{eq:ESf.MC}
\overline{\mathrm{E}}^{(S)}\widehat{f}:=
\sum_{k=1}^Kw_k\mathrm{E}^{(S_k)}\widehat{f}.
\end{equation*}

Let
\begin{align}\label{eq:xi1.MC}
 \xi':=\xi-(1-\tau)D_\mathcal{F}^{({\bf w})}(S,T),
\end{align}
and denote $\wedge$ as the conjunction of two events. According to the triangle inequality, we have
\begin{align*}
&\Big(\big|\tau(\mathrm{E}^{(T)}\widehat{f}
-\mathrm{E}^{(T)}_{N_T}\widehat{f})
+(1-\tau)(\overline{\mathrm{E}}^{(S)}
\widehat{f}-\mathrm{E}^{(S)}_{{\bf w}}\widehat{f})\big|\nonumber\\
&-\big|\tau(\mathrm{E}^{(T)}\widehat{f}
-\mathrm{E'}^{(T)}_{N_T}\widehat{f})
+(1-\tau)(\overline{\mathrm{E}}^{(S)}
\widehat{f}-\mathrm{E'}^{(S)}_{{\bf w}}\widehat{f})\big| \Big)\nonumber\\
\leq&
\big|\tau(\mathrm{E'}^{(T)}_{N_T}\widehat{f}
-\mathrm{E}^{(T)}_{N_T}\widehat{f})
+(1-\tau)(\mathrm{E'}^{(S)}_{{\bf w}}\widehat{f}-\mathrm{E}^{(S)}_{{\bf w}}\widehat{f})\big|,
\end{align*}
and thus for any $\xi'>0$,
\begin{align*}%\label{eq:the:sym.pr1}
 &\left(\mathbf{1}_{|\tau(\mathrm{E}^{(T)}\widehat{f}
-\mathrm{E}^{(T)}_{N_T}\widehat{f})
+(1-\tau)(\overline{\mathrm{E}}^{(S)}
\widehat{f}-\mathrm{E}^{(S)}_{{\bf w}}\widehat{f})|>\xi'}\right)
\left(\mathbf{1}_{|\tau(\mathrm{E}^{(T)}\widehat{f}
-\mathrm{E'}^{(T)}_{N_T}\widehat{f})
+(1-\tau)(\overline{\mathrm{E}}^{(S)}
\widehat{f}-\mathrm{E'}^{(S)}_{{\bf w}}\widehat{f})|<\frac{\xi'}{2}}\right)\nonumber\\
=&\mathbf{1}_{\left\{|\tau(\mathrm{E}^{(T)}\widehat{f}
-\mathrm{E}^{(T)}_{N_T}\widehat{f})
+(1-\tau)(\overline{\mathrm{E}}^{(S)}
\widehat{f}-\mathrm{E}^{(S)}_{{\bf w}}\widehat{f})|>\xi'\right\}
\,\wedge\,\left\{|\tau(\mathrm{E}^{(T)}\widehat{f}
-\mathrm{E'}^{(T)}_{N_T}\widehat{f})
+(1-\tau)(\overline{\mathrm{E}}^{(S)}
\widehat{f}-\mathrm{E'}^{(S)}_{{\bf w}}\widehat{f})|<\frac{\xi'}{2}\right\}}\nonumber\\
\leq&\mathbf{1}_{|\tau(\mathrm{E'}^{(T)}_{N_T}\widehat{f}
-\mathrm{E}^{(T)}_{N_T}\widehat{f})
+(1-\tau)(\mathrm{E'}^{(S)}_{{\bf w}}\widehat{f}-\mathrm{E}^{(S)}_{{\bf w}}\widehat{f})|>\frac{\xi'}{2}}\;.
\end{align*}
Then, taking the expectation with respect to $\{{\bf Z'}_1^{N_k}\}_{k=1}^K$ and ${\bf \overline{Z'}}_{1}^{N_T}$ gives
\begin{align}\label{eq:the:sym.pr2.MC}
   &\left(\mathbf{1}_{|\tau(\mathrm{E}^{(T)}\widehat{f}
-\mathrm{E}^{(T)}_{N_T}\widehat{f})
+(1-\tau)(\overline{\mathrm{E}}^{(S)}
\widehat{f}-\mathrm{E}^{(S)}_{{\bf w}}\widehat{f})|>\xi'}\right)\nonumber\\
&\times\mathrm{Pr'}\left\{\big|\tau(\mathrm{E}^{(T)}\widehat{f}
-\mathrm{E'}^{(T)}_{N_T}\widehat{f})
+(1-\tau)(\overline{\mathrm{E}}^{(S)}
\widehat{f}-\mathrm{E'}^{(S)}_{{\bf w}}\widehat{f})\big|<\frac{\xi'}{2}\right\}\nonumber\\
\leq& \mathrm{Pr'}\left\{\big|\tau(\mathrm{E'}^{(T)}_{N_T}\widehat{f}
-\mathrm{E}^{(T)}_{N_T}\widehat{f})
+(1-\tau)(\mathrm{E'}^{(S)}_{{\bf w}}\widehat{f}-\mathrm{E}^{(S)}_{{\bf w}}\widehat{f})\big|>\frac{\xi'}{2}\right\}.
\end{align}

By Chebyshev's inequality, since $\{{\bf Z'}_1^{N_k}\}_{k=1}^K$ and ${\bf \overline{Z'}}_{1}^{N_T}$ are the sets of i.i.d. samples drawn from the multiple sources $\{\mathcal{Z}^{(S_k)}\}_{k=1}^K$ and the target $\mathcal{Z}^{(T)}$ respectively, we have for any $\xi'>0$,
\begin{align*}%\label{eq:the:sym.pr3.MC}
   &\mathrm{Pr'}\left\{\big|\tau(\mathrm{E}^{(T)}\widehat{f}
-\mathrm{E'}^{(T)}_{N_T}\widehat{f})
+(1-\tau)(\overline{\mathrm{E}}^{(S)}
\widehat{f}-\mathrm{E'}^{(S)}_{{\bf w}}\widehat{f})\big|\geq\frac{\xi'}{2}\right\}\nonumber\\
   \leq&\mathrm{Pr'}\left\{\frac{\tau}{N_T}\sum_{n=1}^{N_T}
|\mathrm{E}^{(T)}\widehat{f}-\widehat{f}({\bf z'}_n^{(T)})|+\sum_{k=1}^K\frac{(1-\tau)w_k}{N_k}
   \sum_{n=1}^{N_k}|\mathrm{E}^{(S_k)}\widehat{f}-\widehat{f}({\bf z'}_{n}^{(k)})|\geq\frac{\xi'}{2}\right\}\nonumber\\
   =&\mathrm{Pr'}\left\{\tau\Big(\prod_{k=1}^KN_k\Big)\sum_{n=1}^{N_T}
|\mathrm{E}^{(T)}\widehat{f}-\widehat{f}({\bf z'}_n^{(T)})|\right.\nonumber\\
&\left.+(1-\tau)\sum_{k=1}^Kw_kN_T
   \Big(\prod_{i\not=k}N_i\Big)\sum_{n=1}^{N_k}|\mathrm{E}^{(S_k)}\widehat{f}-\widehat{f}({\bf z'}_{n}^{(k)})|\geq\frac{\xi'N_T\prod_{k=1}^KN_k}{2}\right\}\nonumber
\end{align*}
\begin{align}\label{eq:the:sym.pr3.MC}
\leq& \frac{4}{N_T^2\big(\prod_{k=1}^KN_k\big)^2(\xi')^2}
\mathrm{E}\left\{\tau^2\big(\prod_{k=1}^KN_k\big)^2\sum_{n=1}^{N_T}
|\mathrm{E}^{(T)}\widehat{f}-\widehat{f}({\bf z'}_n^{(T)})|^2\right.\nonumber\\
&\qquad\qquad\qquad\left.+
(1-\tau)^2N^2_T\sum_{k=1}^K\sum_{n=1}^{N_k}w_k^2\big(\prod_{i\not=k}N_i\big)^2
\big|\mathrm{E}^{(S_k)}\widehat{f}-\widehat{f}({\bf z'}_n^{(k)})\big|^2\right\}\nonumber\\
\leq& \frac{4\left(\tau^2\big(\prod_{k=1}^{K}N_k\big)^2N_T(b-a)^2+(1-\tau)^2N^2_T\sum_{k=1}^Kw^2_k
\big(\prod_{i\not=k}N_i\big)^2
N_k\left(b-a\right)^2\right)}{N_T^2\big(\prod_{k=1}^KN_k\big)^2(\xi')^2}\nonumber\\
=& \frac{4\tau^2(b-a)^2}{N_T(\xi')^2}+\sum_{k=1}^K\frac{4(1-\tau)^2w_k^2(b-a)^2}{N_k(\xi')^2}.
%\leq&\frac{4(b-a)^2}{\xi^2\sum_{k=1}^KN_k}
\end{align}

Subsequently, according to \eqref{eq:the:sym.pr2.MC} and \eqref{eq:the:sym.pr3.MC}, we have for any $\xi'>0$,
\begin{align}\label{eq:the:sym.pr4.MC}
   &\mathrm{Pr'}\left\{\big|\tau(\mathrm{E'}^{(T)}_{N_T}\widehat{f}
-\mathrm{E}^{(T)}_{N_T}\widehat{f})
+(1-\tau)(\mathrm{E'}^{(S)}_{{\bf w}}\widehat{f}-\mathrm{E}^{(S)}_{{\bf w}}\widehat{f})\big|>\frac{\xi'}{2}\right\}\nonumber\\
\geq&
   \left(\mathbf{1}_{|\tau(\mathrm{E}^{(T)}\widehat{f}
-\mathrm{E}^{(T)}_{N_T}\widehat{f})
+(1-\tau)(\overline{\mathrm{E}}^{(S)}
\widehat{f}-\mathrm{E}^{(S)}_{{\bf w}}\widehat{f})|>\xi'}\right)\left(1-\left(\frac{4\tau^2(b-a)^2}{N_T(\xi')^2}
+\sum_{k=1}^K\frac{4(1-\tau)^2w_k^2(b-a)^2}{N_k(\xi')^2}\right)\right).
\end{align}
According to \eqref{eq:sym01.MC}, \eqref{eq:xi1.MC} and \eqref{eq:the:sym.pr4.MC}, taking the expectation with respect to $\{{\bf Z}_1^{N_k}\}_{k=1}^K$ and  ${\bf \overline{Z'}}_{1}^{N_T}$ and letting
$$\frac{4\tau^2(b-a)^2}{N_T(\xi')^2}
+\sum_{k=1}^K\frac{4(1-\tau)^2w_k^2(b-a)^2}{N_k(\xi')^2}\leq \frac{1}{2},$$
we then have for any $\xi>(1-\tau)D_\mathcal{F}^{({\bf w})}(S,T)$,
\begin{align*}%\label{eq:sym.result1.M}
  &\mathrm{Pr}\left\{|\mathrm{E}^{(T)}\widehat{f}-\mathrm{E}^{\tau}_{\bf w}\widehat{f}|>\xi\right\}\nonumber\\
 \leq& \mathrm{Pr}\left\{|\tau(\mathrm{E}^{(T)}\widehat{f}
-\mathrm{E}^{(T)}_{N_T}\widehat{f})
+(1-\tau)(\overline{\mathrm{E}}^{(S)}
\widehat{f}-\mathrm{E}^{(S)}_{{\bf w}}\widehat{f})|>\xi'\right\}\nonumber\\
 \leq& 2\mathrm{Pr'}\left\{\big|\tau(\mathrm{E'}^{(T)}_{N_T}\widehat{f}
-\mathrm{E}^{(T)}_{N_T}\widehat{f})
+(1-\tau)(\mathrm{E'}^{(S)}_{{\bf w}}\widehat{f}-\mathrm{E}^{(S)}_{{\bf w}}\widehat{f})\big|>\frac{\xi'}{2}\right\}
\end{align*}
with $\xi'=\xi-(1-\tau)D_\mathcal{F}^{({\bf w})}(S,T)$.
This completes the proof.
\hfill$\blacksquare$

%%

%%%%% Proof of Bound -Hoeffding
\subsection{Proof of Theorem \ref{thm:Hcrate.MC}}

{\it Proof of Theorem \ref{thm:Hcrate.MC}.} Consider $\epsilon$ as an independent Rademacher random variable, that is, an independent $\{-1,1\}$-valued random variable with equal probability of taking either value. Given sample sets $\{{\bf Z}_{1}^{2N_k}\}_{k=1}^K$ and $\overline{{\bf Z}}_{1}^{2N_T}$, denote for any $f\in\mathcal{F}$ and $1\leq k\leq K$,
\begin{align}\label{eq:epsilon.MC}
    \overrightarrow{\epsilon}^{(k)}:=&(\epsilon^{(k)}_1,\cdots,
    \epsilon^{(k)}_{N_k},
    -\epsilon^{(k)}_1,\cdots,-\epsilon^{(k)}_{N_k})
    \in\{\pm 1\}^{2N_k},\nonumber\\
\overrightarrow{\epsilon}_T:=&(\epsilon_1,\cdots,\epsilon_{N_T},
-\epsilon_1,\cdots,-\epsilon_{N_T})\in\{\pm 1\}^{2N_T},
\end{align}
and
\begin{align}\label{eq:vecf.MC}
  \overrightarrow{f}({\bf Z}_{1}^{2N_k}):=&\big(f({\bf z'}^{(k)}_1),\cdots,f({\bf z'}^{(k)}_{N_k}),f({\bf z}^{(k)}_1),\cdots,f({\bf z}^{(k)}_{N_k})\big)\in[a,b]^{2N_k};\nonumber\\
\overrightarrow{f}({\bf Z}_{1}^{2N_T}):=&\big(f({\bf z}'_1),\cdots,f({\bf z}'_{N_T}),f({\bf z}_1),\cdots,f({\bf z}_{N_T})\big)\in[a,b]^{2N_T}.
\end{align}

According to \eqref{eq:short1.MC}, \eqref{eq:short2.MC} and Theorem \ref{thm:sym.MC}, given any $\xi>(1-\tau)D_\mathcal{F}^{({\bf w})}(S,T)$, we have for any $N_T,N_1,\cdots,N_K\in\mathbb{N}$
such that
\begin{equation*}
  \frac{\tau^2(b-a)^2}{N_T(\xi')^2}
+\sum_{k=1}^K\frac{(1-\tau)^2w_k^2(b-a)^2}{N_k(\xi')^2}\leq \frac{1}{8}
\end{equation*}
 with $\xi'=\xi-(1-\tau)D_\mathcal{F}^{({\bf w})}(S,T)$,
\begin{align}\label{eq:bas1.MC}
    &\mathrm{Pr}\left\{\sup_{f\in\mathcal{F}}
    \big|\mathrm{E}^{(T)}f-\mathrm{E}^{\tau}_{\bf w}f\big|>\xi\right\}\nonumber\\
\leq& 2\mathrm{Pr}\left\{\sup_{f\in\mathcal{F}}
\big|\mathrm{E'}^{\tau}_{\bf w}f-\mathrm{E}_{\bf w}^{\tau}f\big|>\frac{\xi'}{2}\right\}\qquad\mbox{(by Theorem \ref{thm:sym.MC})}\nonumber\\
=&2\mathrm{Pr}\left\{\sup_{f\in\mathcal{F}}
\Big|\frac{\tau}{N_T}\sum_{n=1}^{N_T}\big(f({\bf z'}^{(T)}_n)-f({\bf z}^{(T)}_n)\big)+\sum_{k=1}^K\frac{(1-\tau)w_k}{N_k}
\sum_{n=1}^{N_k}\big(f({\bf z'}^{(k)}_n)-f({\bf z}^{(k)}_n)\big)\Big|>\frac{\xi'}{2}\right\}\nonumber\\
=&2\mathrm{Pr}\left\{\sup_{f\in\mathcal{F}}
\Big|\frac{\tau}{N_T}\sum_{n=1}^{N_T}\epsilon_n\big(f({\bf z'}^{(T)}_n)-f({\bf z}^{(T)}_n)\big)\right.\nonumber\\
&\left.+ \sum_{k=1}^K\frac{(1-\tau)w_k}{N_k}\sum_{n=1}^{N_k}
\epsilon^{(k)}_n\big(f({\bf z'}^{(k)}_n)-f({\bf z}^{(k)}_n)\big)\Big|>\frac{\xi'}{2}\right\}\nonumber\\
=&2\mathrm{Pr}\left\{\sup_{f\in\mathcal{F}}
\Big|\frac{\tau}{2N_T}\langle\overrightarrow{\epsilon}_T,
\overrightarrow{f}({\bf Z}_{1}^{2N_T})
\rangle+\sum_{k=1}^K\frac{(1-\tau)w_k}{2N_k}
\big\langle\overrightarrow{\epsilon}^{(k)},
\overrightarrow{f}({\bf Z}_{1}^{2N_k})\big\rangle\Big|>\frac{\xi'}{4}\right\}.
\end{align}

Fix a realization of $\{{\bf Z}_{1}^{2N_k}\}_{k=1}^K$ and $\overline{{\bf Z}}_{1}^{2N_T}$, and let $\Lambda$ be a $\xi'/8$-radius cover of $\mathcal{F}$ with respect to the $\ell_1^{{\bf w},\tau}(\{{\bf Z}_{1}^{2N_k}\}_{k=1}^K,\overline{{\bf Z}}_{1}^{2N_T})$ norm.
Since $\mathcal{F}$ is composed of the bounded functions with the range $[a,b]$, we assume that the same holds for any $h\in\Lambda$. If $f_0$ is the function that achieves the following supremum
\begin{equation*}
    \sup_{f\in\mathcal{F}}
\Big|\frac{\tau}{2N_T}\langle\overrightarrow{\epsilon}_T,
\overrightarrow{f}(\overline{{\bf Z}}_{1}^{2N_T})
\rangle+\sum_{k=1}^K\frac{(1-\tau)w_k}{2N_k}
\big\langle\overrightarrow{\epsilon}^{(k)},
\overrightarrow{f}({\bf Z}_{1}^{2N_k})\big\rangle\Big|>\frac{\xi'}{4},
\end{equation*}
there must be an $h_0\in\Lambda$ that satisfies
\begin{align*}
    &\frac{\tau}{2N_T}\sum_{n=1}^{N_T}\left(|f_0({\bf z'}^{(T)}_n)-h_0({\bf z'}^{T}_n)|+|f_0({\bf z}^{(T)}_n)-h_0({\bf z}^{(T)}_n)|\right)\nonumber\\
&+\sum_{k=1}^K\frac{(1-\tau)w_k}{2N_k}\sum_{n=1}^{N_k}\left(|f_0({\bf z'}^{(k)}_n)-h_0({\bf z'}^{(k)}_n)|+|f_0({\bf z}^{(k)}_n)-h_0({\bf z}^{(k)}_n)|\right)< \frac{\xi'}{8},
\end{align*}
and meanwhile,
\begin{equation*}
   \Big|\frac{\tau}{2N_T}\big\langle\overrightarrow{\epsilon}_T,
\overrightarrow{h}_0(\overline{{\bf Z}}_{1}^{2N_T})
\big\rangle+ \sum_{k=1}^K\frac{(1-\tau)w_k}{2N_k}
    \big\langle\overrightarrow{\epsilon}^{(k)},
    \overrightarrow{h_0}({\bf Z}_{1}^{2N_k})\big\rangle\Big|>\frac{\xi'}{8}.
\end{equation*}
Therefore, we arrive at
\begin{align}\label{eq:bas2.MC}
&\mathrm{Pr}\left\{\sup_{f\in\mathcal{F}}
\Big|\frac{\tau}{2N_T}\langle\overrightarrow{\epsilon}_T,
\overrightarrow{f}(\overline{{\bf Z}}_{1}^{2N_T})
\rangle+\sum_{k=1}^K\frac{(1-\tau)w_k}{2N_k}
\big\langle\overrightarrow{\epsilon}^{(k)},
\overrightarrow{f}({\bf Z}_{1}^{2N_k})\big\rangle\Big|>\frac{\xi'}{4}\right\}\nonumber\\
\leq&\mathrm{Pr}\left\{\sup_{h\in\Lambda}
\Big|\frac{\tau}{2N_T}\langle\overrightarrow{\epsilon}_T,
\overrightarrow{h}(\overline{{\bf Z}}_{1}^{2N_T})
\rangle+\sum_{k=1}^K\frac{(1-\tau)w_k}{2N_k}
\big\langle\overrightarrow{\epsilon}^{(k)},
\overrightarrow{h}({\bf Z}_{1}^{2N_k})\big\rangle\Big|>\frac{\xi'}{8}\right\}.
\end{align}

 According to \eqref{eq:epsilon.MC}, \eqref{eq:vecf.MC} and Theorem \ref{thm:Hdineq.MC}, we have
\begin{align}\label{eq:bas3.MC}
&\mathrm{Pr}\left\{\sup_{h\in\Lambda}
\Big|\frac{\tau}{2N_T}\langle\overrightarrow{\epsilon}_T,
\overrightarrow{h}(\overline{{\bf Z}}_{1}^{2N_T})
\rangle+\sum_{k=1}^K\frac{(1-\tau)w_k}{2N_k}
\big\langle\overrightarrow{\epsilon}^{(k)},
\overrightarrow{h}({\bf Z}_{1}^{2N_k})\big\rangle\Big|>\frac{\xi'}{8}\right\}\\
=&
\mathrm{Pr}\left\{\sup_{h\in\Lambda}\big|\mathrm{E'}^{\tau}_{\bf w}h-\mathrm{E}^{\tau}_{\bf w}h\big|>
\frac{\xi'}{4}\right\}\nonumber\\
\leq&
\mathrm{Pr}\left\{\sum_{h\in\Lambda}\big|\mathrm{E'}^{\tau}_{\bf w}h-\mathrm{E}^{\tau}_{\bf w}h\big|>
\frac{\xi'}{4}\right\}\nonumber\\
\leq&\mathrm{Pr}
\left\{\sum_{h\in\Lambda}|\mathrm{E}^{(*)}h-\mathrm{E'}^{\tau}_{\bf w}h|
+|\mathrm{E}^{(*)}h-\mathrm{E}^{\tau}_{\bf w}h|
>\frac{\xi'}{4}\right\}\nonumber\\
\leq&
2
\mathrm{Pr}\left\{\sum_{h\in\Lambda}\big|\mathrm{E}^{(*)}h-\mathrm{E}^{\tau}_{\bf w}h\big|>\frac{\xi'}{8}\right\}\nonumber\\
\leq&4\mathcal{N}^{{\bf w},\tau}_1(\mathcal{F},\xi'/8,2{\bf N})\exp\left\{-\frac{(\xi')^2}
{32(b-a)^2\big(\frac{\tau^2}{N_T}+
\sum_{k=1}^{K}\frac{(1-\tau)^2w_k^2}{N_k}\big)}\right\},\nonumber
\end{align}
where $\xi'=\xi-(1-\tau)D^{({\bf w})}_{\mathcal{F}}(S,T)$ and ${\bf N}=N_T+\sum_{k=1}^KN_k$.

The combination of \eqref{eq:bas1.MC}, \eqref{eq:bas2.MC} and \eqref{eq:bas3.MC} leads to the result:  given any $\xi>(1-\tau)D^{({\bf w})}_{\mathcal{F}}(S,T)$ and for any $N_T\big(\prod_{k=1}^KN_k\big)\geq \frac{8\left(b-a\right)^2}{(\xi')^2}$ with $\xi'=\xi-(1-\tau)D^{({\bf w})}_{\mathcal{F}}(S,T)$,
\begin{align}\label{eq:RB1_1.MC}
    &\mathrm{Pr}\left\{\sup_{f\in\mathcal{F}}
    \big|\mathrm{E}^{(T)}f-\mathrm{E}^{\tau}_{\bf w}f\big|>\xi\right\}\nonumber\\
        \leq&8\mathcal{N}^{{\bf w},\tau}_1(\mathcal{F},\xi'/8,2{\bf N})
       \exp\left\{-\frac{(\xi')^2}
{32(b-a)^2\big(\frac{\tau^2}{N_T}+
\sum_{k=1}^{K}\frac{(1-\tau)^2w_k^2}{N_k}\big)}\right\}.
        \end{align}

According to \eqref{eq:RB1_1.MC}, letting
\begin{align*}%\label{eq:epsilon}
    \epsilon:=8\mathcal{N}^{{\bf w},\tau}_1(\mathcal{F},\xi'/8,2{\bf N})  \exp\left\{-\frac{(\xi')^2}
{32(b-a)^2\big(\frac{\tau^2}{N_T}+
\sum_{k=1}^{K}\frac{(1-\tau)^2w_k^2}{N_k}\big)}\right\},
\end{align*}
we then arrive at with probability at least $1-\epsilon$,
\begin{align*}%\label{eq:crate1}
    \sup_{f\in\mathcal{F}}
\big|\mathrm{E}^{\tau}_{\bf w}f-\mathrm{E}^{(T)}f\big|
\leq (1-\tau)D^{({\bf w})}_{\mathcal{F}}(S,T)+ \left(\frac{\ln\mathcal{N}^{{\bf w},\tau}_1(\mathcal{F},\xi'/8,2{\bf N})-\ln(\epsilon/8)}
{\frac{1}{32(b-a)^2\big(\frac{\tau^2}{N_T}+
\sum_{k=1}^{K}\frac{(1-\tau)^2w_k^2}{N_k}\big)}}\right)^{\frac{1}{2}},
\end{align*}
where $\xi'=\xi-D^{({\bf w})}_{\mathcal{F}}(S,T)$.
This completes the proof. \hfill$\blacksquare$

%%%%

%% Proof of Bound - Bennett
%%
\subsection{Proof of Theorem \ref{thm:RB1.MC}}\label{app:pf.RB1.MC}
By using the resulted deviation inequality \eqref{eq:dineq.MC} and the symmetrization inequality, we can achieve the proof of Theorem \ref{thm:RB1.MC}.

{\it Proof of Theorem \ref{thm:RB1.MC}.}
Recall the proof of Theorem \ref{thm:Hcrate.MC} and set $w_k=N_k/\sum_{k=1}^KN_k$ ($1\leq k\leq K$) and $\tau=N_T/(N_T+\sum_{k=1}^KN_k)$. Following the way of obtaining \eqref{eq:bas3.MC}, we have
\begin{align}\label{eq:bas3.MC.B}
&\mathrm{Pr}\left\{\sup_{h\in\Lambda}
\Big|\frac{\tau}{2N_T}\langle\overrightarrow{\epsilon}_T,
\overrightarrow{h}(\overline{{\bf Z}}_{1}^{2N_T})
\rangle+\sum_{k=1}^K\frac{(1-\tau)w_k}{2N_k}
\big\langle\overrightarrow{\epsilon}^{(k)},
\overrightarrow{h}({\bf Z}_{1}^{2N_k})\big\rangle\Big|>\frac{\xi'}{8}\right\}\nonumber\\
\leq&4\mathcal{N}^{{\bf w},\tau}_1(\mathcal{F},\xi'/8,2{\bf N})\exp\left\{\Big(N_T+\sum_{k=1}^KN_k\Big)\Gamma\left( \frac{\xi'}{(b-a)}\right)\right\},
\end{align}
where $\xi'=\xi-(1-\tau)D^{({\bf w})}_{\mathcal{F}}(S,T)$ and ${\bf N}=N_T+\sum_{k=1}^KN_k$. The combination of \eqref{eq:bas1.MC}, \eqref{eq:bas2.MC} and \eqref{eq:bas3.MC.B} leads to the result:  given any $\xi>(1-\tau)D^{({\bf w})}_{\mathcal{F}}(S,T)$ and for any $N_T\big(\prod_{k=1}^KN_k\big)\geq \frac{8\left(b-a\right)^2}{(\xi')^2}$ with $\xi'=\xi-(1-\tau)D^{({\bf w})}_{\mathcal{F}}(S,T)$,
\begin{align*}%\label{eq:RB1_1.MC}
    &\mathrm{Pr}\left\{\sup_{f\in\mathcal{F}}
    \big|\mathrm{E}^{(T)}f-\mathrm{E}^{\tau}_{\bf w}f\big|>\xi\right\}\nonumber\\
        \leq&8\mathcal{N}^{{\bf w},\tau}_1(\mathcal{F},\xi'/8,2{\bf N})
       \exp\left\{\Big(N_T+\sum_{k=1}^KN_k\Big)\Gamma\left( \frac{\xi'}{(b-a)}\right)\right\}.
        \end{align*}
 This completes the proof. \hfill$\blacksquare$

%%%%%%%

\subsection{Proof of Theorem \ref{thm:RB.Rade.MC}}

In order to prove Theorem \ref{thm:RB.Rade.MC}, we also need the following result \citep[see][]{Bousquet04}:

\begin{theorem}\label{thm:b2}
Let $\mathcal{F}\subseteq[a,b]^{\mathcal{Z}}$. For any $\epsilon>0$, with probability at least $1-\epsilon$, there holds that for any $f\in\mathcal{F}$,
\begin{align*}%\label{eq:bound.T2}
\mathrm{E}f\leq&\mathrm{E}_Nf+2\mathcal{R}(\mathcal{F})
+\sqrt{\frac{(b-a)\ln(1/\epsilon)}{2N}}\nonumber\\
\leq& \mathrm{E}_Nf+2\mathcal{R}_N(\mathcal{F})
+3\sqrt{\frac{(b-a)\ln(2/\epsilon)}{2N}}.
\end{align*}
\end{theorem}

By using Theorem \ref{thm:Mcdiarmid} and Theorem \ref{thm:b2}, we prove Theorem \ref{thm:RB.Rade.MC} as follows:

{\it Proof of Theorem \ref{thm:RB.Rade.MC}.} Assume that the function class $\mathcal{F}$ is composed of bounded functions with the range $[a,b]$. Let $\{{\bf Z}_{1}^{N_k}\}_{k=1}^K:=\{\{{\bf z}_{n}^{(k)}\}_{n=1}^{N_k}\}_{k=1}^K$ and $\overline{{\bf Z}}_{1}^{N_{T}}:=\{{\bf z}_n^{(T)}\}_{n=1}^{N_T}$ be the sample sets  drawn from multiple sources $\mathcal{Z}^{(S_k)}$ ($1\leq k\leq K$) and the target domain $\mathcal{Z}^{(T)}$, respectively.

Given $\tau\in[0,1)$ and  ${\bf w}\in[0,1]^{K}$ with $\sum_{k=1}^Kw_k=1$, denote
\begin{equation}\label{eq:Proof.Rad1.MC}
H\big({\bf Z}_{1}^{N_1},\cdots,{\bf Z}_{1}^{N_K},\overline{{\bf Z}}_{1}^{N_{T}}\big):=\sup_{f\in\mathcal{F}}\big|\mathrm{E}_{{\bf w}}^{\tau}f-\mathrm{E}^{(T)}f\big|.
\end{equation}
By \eqref{eq:emrisk.MC}, we have
\begin{equation}\label{eq:Proof.Rad2}
H\big({\bf Z}_{1}^{N_1},\cdots,{\bf Z}_{1}^{N_K}\big)
=\sup_{f\in\mathcal{F}}\Big|\tau(\mathrm{E}^{(T)}_{N_T}f-\mathrm{E}^{(T)}f)
+(1-\tau)\sum_{k=1}^Kw_k\big(\mathrm{E}_{N_k}^{(S_k)}f-\mathrm{E}^{(T)}f\big)\Big|,
\end{equation}
where $\mathrm{E}_{N_k}^{(S_k)}f=\frac{1}{N_k}\sum_{n=1}^{N_k}f({\bf z}_n^{(k)})$. Therefore, it is clear that such $H\big({\bf Z}_{1}^{N_1},\cdots,{\bf Z}_{1}^{N_K},\overline{{\bf Z}}_{1}^{N_{T}}\big)$ satisfies the condition of bounded difference with
\begin{equation*}
   c_n^{(k)}=\frac{(1-\tau)(b-a)w_k}{N_k},\;1\leq k\leq K,\;1\leq n\leq N_k;\quad \overline{c}_n=\frac{\tau(b-a)}{N_T},\;1\leq n\leq N_T.
\end{equation*}
Thus, according to Theorem \ref{thm:Mcdiarmid}, we have for any $\xi>0$,
\begin{align*}
&\mathrm{Pr}\Big\{H\big({\bf Z}_{1}^{N_1},\cdots,{\bf Z}_{1}^{N_K},\overline{{\bf Z}}_{1}^{N_{T}}\big)-\mathrm{E}\big\{H\big({\bf Z}_{1}^{N_1},\cdots,{\bf Z}_{1}^{N_K},\overline{{\bf Z}}_{1}^{N_{T}}\big)\big\}\geq\xi\Big\}\nonumber\\
\leq&\exp\left\{\frac{-2\xi^2}
{(b-a)^2\left(\frac{\tau^2}{N_T}
+\sum_{k=1}^K\frac{(1-\tau)^2w_k^2}{N_k}\right)}\right\},
\end{align*}
which can be equivalently rewritten as with probability at least $1-(\epsilon/2)$,
\begin{align}\label{eq:Proof.Rad3}
&H\big({\bf Z}_{1}^{N_1},\cdots,{\bf Z}_{1}^{N_K},\overline{{\bf Z}}_{1}^{N_{T}}\big)\\
\leq&\mathrm{E}\big\{H\big({\bf Z}_{1}^{N_1},\cdots,{\bf Z}_{1}^{N_K},\overline{{\bf Z}}_{1}^{N_{T}}\big)\big\}+
\sqrt{\frac{(b-a)^2\ln(2/\epsilon)}{2}\left(\frac{\tau^2}{N_T}
+\sum_{k=1}^K\frac{(1-\tau)^2w_k^2}{N_k}\right)}\nonumber\\
=&\mathrm{E}
   \left\{\sup_{f\in\mathcal{F}}\Big|\tau(\mathrm{E}^{(T)}_{N_T}f
   -\mathrm{E}^{(T)}f)
+(1-\tau)\sum_{k=1}^Kw_k\big(\mathrm{E}_{N_k}^{(S_k)}f
-\mathrm{E}^{(T)}f\big)\Big|\right\}\nonumber\\
&+
\sqrt{\frac{(b-a)^2\ln(2/\epsilon)}{2}\left(\frac{\tau^2}{N_T}
+\sum_{k=1}^K\frac{(1-\tau)^2w_k^2}{N_k}\right)}\nonumber\\
\leq&\tau\mathrm{E}^{(T)}\left\{\sup_{f\in\mathcal{F}}|\mathrm{E}^{(T)}_{N_T}f-
\mathrm{E}^{(T)}f|\right\}+(1-\tau)
\mathrm{E}^{(S)}
   \left\{\sup_{f\in\mathcal{F}}\Big|\sum_{k=1}^Kw_k
\big(\mathrm{E}_{N_k}^{(S_k)}f
   -\mathrm{E}^{(S_k)}f\big)\Big|\right\}\nonumber\\
   &+(1-\tau)
   \sum_{k=1}^Kw_k\sup_{f\in\mathcal{F}}\big|\mathrm{E}^{(S_k)}f
-\mathrm{E}^{(T)}f\big|
 +
\sqrt{\frac{(b-a)^2\ln(2/\epsilon)}{2}\left(\frac{\tau^2}{N_T}
+\sum_{k=1}^K\frac{(1-\tau)^2w_k^2}{N_k}\right)}.\nonumber
\end{align}

%  D^{({\bf w})}_{\mathcal{F}}(S,T)

According to \eqref{eq:ExRade}, we have
\begin{align}\label{eq:element1}
   &\mathrm{E}^{(S)}\sup_{f\in\mathcal{F}}\Big|\sum_{k=1}^Kw_k
\big(\mathrm{E}_{N_k}^{(S_k)}f
   -\mathrm{E}^{(S_k)}f\big)\Big|\nonumber\\
   =&\mathrm{E}^{(S)}\sup_{f\in\mathcal{F}}
\Big|\sum_{k=1}^Kw_k\big(\mathrm{E}_{N_k}^{(S_k)}f-\mathrm{E'}^{(S_k)}
   \{\mathrm{E'}_{N_k}^{(S_k)}f\}\big)\Big|\nonumber\\
   \leq&\mathrm{E}^{(S)}\mathrm{E'}^{(S)}\sup_{f\in\mathcal{F}}
   \Big|\sum_{k=1}^Kw_k\big(\mathrm{E}_{N_k}^{(S_k)}f
-\mathrm{E'}_{N_k}^{(S_k)}f\big)\Big|\nonumber\\
   =&\mathrm{E}^{(S)}\mathrm{E'}^{(S)}\sup_{f\in\mathcal{F}}
   \Big|\sum_{k=1}^Kw_k\frac{1}{N_k}\sum_{n=1}^{N_k}\big(f({\bf z}_n^{(k)})-f({\bf z'}_n^{(k)})\big)\Big|\nonumber\\
   =&\mathrm{E}^{(S)}\mathrm{E'}^{(S)}\mathrm{E}_{\sigma}\sup_{f\in\mathcal{F}}
   \Big|\sum_{k=1}^Kw_k\frac{1}{N_k}\sum_{n=1}^{N_k}\sigma_{n}^{(k)}\big(f({\bf z}_n^{(k)})-f({\bf z'}_n^{(k)})\big)\Big|\nonumber\\
   \leq&2\mathrm{E}^{(S)}\mathrm{E}_{\sigma}\sup_{f\in\mathcal{F}}
   \Big|\sum_{k=1}^Kw_k\frac{1}{N_k}\sum_{n=1}^{N_k}\sigma_{n}^{(k)}f({\bf z}_n^{(k)})\Big|\nonumber\\
   \leq&2\mathrm{E}^{(S)}\mathrm{E}_{\sigma}\sum_{k=1}^Kw_k\frac{1}{N_k}\sup_{f\in\mathcal{F}}
   \Big|\sum_{n=1}^{N_k}\sigma_{n}^{(k)}f({\bf z}_n^{(k)})\Big|\nonumber\\
   =&2\sum_{k=1}^Kw_k\mathcal{R}^{(k)}(\mathcal{F}),
\end{align}
and similarly,
\begin{align*}%\label{eq:element1.5}
   \mathrm{E}^{(T)}\left\{\sup_{f\in\mathcal{F}}|\mathrm{E}^{(T)}_{N_T}f-
\mathrm{E}^{(T)}f|\right\}\leq2\mathcal{R}^{(T)}(\mathcal{F}).
\end{align*}

Again, let
$G({\bf z}_1,\cdots,{\bf z}_N)=\mathcal{R}_{N}(\mathcal{F})$. It is clear that $G$ satisfies the condition \eqref{eq:dif.cond} of Theorem \ref{thm:Mcdiarmid0} with $c=(b-a)/N$. Similarly, we have with probability at least $1-\epsilon/2$
\begin{equation}\label{eq:element2}
   \mathcal{R}(\mathcal{F})\leq \mathcal{R}_{N}(\mathcal{F})+
   (b-a)\left(\frac{\ln(4/\epsilon)}{2N}\right)^{\frac{1}{2}}.
\end{equation}

%%

%On the other hand, it is followed from Theorem \ref{thm:b2} that for any $\epsilon>0$, with at least $1-(\epsilon/2)$,
%\begin{equation}\label{eq:element2}
%\sup_{f\in\mathcal{F}}\Big|\mathrm{E}^{(T)}_{N_T}f
%-\mathrm{E}^{(T)}f\Big|\leq2\mathcal{R}^{(T)}_{N_T}(\mathcal{F})
%+3\sqrt{\frac{(b-a)\ln(4/\epsilon)}{2N_T}}.
%\end{equation}

By combining \eqref{eq:Dist.MC}, \eqref{eq:Proof.Rad1.MC}, \eqref{eq:Proof.Rad3}, \eqref{eq:element1} and \eqref{eq:element2}, we arrive at with probability at least $1-\epsilon$,
\begin{align}\label{eq:lastone}
\sup_{f\in\mathcal{F}}\big|\mathrm{E}_{{\bf w}}^{\tau}f-\mathrm{E}^{(T)}f\big|
\leq& (1-\tau) D^{({\bf w})}_{\mathcal{F}}(S,T)+
2(1-\tau)\sum_{k=1}^Kw_k\mathcal{R}^{(k)}(\mathcal{F})\nonumber\\
&+2\tau\mathcal{R}^{(T)}_{N_T}(\mathcal{F})
+2\tau\sqrt{\frac{(b-a)^2\ln(4/\epsilon)}{2N_T}}\nonumber\\
&+\sqrt{\frac{(b-a)^2\ln(2/\epsilon)}{2}\left(\frac{\tau^2}{N_T}
+\sum_{k=1}^K\frac{(1-\tau)^2w_k^2}{N_k}\right)}.
\end{align}
This completes the proof. \hfill$\blacksquare$

%%%%%%

\end{document}